\RequirePackage[svgnames]{xcolor}

\documentclass[11pt,letterpaper]{mystyle}

\usepackage[all]{hypcap}
\usepackage[svgnames]{xcolor}
\usepackage[comma,authoryear,compress]{natbib}
\usepackage{hyperref}[citecolor=lightblue]

\hypersetup{
    colorlinks = true,
    citecolor = {YaleBlue},
}
\usepackage{pythonhighlight}
\usepackage{algorithm}
\usepackage{algorithmicx}
\usepackage{algpseudocode}
\usepackage{microtype}
\usepackage{graphicx}
\expandafter\def\csname ver@subfig.sty\endcsname{}
\usepackage{booktabs} %   
\usepackage{float}
\usepackage{bigstrut}
\usepackage{wrapfig}
\usepackage{amsmath}
\usepackage{amssymb}
\usepackage{mathtools}
\usepackage{amsthm}
\usepackage{mathrsfs}
\usepackage{nicefrac}
\usepackage{dsfont}
\usepackage{enumitem}
\usepackage{subcaption}
\usepackage{graphicx,subfig}
\usepackage{cleveref}
\usepackage{bxcoloremoji}

\usepackage{amsmath} 
\usepackage{amssymb} 
\usepackage{pifont}  
\usepackage{booktabs} 
\usepackage{float}
\usepackage{amssymb} 
\usepackage[utf8]{inputenc}   
\usepackage[T1]{fontenc}      
\usepackage{lmodern}          
\usepackage{hyperref}       %
\usepackage{url}            %
\usepackage{booktabs}       %
\usepackage{amsfonts}       %
\usepackage{nicefrac}       %
\usepackage{microtype}      %
\usepackage{graphicx}
\usepackage{subcaption}
\usepackage{amssymb}
\usepackage{fdsymbol}
\usepackage{wrapfig}
\usepackage{lipsum}
\usepackage{enumitem}
\usepackage{stackengine}
\usepackage[font=small,labelfont=bf]{caption}
\usepackage{color}
\usepackage{adjustbox}
\usepackage{multirow}
\usepackage{rotating}
\usepackage{makecell}
\usepackage[dvipsnames, table]{xcolor}
\usepackage{subcaption}

\definecolor{g}{RGB}{189, 230, 205}
\definecolor{lg}{RGB}{228,238,188}     
\definecolor{y}{RGB}{255,248,197} 
\definecolor{r}{RGB}{255,200,190}
\definecolor{lr}{RGB}{255,248,227} 

\definecolor{myblue}{RGB}{31,119,180}

\definecolor{myellow1}{rgb}{0.94, 0.8, 0.0}
\colorlet{mygray}{white!15}
\colorlet{myellow}{lightgray!15}

\definecolor{rwkv}{RGB}{248, 181, 134}
\definecolor{cnn}{RGB}{238,131,115}     
\definecolor{transformer}{RGB}{131,121,182} 
\definecolor{mamba}{RGB}{76,200,132}
\definecolor{hybrid}{RGB}{46,117,182}

\newcommand{\x}{\mathbf{x}}

\definecolor{blanchedalmond}{rgb}{1.0, 0.92, 0.8}
\definecolor{carmine}{rgb}{0.59, 0.0, 0.09}
\definecolor{lightblue}{rgb}{0.22,0.45,0.70}%

\renewcommand{\mathbf}{\boldsymbol}

\makeatletter
\def\Ddots{\mathinner{\mkern1mu\raise\p@
\vbox{\kern7\p@\hbox{.}}\mkern2mu
\raise4\p@\hbox{.}\mkern2mu\raise7\p@\hbox{.}\mkern1mu}}
\makeatother

\definecolor{amaranth}{rgb}{0.9, 0.17, 0.31}
\definecolor{antiquebrass}{rgb}{0.8, 0.58, 0.46}
\definecolor{antiquefuchsia}{rgb}{0.57, 0.36, 0.51}
\definecolor{chromeyellow}{rgb}{0.31, 0.47, 0.26}

\newcommand{\github}{\raisebox{-1.5pt}{\includegraphics[height=1.05em]{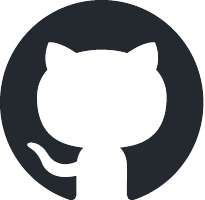}}}

\newcommand{\paperlogo}{\raisebox{-1.5pt}{\includegraphics[height=1.45em]{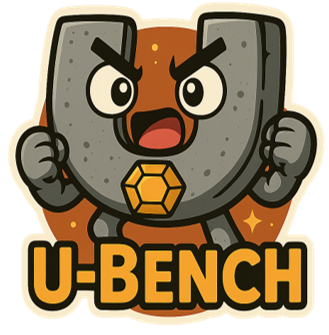}}}
\newcommand{\gold}{\raisebox{-4pt}{\includegraphics[height=1.25em]{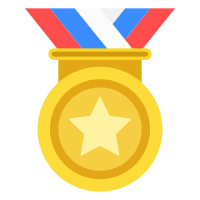}}}

\newcommand{\silver}{\raisebox{-4pt}{\includegraphics[height=1.25em]{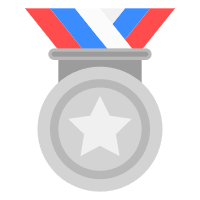}}}

\newcommand{\bronze}{\raisebox{-4pt}{\includegraphics[height=1.25em]{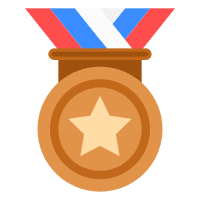}}}

\newtcolorbox{AIbox}[2][]{aibox,title=#2,#1}
\definecolor{lightblue}{rgb}{0.22,0.45,0.70}%
\definecolor{Gray}{gray}{0.95}
\definecolor{Cornsilk}{rgb}{1.0, 0.97, 0.86}

\usepackage{amsmath}

\usepackage[all]{hypcap}

\title{\paperlogo{} U-Bench: A Comprehensive Understanding of U-Net through 100-Variant Benchmarking}

\runningtitle{\paperlogo{} U-Bench: A Comprehensive Understanding of U-Net through 100-Variant Benchmarking}

\author{
  Fenghe Tang$^{1,2}$,
  Chengqi Dong$^{1,2}$,
  Wenxin Ma$^{1,2}$,
  Zikang Xu$^{3}$,
  Heqin Zhu$^{1,2}$, \\
  Zihang Jiang$^{1,2}$,
  Rongsheng Wang$^{1,2}$,
  Yuhao Wang$^{1,2}$,
  Chenxu Wu$^{1,2}$,
  Shaohua Kevin Zhou$^{1,2}$
}

\affil[1]{University of Science and Technology of China~\footnote{School of Biomedical Engineering, Division of Life Sciences and Medicine, University of Science and Technology of China (USTC), Hefei, Anhui, 230026, P.R. China}}
\affil[2]{MIRACLE Center~\footnote{Center for Medical Imaging, Robotics, and Analytic Computing \& LEarning (MIRACLE), Suzhou Institute for Advanced Research, USTC, Suzhou 215123, P.R. China, and Jiangsu Provincial Key Laboratory of Multimodal Digital Twin Technology, Suzhou Jiangsu, 215123, China}}
\affil[3]{HCNS~\footnote{Anhui Province Key Laboratory of Biomedical Imaging and Intelligent Processing, Institute of Artificial Intelligence, Hefei Comprehensive National Science Center, Hefei 230026, China}}

\correspondingauthor{Shaohua Kevin Zhou; Email \href{skevinzhou@ustc.edu.cn}{skevinzhou@ustc.edu.cn}}

\begin{document}

\begin{abstract}
Over the past decade, U-Net has been the dominant architecture in medical image segmentation, leading to the development of thousands of U-shaped variants. Despite its widespread adoption, there is still no comprehensive benchmark to systematically evaluate their performance and utility, largely because of insufficient statistical validation and limited consideration of efficiency and generalization across diverse datasets. To bridge this gap, we present U-Bench, the first large-scale, statistically rigorous benchmark that evaluates 100 U-Net variants across 28 datasets and 10 imaging modalities. Our contributions are threefold: \textbf{(1) Comprehensive Evaluation}: U-Bench evaluates models along three key dimensions: statistical robustness, zero-shot generalization, and computational efficiency. We introduce a novel metric, U-Score, which jointly captures the performance-efficiency trade-off, offering a deployment-oriented perspective on model progress. \textbf{(2) Systematic Analysis and Model Selection Guidance}: We summarize key findings from the large-scale evaluation and systematically analyze the impact of dataset characteristics and architectural paradigms on model performance. Based on these insights, we propose a model advisor agent to guide researchers in selecting the most suitable models for specific datasets and tasks. \textbf{(3) Public Availability}: We provide all code, models, protocols, and weights, enabling the community to reproduce our results and extend the benchmark with future methods. In summary, U-Bench not only exposes gaps in previous evaluations but also establishes a foundation for fair, reproducible, and practically relevant benchmarking in the next decade of U-Net-based segmentation models.

\vspace{2mm}

\textit{Keywords: Benchmark, U-Net, Medical Image Segmentation, U-Score}

\vspace{5mm}

\coloremojicode{1F4C5} \textbf{Date}: October, 2025

\coloremojicode{1F3E0} \textbf{Projects}: \href{https://fenghetan9.github.io/ubench}{https://fenghetan9.github.io/ubench}

\github{} \textbf{Code Repository}: \href{https://github.com/FengheTan9/U-Bench}{https://github.com/FengheTan9/U-Bench}

\coloremojicode{1F917} \textbf{Model Weights \& Checkpoints}: \href{https://huggingface.co/FengheTan9/U-Bench}{https://huggingface.co/FengheTan9/U-Bench}

\coloremojicode{1F4DA} \textbf{Datasets}: \href{https://huggingface.co/FengheTan9/U-Bench}{https://huggingface.co/FengheTan9/U-Bench}

\coloremojicode{1F4E7} \textbf{Contact}: \href{fhtan9@mail.ustc.edu.cn}{fhtan9@mail.ustc.edu.cn}

\end{abstract}

\maketitle
\vspace{3mm}
\section{Introduction}

\begin{figure}[!]
  \centering
   \includegraphics[width=0.96\linewidth]{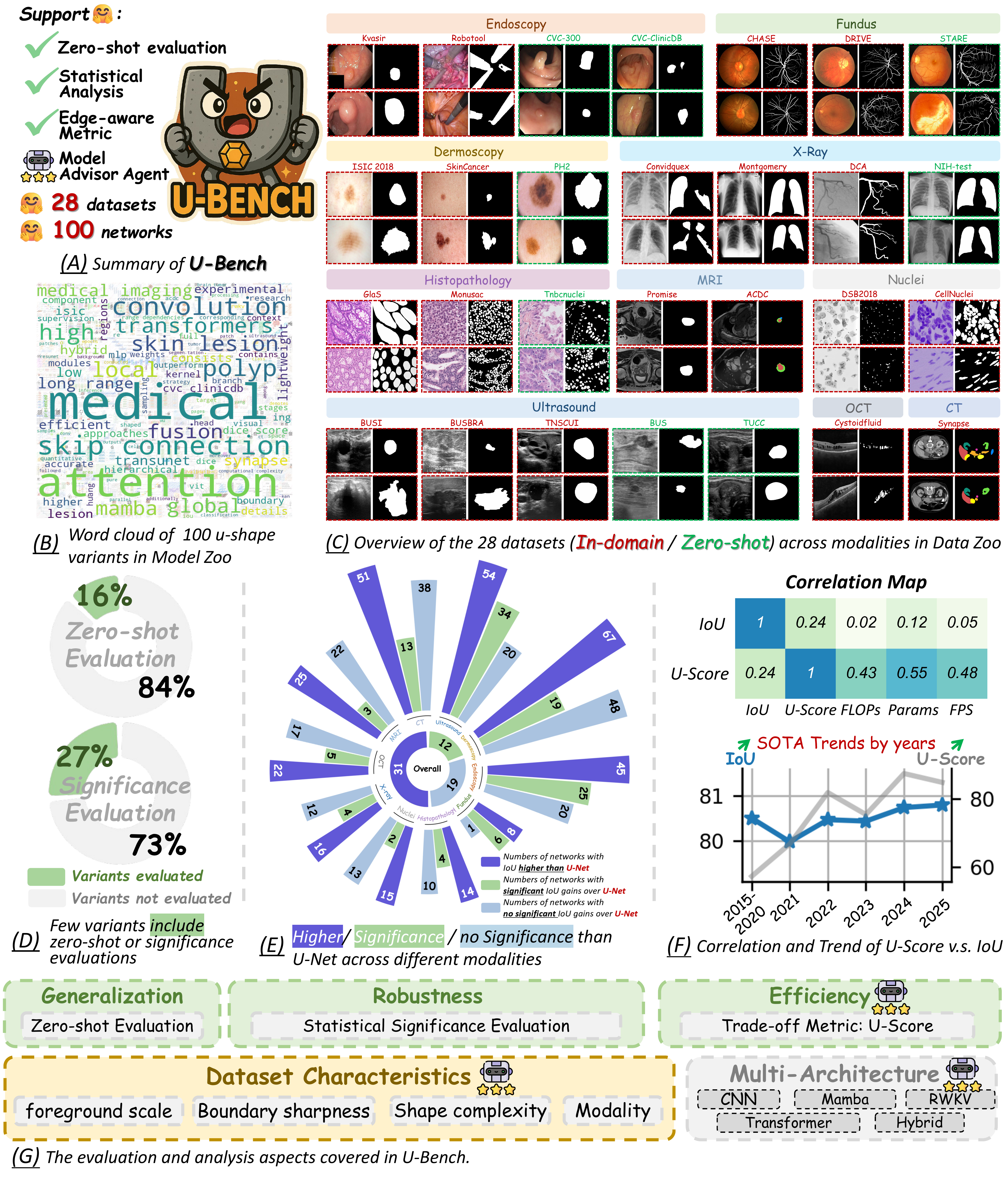}
    \vspace{-2mm}
   \caption{Overview of U-Bench. (A) \textbf{The summary of U-Bench}, which encompasses the most comprehensive large-scale evaluation of U-shaped architectures. (B)\textbf{ Word cloud of 100 published U-shaped variants in U-Bench Model Zoo.} (C) \textbf{Examples of the 28 datasets in U-Bench Data Zoo.} The \textcolor{red}{red} / \textcolor{green}{green} box: in-domain / zero-shot split for evaluation.  (D) \textbf{Literature analysis.} Among 100 recent works, 84\% papers neglect zero-shot evaluation and 73\% papers lack of statistical significance testing.  (E) \textbf{Significance analysis.} Only a minority achieve statistically significant gains over U-Net. (F) \textbf{Overview of a new metric, U-score.} Top: IoU does not account for efficiency, while U-Score demonstrates a strong correlation with both segmentation performance and efficiency metrics. Bottom: while IoU shows a trend of saturation, U-Score highlights the yearly trends toward more efficient models. (G) \textbf{The evaluation and analysis aspects covered in U-Bench.}}
   \label{fig:uteaser}
\end{figure}

Medical image segmentation is a critical and challenging task that can greatly enhance diagnostic efficiency by offering doctors objective and precise references for regions of interest~\citep{zhou2017deep}. Over the past decade, U-Net~\citep{U-Net} has become a cornerstone of medical image segmentation, thanks to its encoder-decoder structure with skip connections that effectively combine multi-scale features. Building on its promising segmentation results across diverse modalities, numerous U-shaped variants have been proposed to further improve performance, with lightweight designs~\citep{UNeXt,CMUNeXt,TinyU-Net,MedT,SwinUnet}, attention mechanisms~\citep{AtU-Net,CMU-Net}, multi-scale feature fusion~\citep{ResNet34UnetPlus,UNet3+}, and more recently Mamba-~\citep{Swin-UMamba,UltraLight-VM-UNet}, RWKV-based~\citep{U-RWKV,RWKV-UNet}, as well as hybrid architectures~\citep{TransUNet,MobileUViT,LGMSNet,llm4seg}. Over the past decade, more than ten thousand U-Net variants have been proposed, and by 2025, nearly a thousand studies have employed U-shaped networks for medical image segmentation.

Among the vast number of U-Net variants, a central challenge remains unresolved: \textit{\textbf{How to conduct a fair and comprehensive comparison across them?}}  Although several benchmarks and surveys have attempted to organize this proliferation (Tab.~\ref{tab:summary}), they mostly lack a large-scale, systematic evaluation. Critical aspects such as robustness of improvements, zero-shot generalization, and computational efficiency are often overlooked, and they also fail to provide complete and in-depth analyses of dataset-specific characteristics and model architectures. Despite reported gains in recent works, many studies report metrics without statistical validation (73\% omit it, Fig.~\ref{fig:uteaser}D), use incomplete baseline comparisons, or rely on limited dataset coverage. Moreover, efficiency, although vital for real-world clinical deployment~\citep{poc,realtime,edge}, is rarely considered. Compounding this issue, evaluations are typically confined to in-distribution settings (84\% of work ignores zero-shot evaluation, Fig.~\ref{fig:uteaser}D), even though clinical practice inevitably involves domain shifts across institutions and annotation protocols~\citep{domainshift,domainshift3}. These gaps leave the robustness and practicality of U-Net variants in real-world scenarios largely unverified~\citep{domainshift2}.

\begin{table*}[!t]
\centering
\caption{Comparisons between U-Bench and other medical image segmentation benchmarks. Details can be found in the Appendix~\ref{app:relate}. \label{tab:summary}}

\centering
\resizebox{\linewidth}{!}{
\begin{tabular}{l l | c c c c c}
\Xhline{1.5px}

\multirow{2}{*}{Category} & \multirow{2}{*}{Item} 
& U-Bench & TorchStone & nnWNet & MedSegBench & nnU-Net Revisited  \\

& & (ours) & \citep{torchstone} & \citep{nnwnet} & \citep{nnwnet} & \citep{nnunetrevisted} \\
 \hline
\textbf{Models} & & \textbf{100} & 19 & 20 & 6 & 19\\
\textbf{Datasets} & & \textbf{28} & 3 & 8 & 35 & 6\\
 \hline
\multirow{10}{*}{\textbf{Modalities}} & Ultrasound & \ding{51} & & & \ding{51} &   \\
 & Dermoscopy & \ding{51}& & \ding{51} & \ding{51} &  \\
 & Endoscopy & \ding{51}& & \ding{51} & \ding{51} &  \\
 & Fundus & \ding{51} & & \ding{51} & \ding{51} &  \\
 & X-Ray & \ding{51} & & & \ding{51} & \\
 & Histopathology & \ding{51} &  & & \ding{51} &  \\
 & CT & \ding{51} & \ding{51} & \ding{51} & \ding{51} & \ding{51} \\
 & MRI & \ding{51} &  &\ding{51} & \ding{51} & \ding{51}\\
 & Nuclei & \ding{51} & & & \ding{51} & \\
 & OCT & \ding{51} &  & & \ding{51} & \\
  \hline
\multirow{3}{*}{\textbf{Evaluation}} & Robustness & \ding{51} & \ding{51} & \ding{51} & & \ding{51} \\
 & Generalization & \ding{51} & \ding{51} & & & \\
 & Efficiency & \ding{51} & & & & \\
 \hline
 \multirow{5}{*}{\textbf{Architecture Analysis}} & CNN & \ding{51} & \ding{51} & \ding{51} & \ding{51} & \ding{51} \\
 & Transformer & \ding{51} & \ding{51} & \ding{51} & & \ding{51}\\
 & Hybrid & \ding{51} & \ding{51} & \ding{51} & \ding{51} & \ding{51}\\
 & Mamba & \ding{51} & & & & \ding{51}\\
 & RWKV & \ding{51} & \\
  \hline
\multirow{3}{*}{\textbf{Dataset Analysis}} & Scale & \ding{51} & & & \\
 & Boundary & \ding{51} & & &\\
 & Shape & \ding{51} & & &\\
\Xhline{1.5px}
\end{tabular}}

\end{table*}

To systematically and comprehensively evaluate U-shaped medical image segmentation models, we introduce \textbf{U-Bench}, the first large-scale, statistically rigorous, and efficiency-oriented benchmark for U-Net and its variants. U-Bench is built upon three key aspects: \textbf{(1) Broad dataset and model coverage:} we implement \textbf{100} recent U-Net variants and evaluate them on \textbf{28} benchmark datasets covering \textbf{10} diverse imaging modalities (ultrasound, dermoscopy, endoscopy, fundus photography, histopathology, nuclear imaging, X-ray, MRI, CT, and OCT; Fig.~\ref{fig:uteaser}A, C). \textbf{(2) Rigorous and comprehensive evaluation:} all models are implemented to calculate performance gains over the baseline U-Net with statistical significance, ensuring robust and fair comparisons (Fig.~\ref{fig:uteaser}E). To capture clinical utility, we further assess zero-shot generalization across modalities. Additionally, to address practical considerations in real-world edge deployment, we introduce the \underline{U-Score}, a statistically grounded, large-scale metric that jointly accounts for accuracy, parameter numbers, computational cost, and inference speed (Fig.~\ref{fig:uteaser}F). \textbf{(3) Public availability and reproducibility:} U-Bench implements models using official code implementations, pre-trained weights, and deep supervision strategies (if available). At the same time, U-Bench is released with all code, models, and protocols, enabling the community to reproduce our results and extend the benchmark with future methods.

Building on this large-scale evaluation, we identify key findings that challenge common assumptions. Traditional metrics like IoU show signs of saturation, offering a limited discriminative power (Fig.~\ref{fig:uteaser}F). Additionally, reported improvements are often inconsistent or statistically insignificant (Fig.~\ref{fig:uteaser}E). At the same time, an increasing focus on storage and computational cost is reflected in the rising trajectory of U-Score (Fig.~\ref{fig:uteaser}F). To explore these dynamics, we conduct a systematic analysis of U-Net variants, examining the influence of dataset and architectural factors on model performance across different modalities, architectures, and computational resource limitations (Fig.~\ref{fig:uteaser}G). Building on these analyses, we introduce a model advisor agent that suggests suitable architectures based on dataset and task attributes, turning an actionable guidance for practitioners in clinical and research contexts.

Our contribution can be summarized as:
\begin{itemize}
    \item We provide a comprehensive evaluation benchmark of 100 U-shaped variants across 28 datasets from 10 modalities with a rigorous assessment across statistical robustness, zero-shot generalization, and computational efficiency. To better capture the trade-off between accuracy and efficiency, we introduce U-Score, a novel metric grounded in large-scale statistical analysis that enables fair and holistic evaluation. 
    \item We summarize the observations over large-scale evaluation: Most variants show performance gains, but few show in-domain statistical significance over the original U-Net. Zero-shot performances show significant and promising improvements. U-Score shows an increasing trajectory, indicating the shift from purely pursuing accuracy to balancing accuracy with efficiency. 
    \item We disentangle different aspects, including dataset characteristics and architectural designs, revealing their impact on performance and efficiency, and further build a model recommender that helps researchers identify well-suited architectures under diverse data and resource conditions.
    \item We open-source U-Bench and all the pretrained weights, providing a large-scale benchmark with comprehensive evaluation for medical image segmentation, to foster fair, robust, generalizable, and efficient research in the community.
\end{itemize}
\newpage
\section{U-Bench Construction}
\label{sec:method}

\subsection{Preliminaries: u-shaped design}

\begin{wrapfigure}{r}{0.4\textwidth}
  \centering
  \vspace{-6mm}
  \includegraphics[width=0.4\textwidth]{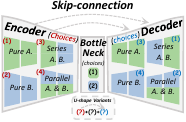}
  \caption{\textbf{Summary of U-shaped networks.} The network comprises an encoder, a bottleneck, and a decoder with skip-connection, each of which can integrate attention gates and multi-scale fusion.}
  \label{fig:combination}
\end{wrapfigure}
A U-shaped model generally comprises three components: hierarchical encoder, decoder, bottleneck, and skip-connection. Given an input image $x \in \mathbb{R}^{3\times H\times W}$, $\text{Encoder}(\cdot)$ extracts multi-scale features $f_{i}$ by $N$ stages from up-bottom, denoted as $\left \{ f_{i} \right \}_{i=1}^{N},f_{i}\in \mathbb{R}^{C_{i}\times \frac{H}{2^{(i-1)}} \times \frac{W}{2^{(i-1)}}}$. $\text{Bottleneck}(\cdot)$ processes the last output feature, and $\text{Decoder}(\cdot)$ is  composed of $N-1$ stages for upsampling decoder features $d_{j}$ from bottom-up, each stage comprises $\text{Skip-connection}(\cdot)$ for feature fusion. Final prediction $\tilde{x}$ is produced by the segmentation head after the top decoder stage.  The differences across variants are illustrated in Fig.~\ref{fig:combination}: Convolutional Neural Networks (CNNs) and related architectures (Attention, Mamba, RWKV) form the core building blocks, which can be organized in pure CNN / Attention, parallel, or sequential configurations for both encoding and decoding. Detailed categorization can be found in the Appendix~\ref{app:relate} and~\ref{app:modelzoo}.

\subsection{Dataset and Model Zoo}
\noindent\textbf{Dataset Zoo.} As shown in Fig.~\ref{fig:uteaser}(C), the U-Bench dataset zoo consists of 28 diverse 2D medical image segmentation datasets spanning a wide range of imaging modalities, including ultrasound, dermoscopy, endoscopy, fundus photography, histopathology, nuclear imaging, X-ray, MRI, CT, and OCT. We train on 20 datasets and evaluate zero-shot generalization on 8 additional ones. Following prior work~\citep{TransUNet,MobileUViT,UNeXt,RWKV-UNet,UCTransNet}, all datasets are resized to 256$\times$256 and augmented by random rotation and flipping; for models with fixed input size, we keep their original resolution (typically 224$\times$224). Official splits are used when available; otherwise, a 7/3 split is applied. All details on datasets and preprocessing are provided in the Appendix~\ref{app:datazoo}.

\noindent\textbf{Model Zoo.} We curate a collection of 100 publicly available and widely adopted U-Net variants, covering CNN-, Transformer-, Mamba-, and RWKV-based architectures, as well as their hybrid designs (Fig.~\ref{fig:uteaser}(B)). To ensure strict reproducibility and fair comparison, we follow the official implementations for all models, adopting their predefined settings, pretrained weights, and deep supervision strategies when available. All model details are provided in the Appendix~\ref{app:modelzoo}.

\subsection{Evaluation Protocol}
\label{sec:u_score}
\noindent\textbf{Evaluation Metrics.} Following previous works~\citep{UTANet,RWKV-UNet,MobileUViT,UNeXt,CMUNeXt}, we evaluate segmentation performance using Intersection over Union (IoU). To evaluate the statistical significance of performance differences between models, we conduct paired sample $t$-tests, comparing each variant to the baseline U-Net. U-Bench also considers computational efficiency metrics, including Parameters (M), FLOPs (G), and FPS. All result details are provided in the Appendix~\ref{app:modelzoo}.

\noindent\textbf{Zero-shot Evaluation.} To evaluate the generalization capability of each model beyond the training distribution, we conduct zero-shot inference on unseen datasets within the same modality and task. Specifically, models are trained on source datasets and then directly evaluated on unseen datasets that share the same modality but differ in acquisition domain. Detailed dataset split can be found in Fig.~\ref{fig:uteaser}(C). This approach aligns with clinical demands, where domain shifts frequently occur in real-world applications due to variations in devices, institutions, and patient populations.

\begin{figure} 
  \centering
  \includegraphics[width=1\textwidth]{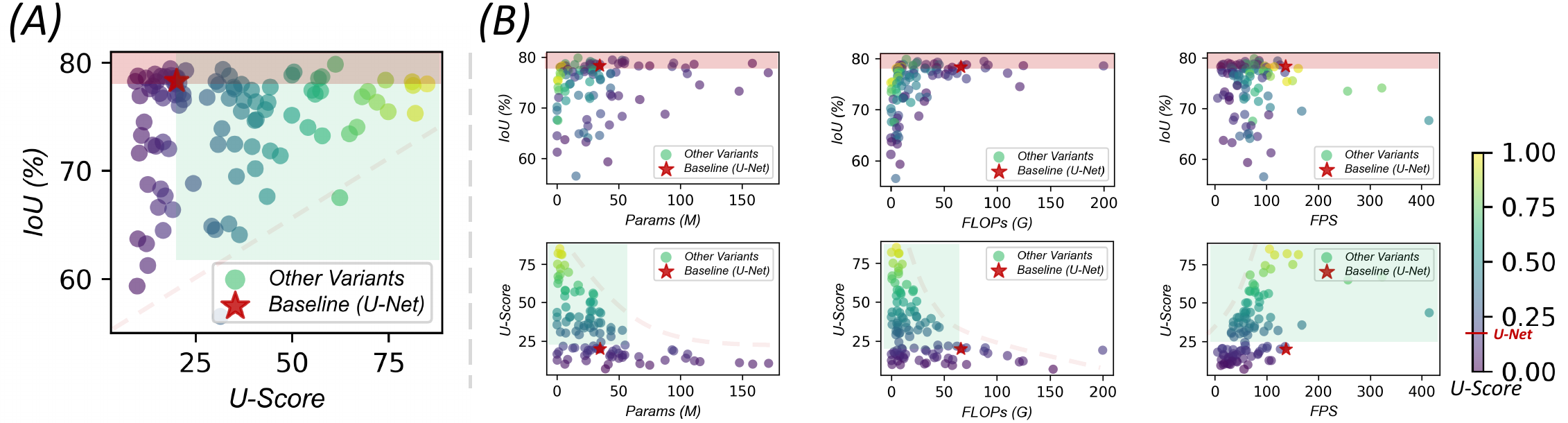}
  \caption{\textbf{Comparison between IoU and U-Score.} Red rectangle indicates the models perform better than U-Net in IoU, and green rectangle indicates the models perform better than U-Net in U-Score. (A) Across 100 variants, few methods show better IoU compared to baseline U-Net, while more than half of the methods show better U-Score. (B) The relationship between performance (IoU) and the increase in computational resources (FLOPs, parameters, FPS) is complex, whereas U-Score offers a clear distribution that effectively distinguishes favorable and unfavorable accuracy-efficiency trade-off.}
  \label{fig:uscore}
\end{figure}

\noindent\textbf{U-Score.} To assess real-world deployability, we propose U-Score, a unified metric that jointly accounts for accuracy and efficiency. For each model $i$, we compute segmentation accuracy evaluated by IoU $A_i$ parameter $P_i$, FLOPs $G_i$, and FPS $S_i$, which are percentile-normalized into $a_i, p_i, g_i, s_i \in [0,1]$ using the 10-th/90-th quantiles across the model zoo. Given a harmonic mean function $\mathcal{H}(\cdot)$ with equal weights for each input, an efficiency subscore $\text{Eff}_i=\mathcal{H}(p_i,g_i,s_i)$ is obtained, ensuring that no single factor dominates. The final U-Score is defined as $\text{U-Score}_i=\mathcal{H}(a_i,\text{Eff}_i)$ to incorporate segmentation accuracy and computational efficiency equally.
This combination rewards models that achieve favorable accuracy-efficiency trade-offs and provides a deployment-oriented alternative to IoU. More details can be found in Appendix~\ref{app:uscore}. 

We report the IoU and U-Score for all models with parameter count, FLOPs, and FPS, as shown in Fig.~\ref{fig:uscore} and Fig~\ref{fig:uteaser}(F). Models with lower computational costs show wide variation in IoU performance, while heavier models tend to achieve higher accuracy at the expense of greater resource consumption. The U-Net baseline, which falls in the middle in terms of computational demand, delivers reasonably strong performance. Although some models surpass U-Net in segmentation accuracy, they require substantially different levels of computational overhead, making direct comparisons with the baseline difficult. Using U-Score, however, reveals that U-Net has a weak accuracy-efficiency trade-off, whereas other models show more distinct and discriminative results, allowing clearer separation between approaches with favorable and unfavorable trade-offs. Notably, the IoU gains of advanced models over U-Net are marginal, suggesting a saturation point, while the large gap between the Top-1 model’s U-Score and U-Net’s highlights that IoU alone is no longer the key bottleneck in medical segmentation tasks. These findings underscore that efficiency is becoming an increasingly critical factor in model development and practical deployment.

\vspace{-2mm}
\section{U-Bench Results \& Discussion}
In this section, we present the results of the U-Bench benchmark across multiple dimensions, including accuracy, efficiency, and generalization. We organize the results as follows: In \ref{sec:results}, we 
present and discuss retrospective analysis of the develop trends and statistical findings over 100 variants spanning different architectures and publication years. In \ref{sec:analysis}, we disentangle influence factors into two aspects: dataset and architecture, and analyze how these factors impact model performance. In \ref{sec:recommend}, we propose our ranking-based advisor agent, offering practical guidance for selecting optimal models based on dataset characteristics and resource constraints.

% \subsection{Progress for Most, Stagnation for Some: Lessons from a Decade of Segmentation}
\subsection{Retrospective Analysis of the Past Decade}
\label{sec:results}

\begin{figure*}[t!]
  \centering
   \includegraphics[width=0.98\linewidth]{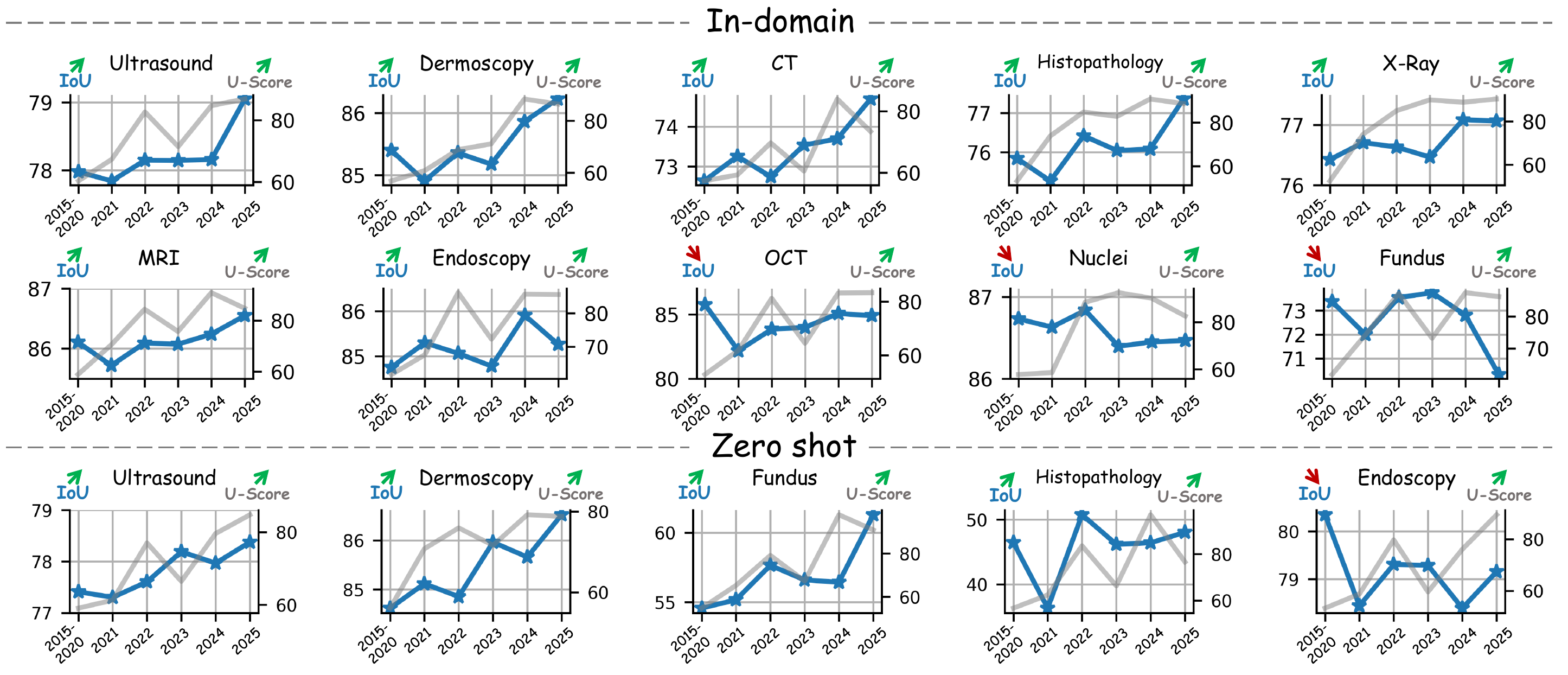}
   \vspace{-2mm}
   \caption{\textbf{Performance trends of SOTA models over the past decade.} The x-axis indicates publication year, with each point marking the yearly best result. The y-axes report two evaluation metrics: \textcolor{myblue}{\textbf{IoU}} (left axis) and \textcolor{gray}{\textbf{U-Score}} (right axis). The trend's summary is shown as arrows at the top of the y-axis, with \textcolor{green}{green} ones highlighting improvements and \textcolor{red}{red} ones indicating stagnation. Source domain performance is show at the top, and zero-shot performance is shown at the bottom.}
   \label{fig:trend}
\end{figure*}

\noindent{\textbf{Finding 1: In-domain Top-1 performance has marginal gains in segmentation accuracy, while zero-shot improves more pronouncedly. }} 
We analyzed 100 variants across different architectures and publication years (the detailed list can be found in the Appendix~\ref{app:relate} of Fig.~\ref{fig:model}), reporting the best-performing variant for each year, as shown in Fig.~\ref{fig:trend}. Over the past decade, 70\% of modalities have demonstrated steady progress of segmentation accuracy in both source and target domains, as reflected in IoU. However, IoU gains have been marginal (on average 1\%-2\%) and inconsistent. Some modalities  (\textit{i.e.} OCT, Nuclei, and Fundus) even show a sign of stagnation. In comparison, when considering zero-shot performances, the improvements have been more obvious (more than 3\% on average) in 80\% of the modalities.

\begin{figure*}[t!]
  \centering
   \includegraphics[width=0.98\linewidth]{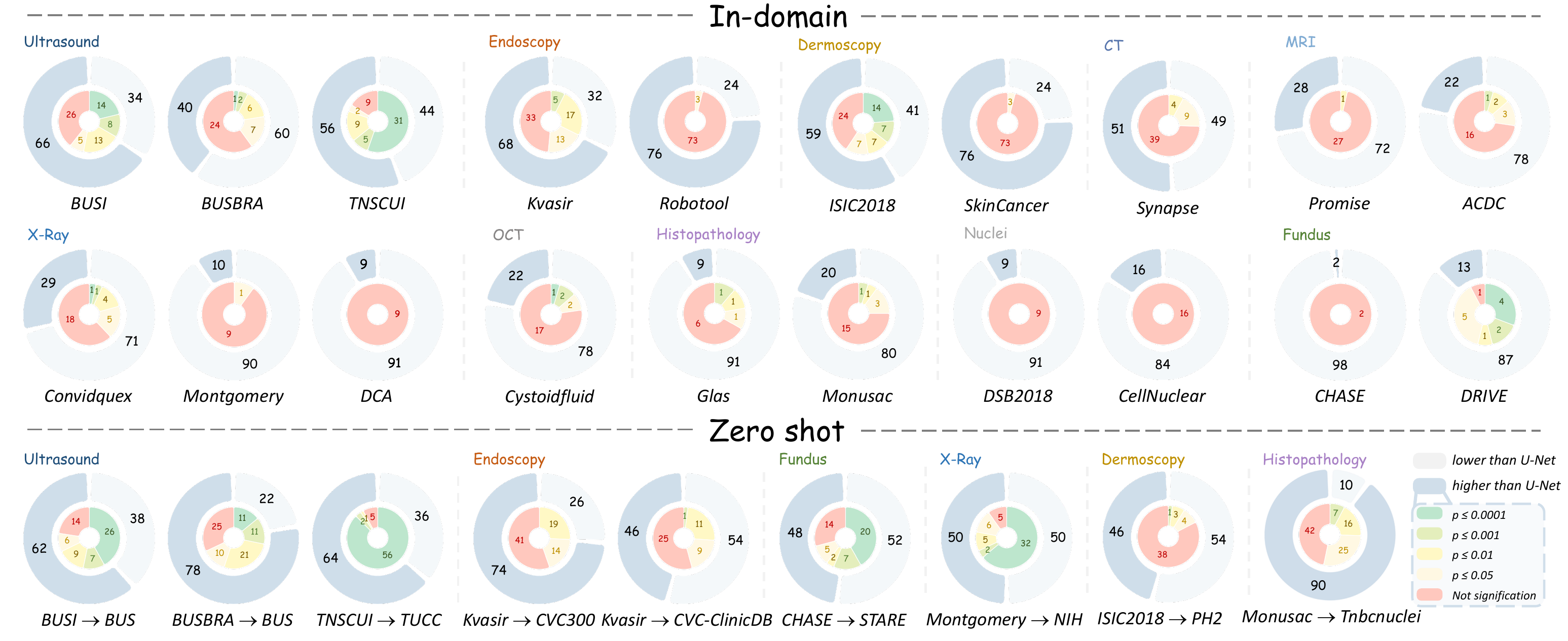}
   \vspace{-2mm}
   \caption{\textbf{Statistical significance analysis against U-Net across 28 datasets across 10 modalities.} The outer blue pie represents the number of variants surpassing U-Net; the inner pie quantifies the statistical significance of the methods with improvements, annotated by \setlength{\fboxsep}{0pt}\colorbox{r}{non-significant} to \setlength{\fboxsep}{0pt}\colorbox{g}{highly significant}, with the number of works annotated in the middle. In general, in-domain improvements show limited statistical significance, while zero-shot performances show more significant improvements.}
  \vspace{-2mm}
   \label{fig:performance}
\end{figure*}

\noindent{\textbf{Finding 2: Although some in-domain improvements exist on average, few reach statistical significance, whereas the average zero-shot improvements remain consistently significant.}} 
To rigorously distinguish modalities with genuine improvements from those with only numerical fluctuations, we perform $t$-tests between each variant and the U-Net baseline. The results are presented in Fig.~\ref{fig:uteaser}(E) and Fig.~\ref{fig:performance}. 
% Surprisingly, while most methods achieve better average results compared to U-Net, we have 
We observe that over 80\% of variants fail to achieve statistically significant improvements. Even in the most heavily studied modalities, such as Ultrasound, Endoscopy, Dermoscopy, CT, and MRI, most gains are marginal and lack significance. Only a handful of datasets (e.g., BUSI, TNSCUI, Kvasir, ISIC2018, Convidquex) exhibit consistent clusters of superior variants. %typically associated with tasks involving localized lesions.
In contrast, in experiments with zero-shot transfer, for variants that outperform U-Net, more than 50\% of the variants are significant across 75\% of the modalities.
% Similarly, in the zero-shot transfer setting, only a small number of variants perform significantly better than U-Net in most target domains.
% In contrast, under zero-shot transfer settings, more than 60\% of recent variants significantly outperform U-Net across multiple target domains.

\noindent{\textbf{Possible explanations for findings 1 \& 2: }} 
We provide a possible explanation for these interesting observations. Considering in-domain evaluations, the improvements with statistical significance is typically associated with the lesion localization tasks which requires global semantic comparison. Specifically, lesions often exhibit significant differences from surrounding normal tissues, requiring a global context to model these distinctions~\cite{zhou2017deep,nnunet}. In recent years, with the growing adoption of long-range modeling techniques (e.g., attention-based Transformers, state-space models such as Mamba, and RNN-inspired hybrids like RWKV), architectural innovations have increasingly focused on capturing long-range dependencies, leading to more pronounced and steady improvements in these lesion segmentation tasks. On the other hand, long-range modeling techniques have been proven to be more generalizable~\citep{generalization1,generalization2,generalization3,generalization4}, leading to improvements in zero-shot generalization ability. By contrast, modalities dominated by repetitive local patterns (e.g., Nuclei, Fundus) benefit far less from global modeling, exhibiting only marginal improvements. This underscores the complementary need for localized mechanisms to achieve precise boundary delineation.

\noindent{\textbf{Finding 3: Increasing attention on efficiency.}} 
While IoU shows marginal improvements, U-Score improvements are more pronounced, with an average increase of 33\%. This trend supports our argument in Sec.~\ref{sec:u_score}, where we suggest that IoU has reached a saturation point and has limited ability to discriminate favorable methods from unfavorable ones, indicating that accuracy alone is no longer the bottleneck of segmentation. The increasing trend of U-Score reflects the growing emphasis and room for improvement on efficient models in the medical community, echoing the practical demand for clinical deployment beyond the lab research.

\renewcommand{\multirowsetup}{\centering} 
\begin{table}[t!]
  \caption{\textbf{Top-10 variants ranked by performance (IoU) and efficiency (U-Score) under in-domain and zero-shot settings. }Variants cover \textcolor{cnn}{CNN}, \textcolor{transformer}{Transformer}, \textcolor{mamba}{Mamba}, \textcolor{rwkv}{RWKV}, and \textcolor{hybrid}{Hybrid} architectures.}
  \centering
  \label{tab:rank5}
  \noindent
\begin{minipage}[t]{0.251\textwidth}
    \centering
    \resizebox{\linewidth}{!}{
        \begin{tabular}{r  c r}
            \Xhline{1px}
            \multicolumn{3}{c}{Source (IoU)} \\
            \hline
            \multirow{1}{*}{Rank} & \multicolumn{1}{c}{Variants} & \multicolumn{1}{c}{\# Volume (year)}  \\
            \hline
            \gold& \textcolor{rwkv}{RWKV-UNet} & Arxiv (2025)  \\
            \silver& \textcolor{hybrid}{AURA-Net} & ISBI (2021) \\
            \bronze& \textcolor{cnn}{UTANet} & AAAI (2025) \\
            \#4 & \textcolor{cnn}{MEGANet} & WACV (2024) \\
            \#5 & \textcolor{mamba}{Swin-umamba} & MICCAI (2024) \\
            \#6 & \textcolor{cnn}{MFMSNet} & CIBM (2024) \\
            \#7 & \textcolor{hybrid}{TransResUNet} & Arxiv (2022) \\
            \#8 & \textcolor{hybrid}{EViT-UNet} & ISBI (2025)  \\
            \#9 & \textcolor{hybrid}{FCBFormer} & MIUA (2022) \\
            \#10 & \textcolor{hybrid}{DA-TransUNet} & FBB (2024) \\
            \rowcolor{gray!10} \#25 & \textcolor{cnn}{U-Net} & MICCAI (2025) \\
            \Xhline{1px}
        \end{tabular}
    }
\end{minipage}%
\hfill
\begin{minipage}[t]{0.251\textwidth}
    \centering
    \resizebox{\linewidth}{!}{
        \begin{tabular}{r  c r}
            \Xhline{1px}
            \multicolumn{3}{c}{Target (IoU)} \\
            \hline
            \multirow{1}{*}{Rank} & \multicolumn{1}{c}{Variants} & \multicolumn{1}{c}{\# Volume (year)} \\
            \hline
            \gold& \textcolor{rwkv}{RWKV-UNet} & Arxiv (2025) \\
            \silver& \textcolor{hybrid}{G-CASCADE} & WACV (2024) \\
            \bronze& \textcolor{mamba}{Swin-umamba} & MICCAI (2024) \\
            \#4 & \textcolor{cnn}{MEGANet} & WACV (2024) \\
            \#5 & \textcolor{hybrid}{CASCADE} & WACV (2023) \\
            \#6 & \textcolor{cnn}{MFMSNet} & CIBM (2024) \\
            \#7 & \textcolor{hybrid}{TransResUNet} & Arxiv (2022) \\
            \#8 & \textcolor{hybrid}{DS-TransUNet} & TIM (2022) \\
            \#9 & \textcolor{hybrid}{CENet} & MICCAI (2025) \\
            \#10 & \textcolor{cnn}{PraNet} & MICCAI (2020) \\
            \rowcolor{gray!10} \#72 & \textcolor{cnn}{U-Net} & MICCAI (2025) \\
            \Xhline{1px}
        \end{tabular}
    }
\end{minipage}%
\hfill
\begin{minipage}[t]{0.245\textwidth}
    \centering
    \resizebox{\linewidth}{!}{
        \begin{tabular}{r  c r}
            \Xhline{1px}
            \multicolumn{3}{c}{Source (U-Score)} \\
            \hline
            \multirow{1}{*}{Rank} & \multicolumn{1}{c}{Variants} & \multicolumn{1}{c}{\# Volume (year)} \\
            \hline
            \gold & \textcolor{hybrid}{LGMSNet} & ECAI (2025) \\
            \silver& \textcolor{cnn}{MBSNet} & MSSP (2021)  \\
            \bronze& \textcolor{cnn}{CMUNeXt} & ISBI (2024) \\
            \#4 & \textcolor{cnn}{LV-UNet} & BIBM (2024) \\
            \#5 & \textcolor{hybrid}{Mobile U-ViT} & ACM MM (2025) \\
            \#6 & \textcolor{cnn}{Tinyunet} & MICCAI (2024) \\
            \#7 & \textcolor{rwkv}{U-RWKV} & MICCAI (2025) \\
            \#8 & \textcolor{cnn}{U-KAN} & AAAI (2025)\\
            \#9 & \textcolor{cnn}{DCSAU-Net} & CIBM (2023) \\
            \#10 & \textcolor{rwkv}{RWKV-UNet} & Arxiv (2025) \\
            \rowcolor{gray!10} \#64 & \textcolor{cnn}{U-Net} & MICCAI (2025) \\
            \Xhline{1px}
        \end{tabular}
    }
\end{minipage}
\hfill
\begin{minipage}[t]{0.245\textwidth}
    \centering
    \resizebox{\linewidth}{!}{
        \begin{tabular}{r  c r}
            \Xhline{1px}
            \multicolumn{3}{c}{Target (U-Score)} \\
            \hline
            \multirow{1}{*}{Rank} & \multicolumn{1}{c}{Variants} & \multicolumn{1}{c}{\# Volume (year)} \\
            \hline
            \gold& \textcolor{hybrid}{LGMSNet} & ECAI (2025) \\
            \silver& \textcolor{cnn}{LV-UNet} & BIBM (2024) \\
            \bronze& \textcolor{cnn}{U-KAN} & AAAI (2025) \\
            \#4 & \textcolor{hybrid}{Mobile U-ViT} & ACM MM (2025) \\
            \#5 & \textcolor{rwkv}{RWKV-UNet} & Arxiv (2025) \\
            \#6 & \textcolor{cnn}{MBSNet} & MSSP (2021) \\
            \#7 & \textcolor{hybrid}{SwinUNETR} & CVPR (2022) \\
            \#8 & \textcolor{cnn}{CMUNeXt} & ISBI (2024) \\
            \#9 & \textcolor{hybrid}{G-CASCADE} & WACV (2024) \\
            \#10 & \textcolor{cnn}{TA-Net} & CIBM (2021) \\
            \rowcolor{gray!10} \#69 & \textcolor{cnn}{U-Net} & MICCAI (2025) \\
            \Xhline{1px}
        \end{tabular}
    }
\end{minipage}%
\end{table}

\begin{figure*}[t!]
  \centering
   \vspace{-2mm}
   \includegraphics[width=0.98\linewidth]{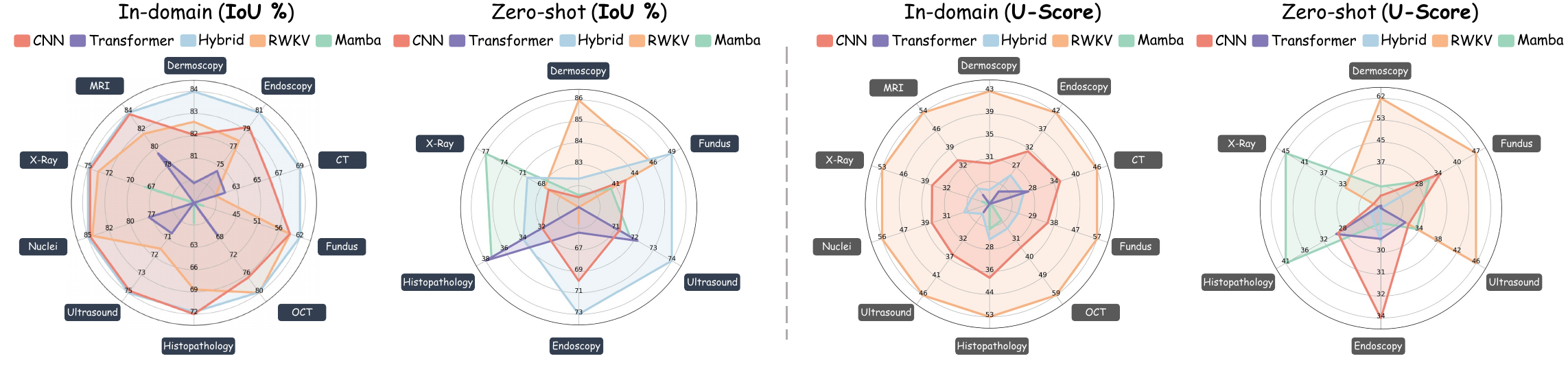}
   \caption{\textbf{Average performance of different architectures across modalities under in-domain and zero-shot settings. }Left: IoU-based comparison; Right: U-Score-based comparison of different architecture strengths in source-domain and zero-shot.}
   \label{fig:radir}
\end{figure*}

\subsection{Influencing Factor Analysis: Architectures and Data Characteristics}
\label{sec:analysis}

\subsubsection{Architectures}
To analyze the performance regarding architectural choices, we divide 100 models into five families: \textcolor{cnn}{CNN}, \textcolor{transformer}{Transformer}, \textcolor{mamba}{Mamba}, \textcolor{rwkv}{RWKV}, and \textcolor{hybrid}{Hybrid}. The detailed descriptions are summarized in Appendix~\ref{app:modelzoo}. We present the top-10 variants across all datasets ranked by IoU and U-Score under in-domain and zero-shot settings, as shown in Tab.~\ref{tab:rank5}, and we calculate the average performance of each architecture family, as shown in Fig.~\ref{fig:radir}.

Considering segmentation performance (IoU), \textcolor{hybrid}{Hybrid} architectures achieve the highest accuracy by combining local priors with global attention. As shown in Tab.~\ref{tab:rank5}(Left), 5 of the top 10 models in both in-domain and zero-shot are hybrid, highlighting their high potential. On average, the hybrid family consistently delivers the best in-domain performance and competitive zero-shot generalization (Fig.~\ref{fig:radir}(Left)), particularly excelling on lesion-centric tasks such as Ultrasound and Endoscopy. The newly proposed \textcolor{rwkv}{RWKV} family ranks first in IoU for both in-domain and zero-shot evaluations, indicating promising potential despite limited prior research. In contrast, \textcolor{mamba}{Mamba} family shows weaker segmentation performance, which may be attributed to its architectural design, which, despite its strengths in certain tasks, might struggle with capturing fine-grained details or handling complex patterns in segmentation tasks.

Once computational demands are taken into account, as shown in Tab.~\ref{tab:rank5}(Right), the U-Score-based leaderboard is reshuffled, with the \textcolor{cnn}{CNN} family leading in performance, comprising 7 / 5 out of the top 10 models in in-domain / zero-shot settings, respectively. The newly proposed \textcolor{rwkv}{RWKV} family achieves the best average in-domain results and competitive zero-shot performance (Fig.~\ref{fig:radir}(Right)), further supporting its structural superiority and potential. In contrast, inefficient long-range modeling methods, including \textcolor{transformer}{Transformer}, and \textcolor{hybrid}{Hybrid} architectures, face higher computational demands, leading to reduced performance when evaluated by U-Score. Although \textcolor{mamba}{Mamba} excels in efficiency, its inconsistent accuracy undermines the U-Score, offsetting its efficiency advantage.

\subsubsection{Data Characteristics}

\begin{figure}[t]
  \centering
  \includegraphics[width=1\textwidth]{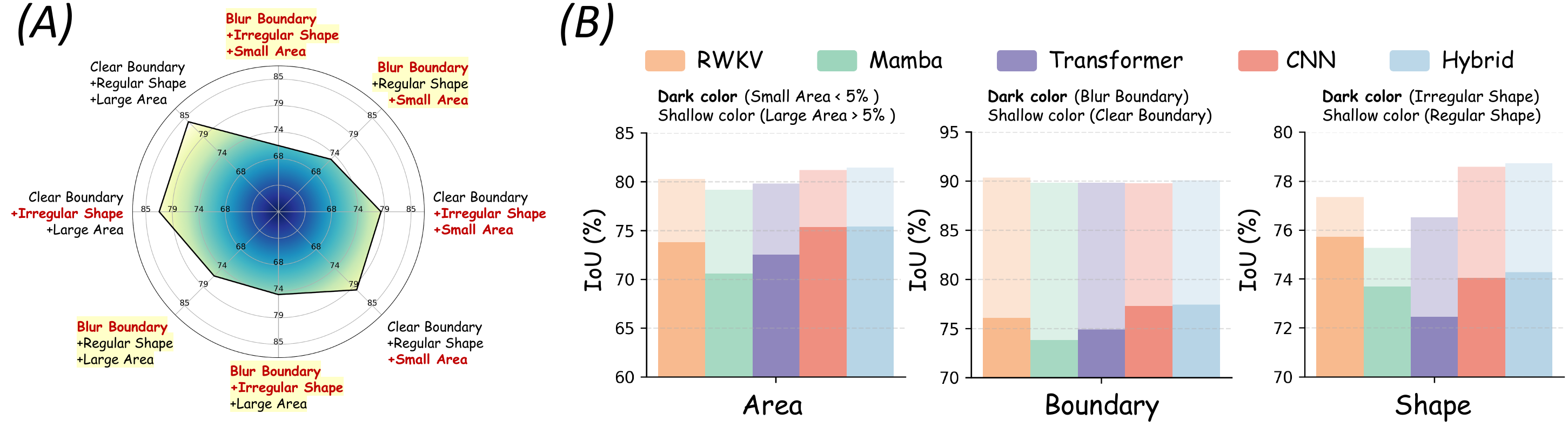}
  \caption{\textbf{Performance analysis under varying foreground properties.} (A) Foreground properties influence segmentation task difficulty. The yellow background indicates the challenge segmentation case. (B) Architectural influence on segmentation difficulty across diverse foreground properties. Dark / Shallow: hard / easy case.}
  \vspace{-4mm}
  \label{fig:abs}
\end{figure}
We further investigate how performances vary with distinct foreground characteristics with three aspects: foreground scale, boundary sharpness, and shape complexity. The Appendix~\ref{app:characterization} provides detailed definitions for the different scales of target area, edge, and shape regularity.

Figure~\ref{fig:abs}(A) summarizes the characteristics of challenging cases: blurry boundaries are the dominant factor, with often causing substantial drops in segmentation performance, while small object size and irregular shapes further exacerbate the difficulty. When these foreground properties shift across datasets, different models exhibit varying performance patterns. As shown in Fig.~\ref{fig:abs}(B), consistent with our earlier findings, hybrid architectures dominate both in easier and more challenging cases, proving that local and global fusion mechanism enables greater adaptability across diverse foreground properties, particularly for blurry boundaries. RWKV-based models show specific strength in capturing irregular but well-defined shapes, reflecting their ability to model long-range contours. Nonetheless, boundary ambiguity, along with small and irregular targets, remains the central challenge; given its prevalence in medical images, uncertainty-aware designs are needed. Since architectural strengths are dataset-dependent, these observations highlight the importance of task-aware advising mechanisms that can match models to dataset properties.

\subsection{Model Advisor Agent}
\label{sec:recommend}

\begin{wrapfigure}{r}{0.45\textwidth}
  \vspace{-4mm}
  \centering
  \includegraphics[width=\linewidth]{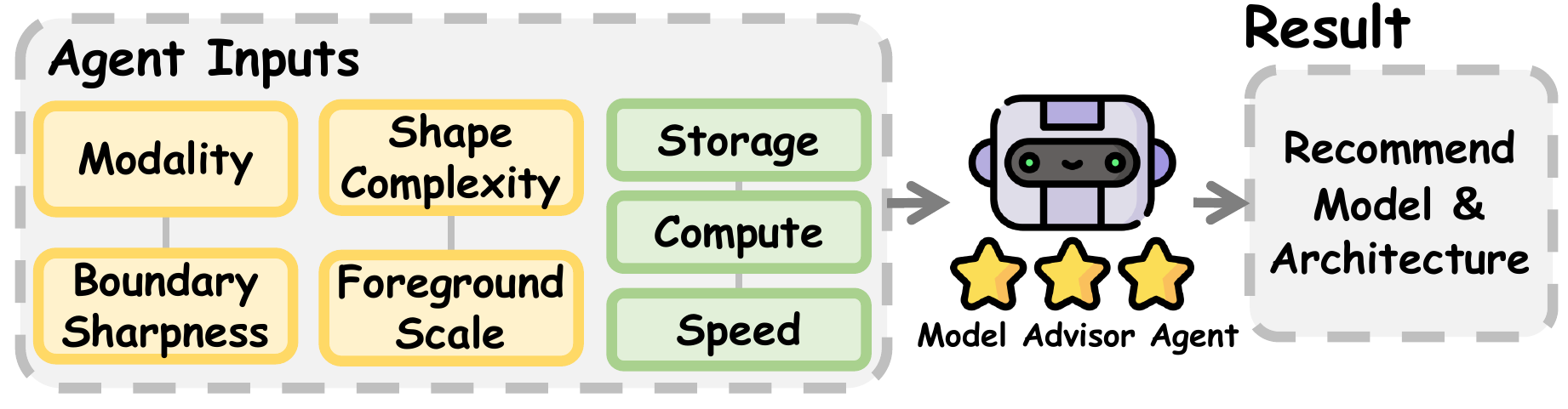}
  \caption{Our model advisor agent. \label{fig:recommend_wrap}}
\end{wrapfigure} 
Based on our analysis, we introduce a ranking-based model advisor agent, designed to guide the community in selecting the most suitable models based on dataset characteristics and task requirements. This tool not only streamlines model selection but also helps users navigate the trade-offs between performance and efficiency, ensuring more informed, task-aware decisions.

The system overview is shown in Fig.~\ref{fig:recommend_wrap}. Our advisor agent system utilizes dataset-level characteristics (e.g., modality, boundary sharpness, shape complexity, and foreground scale) along with resource constraints (storage, computation, and speed) to predict the suitability of various U-shaped architectures. 
Rather than relying on a manual trial-and-error approach, our framework leverages XGBoost~\citep{xgboost} as the recommended backbone and outputs candidate models and architectures that best satisfy the specified requirements.
Crucially, the output is not a single ``best'' model but a prioritized list, offering more flexibility in choices to practitioners. 
%This helps them narrow the search and focus on architectures most likely to perform well under their specific conditions.
Further details on the recommendation setup, dataset construction, implementation details, and evaluation metrics are provided in the Appendix~\ref{app:agent} and ~\ref{app:f-metrics}.

\begin{wraptable}{r}{0.45\textwidth}
  \centering
  \vspace{-4mm}
  \caption{NDCG, MAP, and Spearman correlation of our advisor agent. 
  \label{tab:recommend_wrap}}
  \resizebox{\linewidth}{!}{
    \begin{tabular}{lcccc}
      \Xhline{1.5pt}
      \multirow{2}{*}{Ranking Metric} & \multicolumn{2}{c}{NDCG} & \multirow{2}{*}{MAP} & \multirow{2}{*}{Spearman} \\
      \cline{2-3}
       & $@5$ & $@20$ & & \\
      \hline
      IoU     & 0.75 & 0.76 & 0.24 & 0.36 \\
      U-Score & 0.74 & 0.79 & 0.43 & 0.52 \\
      \Xhline{1.5pt}
    \end{tabular}
  }
  \vspace{-4mm}
\end{wraptable}

We design a set of experiments to validate the feasibility of automatic model suggestion in medical image segmentation. Our setup uses 18 in-domain datasets for training and holds out 2 datasets for validation. We use Normalized Discounted Cumulative Gain (NDCG), mean average precision (MAP) and Spearman correlation for evaluation (See Appendix~\ref{app:f-metrics}). As shown in Tab.~\ref{tab:recommend_wrap}, our experiments demonstrate that the proposed model advisor agent effectively recovers ranking orders that align with ground-truth IoU and U-Score rank in our benchmark. The results validate that our advisor agent system is able to prioritize suitable models across different task requirements, making it a reliable tool for model selection and deployment.
% \clearpage
\section{Conclusion}
\label{sec:conclusion}

A key challenge in the field of medical image segmentation remains: \textit{How can we conduct a fair and comprehensive comparison across the numerous U-shaped variants} ? To address this, we introduce U-Bench, a framework that fills critical gaps in prior evaluations by offering a comprehensive, statistically rigorous, and efficiency-oriented approach. Our results challenge common assumptions in the field, revealing that while many variants show performance gains, few achieve statistical significance in-domain. In contrast, zero-shot generalization demonstrates substantial improvements, highlighting the potential for better model generalization across domains. In addition, the newly proposed U-Score metric, which emphasizes efficiency alongside performance, signals a paradigm shift from models focused solely on accuracy to those that balance both performance and  efficiency. Leveraging insights from our analysis of model architecture and dataset characteristics, we propose a ranking-based model agent that transforms our large-scale evaluation into actionable guidance for selecting models tailored to specific tasks. By releasing U-Bench as an open-source platform, we provide the community with a robust, reproducible tool to advance research in segmentation, enabling the development of models that are both accurate and feasible for clinical deployment.

\section{Acknowledgment}

Supported by Natural Science Foundation of China under Grant 62271465, National Key R\&D Program of China under Grant 2025YFC3408300, and Suzhou Basic Research Program under Grant SYG202338.

\textbf{Fenghe Tang}: Writing \- review \& editing, Methodology, Conceptualization, Visualization, Formal analysis, Validation, Data curation, Writing \- original draft, Investigation. \textbf{Chengqi Dong}: Writing \- review \& editing, Methodology, Visualization, Data curation, Writing \- original draft. \textbf{Wenxin Ma}: Writing \- review \& editing, Conceptualization, Formal analysis, Writing \- original draft. \textbf{Zikang Xu}: Writing \- review \& editing, Formal analysis. \textbf{Heqin Zhu, Zihang Jiang, Rengsheng Wang, Yuhao Wang, and Chenxu Wu}: Writing \- review \& editing. \textbf{Shaohua Kevin Zhou}: Supervision,  Writing \- review \& editing, Funding acquisition.

\clearpage

\appendix
\section{Appendix}

In this appendix, we provide additional details and results to complement the main paper. 
The content is organized as follows:

\noindent\textbf{Appendix~\ref{app:relate}}: Relate Work.

\noindent\textbf{Appendix~\ref{app:datazoo}}: Details of U-Bench Data Zoo.

\noindent\textbf{Appendix~\ref{app:modelzoo}}: Details of U-Bench Model Zoo.

\noindent\textbf{Appendix~\ref{app:uscore}}: Details of U-Score.

\noindent\textbf{Appendix~\ref{app:implementation}}: Implementation and Evaluation Details.

\noindent\textbf{Appendix~\ref{app:result}}: Additional Results.

\noindent\textbf{Appendix~\ref{app:reproduce}}: Reproducibility Checklist.

% \noindent\textbf{Appendix~\ref{app:llmuse}}:Explanation of the Use of LLM.

\clearpage
\section{Related Work}\label{app:relate}

In Appendix~\ref{app:relate}, we present a broad view of the variations of U-shape networks, including the architecture of the network and existing medical segmentation benchmarks.

\subsection{Model Architecture}

\begin{figure}[htbp]
    \centering
    \includegraphics[width=\linewidth]{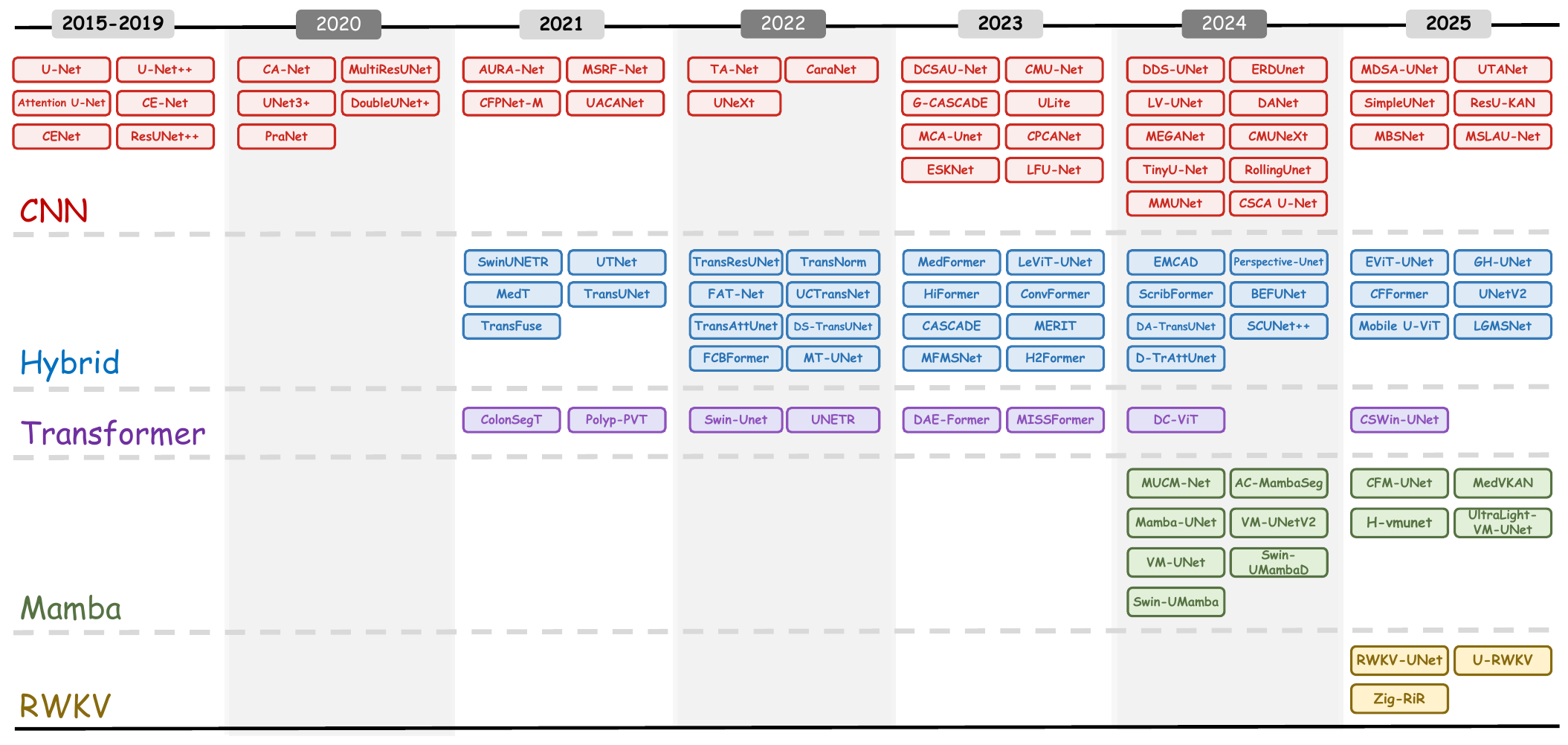}  
    \caption{Time and architecture distribution of all evaluated models.}  
    \label{fig:model}
\end{figure}

As the core architecture for medical image segmentation, the U-Net has evolved into numerous variants in recent years, driven by advancements in feature representation capabilities, long-range dependency modeling techniques, and the trade-off between efficiency and accuracy. This section categorizes and organizes these U-Net variants based on their core paradigms and design motivations, systematically tracing their evolutionary path from foundational construction to integrated innovation. Fig.~\ref{fig:model} summarizes the evolution of U-Net variants over time.

\begin{enumerate}
\item \textbf{U-shaped Networks Dominated by Convolutional Neural Networks (CNNs)} (2015-2021: Foundational Laying)\par

U-shaped networks, using convolutional neural networks as their sole backbone, extract local features through convolution operations and utilize fixed skip connections to fuse multi-scale information, laying the foundation for the "encoder-decoder" paradigm in medical image segmentation. Their advantages lie in their ability to accurately capture local details (such as textures and edges), their relatively lightweight architecture, and their stable training process, providing a solid design baseline for subsequent complex variants. However, the local receptive field of convolutional operations limits the ability to model global semantic information, and fixed skip connections easily lead to a semantic gap between the encoder and decoder, limiting performance.

\item \textbf{Transformer-driven U-Networks} (2021-2023: Paradigm Shift)\par

This variant introduces the Transformer architecture (including variants such as Vision Transformer~\citep{vit} and Swin Transformer~\citep{swin}) to replace or enhance the traditional CNN backbone, leveraging the self-attention mechanism to effectively model long-range dependencies. However, the computational complexity of the self-attention mechanism grows quadratically with sequence length, resulting in low inference efficiency. Furthermore, this type of model is poorly adaptable to small-scale medical datasets and is prone to overfitting due to insufficient data, making it difficult to meet the stringent real-time requirements of clinical edge devices.

\item \textbf{U-Networks Based on State-Space Models (SSMs) and Recurrent Paradigms} (2023-2025: Efficiency-Oriented)\par

Recent research explores replacing the quadratic-cost self-attention with linear-time alternatives. One line leverages state-space models (e.g., Mamba~\citep{mamba}) that adopt selective state updates to capture long-range dependencies while achieving linear complexity, markedly improving inference efficiency and adaptability to small sample sizes. Another complementary line introduces RWKV~\citep{rwkv}, a recurrent-inspired model that combines Transformer-like expressiveness with RNN-style recurrence, enabling efficient sequential processing and stronger generalization across varying input lengths. Together, these paradigms alleviate the computational and data-dependency limitations of Transformers.

\item \textbf{Multi-Paradigm Fusion U-Networks (Hybrid Networks)} (2020-2025: Fusion and Innovation)\par

This phase aims to integrate the advantages of CNNs in local feature extraction, the global semantic modeling capabilities of Transformers. The goal is to achieve a balance between accuracy, efficiency, and generalization by fusing different architectures. This type of network variant can adapt to complex clinical scenarios such as multimodal imaging and cross-center data heterogeneity, significantly improving the practical value of segmentation results. However, the architectural design complexity increases significantly, and the coordination mechanisms between modules of different paradigms (such as the timing of feature interactions and weight distribution) still need further optimization.
\end{enumerate}

The development of the four types of U-shaped network variants follows a technological evolutionary path of "local refinement $\to$ global correlation $\to$ efficiency considerations $\to$ multi-paradigm collaboration," reflecting the shift in clinical needs from static, single-scenario segmentation toward more efficient, generalized solutions adaptable across diverse conditions.

\subsection{Medical Segmentation Benchmarks}
To fill the research gap in the evaluation of U-net systems, we comprehensively compare previous segmentation evaluation benchmarks with the U-Bench proposed in this paper, thereby clarifying the innovative positioning of U-Bench.

\subsubsection{Related Work}
Medical image segmentation has seen rapid progress, driven by deep learning architectures and large-scale datasets. However, the validity and reproducibility of many reported advances have been challenged due to inconsistent evaluation protocols, limited dataset diversity, and insufficient consideration of deployment constraints.

TorchStone \citep{torchstone} addressed some of these limitations by introducing a large-scale collaborative benchmark for abdominal organ segmentation, leveraging diverse CT scans from multiple hospitals worldwide. While it emphasized out-of-distribution generalization and standardized evaluation, its scope was limited to a single anatomical region and modality, making it less suitable for assessing broader architectural capabilities.
MedSegBench \citep{nnwnet} expanded coverage across modalities, incorporating 35 datasets from ultrasound, MRI, X-ray, and others. It provided standardized splits and evaluated multiple encoder-decoder variants, aiming to foster universal segmentation models. However, its focus remained on a smaller set of architectures and lacked comprehensive analysis of robustness, efficiency, and cross-paradigm comparisons.
nnWNet \citep{nnwnet} proposed architectural modifications to integrate convolutions and transformers within a U-Net framework, addressing the need for continuous transmission of local and global features. Although it benchmarked on multiple 2D and 3D datasets, its evaluation was limited to a small number of models and lacked systematic efficiency analysis.
nnU-Net Revisited \citep{nnunetrevisted} critically examined recent architectural claims, showing that properly configured CNN-based U-Nets could still match or outperform newer transformer and Mamba-based models when trained with sufficient resources. This study highlighted the importance of rigorous baselines and computational reproducibility, yet it did not provide a multi-modal, multi-dataset framework for comparing a large number of variants.
Collectively, these efforts underscore the need for a unified, statistically rigorous, and comprehensive benchmark that systematically evaluates a broad spectrum of U-Net variants across diverse modalities, datasets, and deployment metrics.

\subsubsection{\textbf{Targeted Improvements of U-Bench}}
As summarized in Tab.~\ref{tab:summary}, existing medical image segmentation benchmarks suffer from limited modality coverage, insufficient evaluation diversity, narrow architectural scope, and lack of dataset-specific analysis—all of which hinder comprehensive assessment of model generalization. To address these gaps, U-Bench is designed with three targeted innovations, establishing a more comprehensive and clinically relevant evaluation framework while aligning with its core goals: evaluating 100 U-Net variants across 28 datasets and 10 modalities, introducing the performance-efficiency balanced U-Score, and enabling fair, reproducible benchmarking.
\begin{enumerate}
\item \textbf{Multimodality and Full Task Coverage}\par
U-Bench encompasses 10 major medical imaging modalities (ultrasound, dermoscopy, endoscopy, fundus, histopathology, nuclei, X-Ray, MRI, CT, OCT) and integrates 28 datasets (sample sizes: 20-17,000). It covers tasks from macroscopic organ segmentation (e.g., lung CT, cardiac MRI) to microscopic structure segmentation (e.g., histopathological nuclei, retinal microvasculature), with standardized train/test splits. This design tests cross-modality adaptability of models, matching real-world clinical multimodal diagnostic workflows.
\item \textbf{Multi-Dimensional Evaluation System}\par
Beyond traditional accuracy metrics (IoU, Dice), U-Bench introduces three critical evaluation dimensions and a unified U-Score to quantify clinical utility:
\textit{Computational Efficiency}: Standardized reporting of model parameters (M), inference FLOPs (G), and FPS to reflect deployability on resource-constrained devices.
\textit{Generalization Performance}: Zero-shot transfer tests on 8 unseen target datasets (distinct from 20 training source datasets) to assess robustness to domain shifts (e.g., cross-center ultrasound, unseen dermoscopic lesions).
\textit{Statistical Significance}: Paired t-tests between each variant and the original U-Net (p $<$ 0.05 as significant) to validate reliable performance gains.
\textit{U-Score}: A comprehensive metric using quantile normalization and weighted harmonic mean to balance accuracy and efficiency, bridging academic performance and clinical deployment value.
\item \textbf{Large-Scale Reproducible Validation}\par
U-Bench includes 100 publicly available U-Net variants, covering mainstream architectures from 2015 to 2025 (CNN, Transformer, Mamba, RWKV, hybrid designs). To ensure reproducibility, all models adopt official implementations, pre-trained weights (if available), and deep supervision strategies (if applicable).
\end{enumerate}

\section{Details of Data Zoo}\label{app:datazoo}

We summarize the dataset statistics used in this paper in Table~\ref{tab:datasets_summary}. This table details the datasets used for experimental evaluation, covering 10 core imaging modalities, including ultrasound (e.g., BUSI), dermoscopy (e.g., ISIC2018), endoscopy (e.g., Kvasir-SEG), fundus (e.g., CHASE), histopathology (e.g.,  Glas), nuclear (e.g., DSB2018), X-ray (e.g., Montgomery), MRI (e.g., ACDC,), CT (e.g., Synapse), and OCT (e.g., Cystoidfluid). For each dataset, we provide key information such as the segmentation class (binary or multiclass), the number of samples, the year of publication, and a basic description. All datasets used are publicly available. Therefore, we provide access links in the relevant references and supplementary tables. The details are available in Tab.~\ref{tab:datasets_summary}. A brief description of the dataset is as follows:

\noindent\textbf{BUSI.} The Breast Ultrasound Images (BUSI) dataset \citep{al2020dataset}, collected from 600 female patients in 2018, contains 133 normal, 487 benign, and 210 malignant cases with corresponding ground truth labels. The data labels are obtained using ultrasound scans to examine breast cancer lesion areas.

\noindent\textbf{BUS.} The Breast UltraSound (BUS) public dataset \citep{zhang2022busis} includes 562 images (306 benign, 256 malignant) collected via five ultrasound devices, used for generalization experiments. The data labels are obtained using ultrasound scans to examine breast cancer lesion  (or non-lesion) areas.

\noindent\textbf{BUSBRA.} The BUS-BRA dataset \citep{gomez2024bus} comprises 1875 anonymized images from 1064 patients (corresponding to 722 benign and 342 malignant cases) acquired via four ultrasound scanners. The data labels are obtained using ultrasound scans to examine breast cancer lesion (or non-lesion) areas.

\noindent\textbf{TNSCUI.} The Thyroid Nodule Segmentation and Classification in Ultrasound Images 2020 dataset\footnote{Available at: \url{https://tn-scui2020.grand-challenge.org/Home}.} includes 3644 cases from the Chinese Artificial Intelligence Alliance for Thyroid and Breast Ultrasound. The data label is the thyroid nodule area obtained by thyroid ultrasound.

\noindent\textbf{TUCC.} The Thyroid Ultrasound (TUCC) dataset\footnote{Available at: \url{https://aimi.stanford.edu/datasets/thyroid-ultrasound-cine-clip}.} collects data from 167 patients, including 192 biopsy-confirmed nodules. The data label is the thyroid nodule area obtained by thyroid ultrasound.

\noindent\textbf{ISIC2018.} The ISIC 2018 dataset \citep{gutman2016skin} is a large-scale dermoscopy dataset for lesion segmentation, containing 2594 skin lesion images. The data label is the melanoma (or non-lesion) area of the skin disease obtained by dermoscopy imaging.

\noindent\textbf{PH2.} The PH$^2$ database \citep{mendoncca2013ph} includes 200 dermoscopic images with manual segmentation and clinical diagnosis. The data label is the melanoma (or non-lesion) area of the skin disease obtained by dermoscopy imaging.

\noindent\textbf{SkinCancer.} The SkinCancer dataset \citep{kucs2024medsegbench} contains 206 dermoscopic samples extracted from DermIS and DermQuest. The data label is the melanoma (or non-lesion) area of the skin disease obtained by dermoscopy imaging

\noindent\textbf{Covidquex.} The Covidquex dataset \citep{kucs2024medsegbench} includes 2,913 chest X-ray images ($256 \times 256$ pixels) for binary segmentation. The dataset is labeled with COVID-infected areas on chest X-rays.

\noindent\textbf{Montgomery.} The Montgomery dataset \citep{jaeger2014two} contains 138 chest X-rays (80 normal, 58 with tuberculosis). The data label is the tuberculosis lesion (or non-lesion) area on the lung X-ray.

\noindent\textbf{NIH-test.} The NIH-test dataset \citep{tang2019xlsor} is a manually annotated chest X-ray dataset with 100 lung masks. The data labels are lung segmentations from chest X-rays.

\noindent\textbf{DCA.} The DCA dataset \citep{kucs2024medsegbench} contains 134 fundus images ($300 \times 300$ pixels). The data label is the blood vessel segmentation of the fundus image.

\noindent\textbf{Kvasir.} The Kvasir dataset \citep{jha2020kvasir} contains 1000 gastrointestinal polyp images and corresponding ground truth. The data labels are pathological areas of gastrointestinal endoscopic imaging.

\noindent\textbf{CVC-300.} The CVC-300 dataset \citep{vazquez2017benchmark} comprises 60 colonoscopy polyp images ($500 \times 574$ pixels).  The data labels are pathological areas of gastrointestinal endoscopic imaging.

\noindent\textbf{CVC-ClinicDB.} The CVC-ClinicDB dataset \citep{bernal2015wm} includes 612 images from 29 colonoscopy sequences. The data labels are pathological areas of gastrointestinal endoscopic imaging.

\noindent\textbf{Robotool.} The Robotool dataset \citep{kucs2024medsegbench} consists of 500 images extracted from multiple surgical videos. The data label is the instrument area imaged by the endoscope.

\noindent\textbf{Promise.} The Promise dataset \citep{kucs2024medsegbench} includes 1,473 prostate MRI samples ($512 \times 512$ pixels).

\noindent\textbf{ACDC.} The ACDC dataset \citep{bernard2018deep} contains 100 cardiac MRI scans. The data labels for left ventricle (LV), right ventricle (RV), and myocardium (MYO) in heart segmentation.

\noindent\textbf{CHASE.} The CHASE dataset \citep{CHASEDB1} includes 28 retinal images (one per eye from 14 children). The data label is the vascular area of the fundus image.

\noindent\textbf{Stare.} The Stare dataset \citep{hoover2000locating} includes 20 ocular fundus vessel images with manual annotations. The data label is the vascular area of the fundus image.

\noindent\textbf{DRIVE.} The DRIVE dataset \citep{staal2004ridge} is collected from a Dutch diabetic retinopathy screening program. The data label is the vascular area of the fundus image.

\noindent\textbf{Cell.} The Cell dataset \citep{kucs2024medsegbench} consists of 670 nuclei images with a resolution of 320$\times$256 pixels. The data label is the cell nucleus segmentation area.

\noindent\textbf{Glas.} The Glas dataset \citep{sirinukunwattana2015glas} contains 165 H\&E stained slide images for gland segmentation. The data label is the glandular lesion (or non-lesion) area of the Hematoxylin and Eosin image.

\noindent\textbf{Monusac.} The Monusac dataset \citep{kucs2024medsegbench} includes 310 H\&E stained digital tissue images. The data labels are the nucleus regions of H\&E stained histology images.

\noindent\textbf{Tnbcnuclei.} The Tnbcnuclei dataset \citep{kucs2024medsegbench} contains 50 pathological samples for binary segmentation. The data labels are the cell nucleus regions of Hematoxylin and Eosin stained histology images.

\noindent\textbf{Synapse.} The Synapse multi-organ dataset includes 30 abdominal CT scans with 8-organ segmentation. The data labels are 8 abdominal organs (aorta, gallbladder, left kidney, right kidney, liver, pancreas, spleen, stomach).

\noindent\textbf{Cystoidfluid.} The Cystoidfluid dataset \citep{kucs2024medsegbench} contains 1,006 Optical Coherence Tomography images. The dataset is labeled the Cystoid Macular Edema (CME) region of the retina.

\noindent\textbf{DSB2018.} The DSB2018 dataset \citep{hamilton2018kaggle} includes 670 Hematoxylin and Eosin (H\&E)-stained nuclear images. The data label is the cell nucleus segmentation area.

\begin{table}[!t]
\renewcommand{\arraystretch}{1.2}
\caption{Dataset information summary, where 'O' in split type represents ourself-split, and 'S' represents splitting by data source}
\label{tab:datasets_summary}
\centering
\begin{tabular}{l l c l l c c} \hline

\textbf{Modal} & \textbf{Dataset} & \textbf{Category} & \textbf{Quantity} & \textbf{Year} & \textbf{Split type} & \textbf{Source} \\ \hline
\multirow{5}{*}{Ultrasound} 
& BUSI & Binary & 0.5k$\sim$1k & 2020 & O & \href{https://scholar.cu.edu.eg/?q=afahmy/pages/dataset}{link} \\
& BUS & Binary & 0.5k$\sim$1k & 2022 & O & \href{http://cvprip.cs.usu.edu/busbench/}{link} \\
& BUSBRA & Binary & 1k$\sim$2k & 2024 & O & \href{https://zenodo.org/records/8231412}{link} \\
& TNSCUI & Binary & 3k$\sim$4K & 2020 & O & \href{https://tn-scui2020.grand-challenge.org/Home/}{link} \\
& TUCC & Binary & 10k$\sim$20k & - & O & \href{https://aimi.stanford.edu/datasets/thyroid-ultrasound-cine-clip}{link} \\
\hline
\multirow{3}{*}{Dermoscopy} 
& ISIC2018 & Binary & 2k$\sim$3k & 2018 & O & \href{https://challenge.isic-archive.com/data/#2018}{link} \\
& PH$^2$ & Binary & $<$0.5k & 2013 & S & \href{https://www.fc.up.pt/addi/ph2%20database.html}{link} \\
& SkinCancer & Binary & 206 & 2024 & S & \href{https://zenodo.org/records/13358372/files/uwaterlooskincancer_256.npz}{link} \\
\hline
\multirow{4}{*}{X-Ray} 
& Covidquex & Binary & 2k $\sim$ 3k & 2021 & S & \href{https://zenodo.org/records/13358372/files/covidquex_256.npz}{link} \\
& Montgomery & Binary & $<$0.5k & 2014 & S & \href{https://github.com/rsummers11/CADLab/tree/master/Lung_Segmentation_XLSor}{link} \\
& NIH-test & Binary & $<$0.5k & 2019 & S & \href{https://github.com/rsummers11/CADLab/tree/master/Lung_Segmentation_XLSor}{link} \\
& DCA & Binary & $<$0.5k & 2019 & S & \href{https://github.com/rsummers11/CADLab/tree/master/Lung_Segmentation_XLSor}{link} \\
\hline
\multirow{4}{*}{Endoscopy} 
& Kvasir-SEG & Binary & 1k$\sim$2k & 2020 & S & \href{https://datasets.simula.no/kvasir-seg/}{link} \\
& CVC-300 & Binary & $<$0.5k & 2017 & S & \href{https://www.kaggle.com/datasets/nourabentaher/cvc-300}{link} \\
& CVC-ClinicDB & Binary & 0.5k$\sim$1k & 2015 & S & \href{https://www.kaggle.com/datasets/balraj98/cvcclinicdb}{link} \\
& Robotool & Binary & 0.5k$\sim$1k & 2021 & S & \href{https://zenodo.org/records/13358372/files/robotool_256.npz}{link} \\
\hline
\multirow{2}{*}{MRI} 
& Promise & Binary & 1k$\sim$2k & 2024 & S & \href{https://zenodo.org/records/13358372/files/promise12_256.npz}{link} \\
& ACDC & 4-Class & $<$0.5k & 2018 & S & \href{https://github.com/SLDGroup/CASCADE}{link} \\
\hline
\multirow{3}{*}{Fundus} 
& CHASE & Binary & $<$0.5k & 2012 & S & \href{https://github.com/clguo/SA-UNet}{link} \\
& Stare & Binary & $<$0.5k & 2000 & S & \href{https://cecas.clemson.edu/$\sim$ahoover/stare/probing/index.html}{link} \\
& DRIVE & Binary & $<$0.5k & - & S & \href{https://github.com/clguo/SA-UNet}{link} \\
\hline
CT & Synapse & 9-Class & 3k$\sim$4k & 2023 & S & \href{https://www.synapse.org/#!Synapse:syn3193805/wiki/217789}{link} \\
\hline
OCT & Cystoidfluid & Binary & 1k$\sim$2k & 2024 & S & \href{https://zenodo.org/records/13358372/files/cystoidfluid_256.npz}{link} \\
\hline
\multirow{2}{*}{Nuclear}
& DSB2018 & Binary & 0.5k$\sim$1k & 2018 & S & \href{https://github.com/Siyavashshabani/PanNuke_MoNuSeg_cpm17_ConSep_Preprocess}{link} \\
& Cell & Binary & 0.5k$\sim$1k & 2018 & S & \href{https://zenodo.org/records/13358372/files/cellnuclei_256.npz}{link} \\ \hline
\multirow{3}{*}{Histopathology} 
& Monusac & Binary & $<$0.5k & 2016 & S & \href{https://zenodo.org/records/13358372/files/monusac_256.npz}{link} \\
& Tnbcnuclei & Binary & $<$0.5k & 2018 & S & \href{https://zenodo.org/records/13358372/files/tnbcnuclei_256.npz}{link} \\
& Glas & Binary & $<$0.5k & 2015 & S & \href{https://github.com/twpkevin06222/Gland-Segmentation}{link} \\

\hline
\end{tabular}
\end{table}

\section{Details of Model Zoo \label{app:modelzoo}}

We conducted a comprehensive statistical analysis of the 100 models evaluated by U-Stone, as shown in  Tab.~\ref{tab:model_comparison_1}, ~\ref{tab:model_comparison_2} and ~\ref{tab:model_data}.

\begin{itemize}
    \item Tab.~\ref{tab:model_comparison_1} and Tab.~\ref{tab:model_comparison_2} summarize the basic information of the single architecture and hybrid architecture respectively, quantifying critical metrics including deep supervision adoption, pre-training status, zero-shot capability, statistical significance (P-value), parameter count (Params), computational cost (FLOPs), and inference speed (FPS);  
    \item Tab.~\ref{tab:model_data} further clarifies the training foundation of all evaluated models, documenting their publication year, venue, target dataset modality, and open-source repository links for reproducibility.

\end{itemize}

\begin{table*}[htbp]
\centering
\caption{Single-architecture model comparison.}
\label{tab:model_comparison_1}
\resizebox{\textwidth}{!}{
\begin{tabular}{l|l|cccc|rrr}
\Xhline{1px}
Architecture & Model & Deep Supervision & Pre-training & Zero-shot & P-value & Params (M) & FLOPs (G) & FPS \\
\hline
\multirow{44}{*}{CNN} 

&	AtU-Net	~\citep{AtU-Net}	&	 	&	 	&	 	&	 \ding{51}	&	34.88	&	66.63	&	126.09	\\
&	AURA-Net	~\citep{AURA-Net}	&	 	&	 	&	 	&	 	&	52.84	&	25.15	&	121.63	\\
&	CA-Net	~\citep{CA-Net}	&	 	&	 	&	 	&	 \ding{51}	&	2.79	&	5.99	&	31.71	\\
&	CaraNet	~\citep{CaraNet}	&	 \ding{51}	&	 \ding{51}	&	 	&	 	&	44.59	&	11.50	&	26.82	\\
&	CENet	~\citep{CENet}	&	 	&	 	&	 	&	 	&	33.36	&	10.64	&	23.63	\\
&	CE-Net	~\citep{CE-Net}	&	 	&	 \ding{51}	&	 	&	 	&	29.00	&	8.90	&	103.22	\\
&	CFPNet-M	~\citep{CFPNet-M}	&	 	&	 	&	 	&	 	&	0.76	&	3.47	&	73.13	\\
&	CMU-Net	~\citep{CMU-Net}	&	 	&	 	&	 	&	 	&	49.93	&	91.25	&	83.28	\\
&	CMUNeXt	~\citep{CMUNeXt}	&	 	&	 	&	 	&	 	&	3.15	&	7.42	&	161.14	\\
&	CPCANet	~\citep{CPCANet}	&	 	&	 	&	 	&	 	&	43.39	&	13.36	&	16.23	\\
&	CSCA U-Net	~\citep{CSCAU-Net}	&	 \ding{51}	&	 	&	 	&	 	&	35.27	&	13.74	&	44.99	\\
&	DANet	~\citep{DANet}	&	 \ding{51}	&	 	&	 	&	 \ding{51}	&	94.51	&	33.24	&	48.18	\\
&	DCSAU-Net	~\citep{DCSAU-Net}	&	 	&	 	&	 	&	 	&	10.81	&	23.83	&	56.84	\\
&	DDS-UNet	~\citep{DDS-UNet}	&	 	&	 	&	 	&	 	&	43.62	&	17.40	&	36.87	\\
&	DoubleUNetPlus	~\citep{DoubleUNetPlus}	&	 \ding{51}	&	 	&	 \ding{51}	&	 	&	29.29	&	53.96	&	100.99	\\
&	ERDUnet	~\citep{ERDUnet}	&	 	&	 \ding{51}	&	 	&	 	&	10.21	&	10.29	&	43.18	\\
&	ESKNet	~\citep{ESKNet}	&	 	&	 	&	 \ding{51}	&	 \ding{51}	&	26.71	&	45.28	&	75.38	\\
&	G-CASCADE	~\citep{G-CASCADE}	&	 \ding{51}	&	 \ding{51}	&	 	&	 	&	26.63	&	5.54	&	62.77	\\
&	LFU-Net	~\citep{LFU-Net}	&	 	&	 	&	 	&	 	&	0.05	&	0.76	&	167.91	\\
&	LV-UNet	~\citep{LV-UNet}	&	 	&	 	&	 	&	 	&	0.92	&	0.21	&	139.30	\\
&	MALUNet	~\citep{MALUNet}	&	 	&	 	&	 	&	 	&	0.18	&	0.08	&	108.64	\\
&	MBSNet	~\citep{MBSNet}	&	 	&	 	&	 	&	 	&	3.98	&	6.86	&	115.10	\\
&	MCA-Unet	~\citep{MCA-Unet}	&	 \ding{51}	&	 	&	 	&	 	&	8.66	&	58.02	&	12.26	\\
&	MDSA-UNet	~\citep{MDSA-UNet}	&	 	&	 	&		&		&	6.58	&	5.65	&	77.36	\\
&	MEGANet	~\citep{MEGANet}	&	 \ding{51}	&	 	&	 	&	 	&	29.27	&	11.71	&	59.62	\\
&	MMUNet	~\citep{MMUNet}	&	 	&	 	&	 	&	 	&	17.73	&	24.04	&	46.93	\\
&	MSLAU-Net	~\citep{MSLAU-Net}	&	 	&	 \ding{51}	&	 	&	 	&	21.88	&	6.27	&	35.34	\\
&	MSRF-Net	~\citep{MSRF-Net}	&	 \ding{51}	&	 	&	 \ding{51}	&	 \ding{51}	&	22.50	&	109.73	&	33.16	\\
&	MultiResUNet	~\citep{MultiResUNet}	&	 	&	 	&	 	&	 	&	7.25	&	18.76	&	84.31	\\
&	PraNet	~\citep{PraNet}	&	 \ding{51}	&	 \ding{51}	&	 \ding{51}	&	 	&	50.01	&	11.96	&	27.55	\\
&	ResNet34UnetPlus	~\citep{ResNet34UnetPlus}	&	 	&	 	&	 	&	 \ding{51}	&	26.90	&	37.63	&	84.54	\\
&	ResU-KAN	~\citep{ResU-KAN}	&	 	&	 	&	 	&	 	&	18.59	&	7.78	&	67.56	\\
&	ResUNetPlusPlus	~\citep{ResUNetPlusPlus}	&	 	&	 	&	 	&	 	&	14.48	&	70.99	&	91.72	\\
&	RollingUnet	~\citep{RollingUnet}	&	 	&	 	&	 	&	 	&	7.10	&	8.28	&	31.85	\\
&	SimpleUNet	~\citep{SimpleUNet}	&	 	&	 	&	 	&	 	&	0.06	&	0.74	&	414.31	\\
&	TA-Net	~\citep{TA-Net}	&	 	&	 	&	 	&	 	&	29.57	&	9.32	&	94.43	\\
&	TinyU-Net	~\citep{TinyU-Net}	&	 	&	 	&	 	&	 	&	0.48	&	1.66	&	150.32	\\
&	UACANet	~\citep{UACANet}	&	 \ding{51}	&	 \ding{51}	&	 \ding{51}	&	 \ding{51}	&	67.11	&	31.55	&	27.79	\\
&	U-KAN	~\citep{U-KAN}	&	 	&	 	&	 	&	 	&	9.38	&	6.89	&	93.47	\\
&	ULite	~\citep{ULite}	&	 	&	 	&	 	&	 	&	0.88	&	0.76	&	323.06	\\
&	U-Net	~\citep{U-Net}	&	 	&	 	&	 	&	 	&	34.53	&	65.52	&	137.05	\\
&	UNet3+	~\citep{UNet3+}	&	 	&	 	&	 	&	 	&	26.97	&	199.74	&	50.70	\\
&	UNeXt	~\citep{UNeXt}	&	 	&	 	&	 	&	 \ding{51}	&	1.47	&	0.57	&	256.68	\\
&	UTANet	~\citep{UTANet}	&	 	&	 	&	 	&	 \ding{51}	&	45.03	&	87.59	&	85.63	\\

\hline
\multirow{11}{*}{Mamba} 

&	AC-MambaSeg	~\citep{AC-MambaSeg}	&	 	&	 	&	 	&	 	&	7.42	&	6.27	&	35.64	\\
&	CFM-UNet	~\citep{CFM-UNet}	&	 	&	 	&	 \ding{51}	&	 \ding{51}	&	52.96	&	6.17	&	38.08	\\
&	H-vmunet	~\citep{H-vmunet}	&	 	&	 	&	 	&	 	&	6.44	&	0.74	&	16.30	\\
&	Mamba-UNet	~\citep{Mamba-UNet}	&	 	&	 \ding{51}	&	 	&	 	&	15.48	&	4.60	&	94.47	\\
&	MedVKAN	~\citep{MedVKAN}	&	 	&	 	&	 	&	 	&	43.58	&	14.13	&	45.91	\\
&	MUCM-Net	~\citep{MUCM-Net}	&	 	&	 	&	 	&	 	&	0.08	&	0.06	&	107.24	\\
&	Swin-UMamba	~\citep{Swin-UMamba}	&	 	&	 \ding{51}	&	 	&	 	&	55.06	&	43.93	&	58.52	\\
&	Swin-UMambaD	~\citep{Swin-UMambaD}	&	 	&	 \ding{51}	&	 	&	 	&	21.74	&	6.20	&	65.48	\\
&	UltraLight-VM-UNet	~\citep{UltraLight-VM-UNet}	&	 	&	 	&	 \ding{51}	&	 	&	0.04	&	0.06	&	82.45	\\
&	VM-UNet	~\citep{VM-UNet}	&	 	&	 \ding{51}	&	 	&	 \ding{51}	&	34.62	&	7.56	&	48.08	\\
&	VM-UNetV2	~\citep{VM-UNetV2}	&	 	&	 \ding{51}	&	 	&	 	&	17.91	&	4.40	&	62.85	\\

\hline
\multirow{3}{*}{RWKV} 
% & Zig-RiR~\citep{Zig-RiR}  &  &  &  & \ding{51} & 24.25 & 3.31 & 35.59 \\
% & RWKV-UNet~\cite{RWKV-UNet} &  & \ding{51} &  &  & 17.10 & 14.58 & 76.14 \\
% & U-RWKV~\cite{U-RWKV} &  &  &  &  & 1.47 & 0.57 & 256.68 \\
&	RWKV-UNet	~\citep{RWKV-UNet}	&	 	&	 \ding{51}	&	 	&	 	&	17.10	&	14.58	&	76.14	\\
&	U-RWKV	~\citep{U-RWKV}	&	 	&	 	&	 	&	 	&	2.82	&	6.90	&	107.52	\\
&	Zig-RiR	~\citep{Zig-RiR}	&	 	&	 	&	 	&	 \ding{51}	&	24.25	&	3.31	&	35.59	\\

\hline
\multirow{8}{*}{Transformer} 

&	DC-ViT	~\citep{DC-ViT}	&	 	&	 	&	 \ding{51}	&	 	&	6.84	&	20.87	&	65.34	\\
&	ColonSegT	~\citep{ColonSegT}	&	 	&	 	&	 	&	 	&	5.01	&	62.16	&	135.43	\\
&	ConvFormer	~\citep{ConvFormer}	&	 	&	 \ding{51}	&	 	&	 	&	115.61	&	121.13	&	43.05	\\
&	CSWin-UNet	~\citep{CSWin-UNet}	&	 	&	 \ding{51}	&	 	&	 	&	23.57	&	6.14	&	33.05	\\
&	DAE-Former	~\citep{DAE-Former}	&	 	&	 \ding{51}	&	 	&	 \ding{51}	&	29.69	&	34.10	&	55.23	\\
&	MISSFormer	~\citep{MISSFormer}	&	 	&	 \ding{51}	&	 	&	 \ding{51}	&	35.45	&	7.25	&	57.68	\\
&	Polyp-PVT	~\citep{Polyp-PVT}	&	 \ding{51}	&	 \ding{51}	&	 \ding{51}	&	 \ding{51}	&	25.11	&	5.30	&	67.64	\\
&	SwinUnet	~\citep{SwinUnet}	&	 	&	 \ding{51}	&	 	&	 	&	41.34	&	8.69	&	63.25	\\
&	UNETR	~\citep{UNETR}	&	 	&	 	&	 	&	 \ding{51}	&	87.51	&	26.41	&	104.25	\\

\Xhline{1px}
\end{tabular}
}
\end{table*}

\begin{table*}[htbp]
\centering
\caption{Hybrid architecture model comparison.}
\resizebox{\linewidth}{!}{
\begin{tabular}{l|cccc|rrr}
\Xhline{1px}
 Model & Deep Supervision & Pre-training & Zero-shot & P-value & Params (M) & FLOPs (G) & FPS \\
\hline

BEFUNet	~\citep{BEFUNet}	&	 	&	 \ding{51}	&	 	&	 \ding{51}	&	42.61	&	7.95	&	69.89	\\
CASCADE	~\citep{CASCADE}	&	 \ding{51}	&	 \ding{51}	&	 	&	 	&	35.27	&	8.15	&	57.91	\\
CFFormer	~\citep{CFFormer}	&	 	&	 	&	 \ding{51}	&	 \ding{51}	&	158.44	&	71.17	&	30.28	\\
DA-TransUNet	~\citep{DA-TransUNet}	&	 	&	 	&	 	&	 \ding{51}	&	2.60	&	6.92	&	67.48	\\
DS-TransUNet	~\citep{DS-TransUNet}	&	 \ding{51}	&	 \ding{51}	&	 \ding{51}	&	 	&	171.34	&	51.15	&	24.28	\\
D-TrAttUnet	~\citep{D-TrAttUnet}	&	 \ding{51}	&	 	&	 	&	 \ding{51}	&	104.16	&	54.00	&	53.85	\\
EMCAD	~\citep{EMCAD}	&	 \ding{51}	&	 \ding{51}	&	 	&	 	&	26.76	&	5.60	&	56.17	\\
EViT-UNet	~\citep{EViT-UNet}	&	 	&	 \ding{51}	&	 	&	 	&	54.79	&	8.36	&	16.73	\\
FAT-Net	~\citep{FAT-Net}	&	 	&	 \ding{51}	&	 	&	 	&	29.62	&	42.80	&	76.01	\\
FCNFormer	~\citep{FCNFormer}	&	 	&	 \ding{51}	&	 \ding{51}	&	 	&	52.94	&	40.88	&	25.70	\\
GH-UNet	~\citep{GH-UNet}	&	 \ding{51}	&	 	&	 	&	 \ding{51}	&	12.81	&	21.58	&	14.61	\\
H2Former	~\citep{H2Former}	&	 	&	 	&	 	&	 	&	33.63	&	32.25	&	55.26	\\
HiFormer	~\citep{HiFormer}	&	 	&	 \ding{51}	&	 	&	 	&	34.14	&	17.75	&	68.12	\\
LeViT-UNet	~\citep{LeViT-UNet}	&	 	&	 \ding{51}	&	 	&	 	&	17.53	&	27.24	&	102.91	\\
LGMSNet	~\citep{LGMSNet}	&	 	&	 	&	 \ding{51}	&	 \ding{51}	&	2.32	&	4.89	&	105.04	\\
MedFormer	~\citep{MedFormer}	&	 	&	 	&	 	&	 	&	28.07	&	21.79	&	59.85	\\
MedT	~\citep{MedT}	&	 	&	 \ding{51}	&	 	&	 	&	1.37	&	2.41	&	5.15	\\
MERIT	~\citep{MERIT}	&	 	&	 \ding{51}	&	 	&	 \ding{51}	&	147.68	&	33.28	&	18.69	\\
MFMSNet	~\citep{MFMSNet}	&	 \ding{51}	&	 \ding{51}	&	 \ding{51}	&	 	&	31.56	&	10.08	&	13.44	\\
MobileUViT	~\citep{MobileUViT}	&	 	&	 	&	 \ding{51}	&	 	&	6.21	&	10.43	&	96.80	\\
MT-UNet	~\citep{MT-UNet}	&	 	&	 	&	 	&	 \ding{51}	&	75.07	&	57.72	&	11.23	\\
Perspective-Unet	~\citep{Perspective-Unet}	&	 	&	 	&	 	&	 	&	111.08	&	124.48	&	41.80	\\
ScribFormer	~\citep{ScribFormer}	&	 \ding{51}	&	 	&	 	&	 \ding{51}	&	47.91	&	44.63	&	35.25	\\
SCUNet++	~\citep{SCUNet++}	&	 	&	 \ding{51}	&	 	&	 	&	43.54	&	16.68	&	59.66	\\
SwinUNETR	~\citep{SwinUNETR}	&	 	&	 	&	 	&	 	&	6.29	&	4.86	&	84.41	\\
TransAttUnet	~\citep{TransAttUnet}	&	 	&	 \ding{51}	&	 	&	 	&	22.65	&	88.78	&	99.76	\\
TransFuse	~\citep{TransFuse}	&	 \ding{51}	&	 \ding{51}	&	 	&	 	&	26.17	&	11.53	&	59.97	\\
TransNorm	~\citep{TransNorm}	&	 	&	 \ding{51}	&	 	&	 	&	105.59	&	39.28	&	42.59	\\
TransResUNet	~\citep{TransResUNet}	&	 	&	 \ding{51}	&	 	&	 	&	27.07	&	24.06	&	85.84	\\
TransUNet	~\citep{TransUNet}	&	 	&	 \ding{51}	&	 	&	 	&	93.23	&	32.23	&	58.45	\\
UCTransNet	~\citep{UCTransNet}	&	 	&	 	&	 	&	 \ding{51}	&	66.24	&	43.06	&	35.12	\\
UNetV2	~\citep{UNetV2}	&	 \ding{51}	&	 	&	 \ding{51}	&	 \ding{51}	&	25.13	&	5.40	&	60.33	\\
UTNet	~\citep{UTNet}	&	 	&	 	&	 	&	 	&	14.41	&	20.49	&	76.67	\\

\Xhline{1px}
\end{tabular}
}
\label{tab:model_comparison_2}
\end{table*}

\begin{table}[htbp]
\centering
\caption{Training modal information of all evaluation models.}
\small
\label{tab:model_data}

\resizebox{0.68\textwidth}{!}{%
  \begin{tabular}{lllllc}
    \Xhline{1px}
    Architecture & Model & Year & Publication & Modality & Github \\
    \Xhline{1px}
    \multirow{45}{*}{CNN} 
    & AttU-Net & 2018 & MIDL &  CT & \href{https://github.com/ozan-oktay/Attention-Gated-Networks}{[link]} \\
    & AURA-Net & 2021 & ISBI & Microscopy & \href{https://arxiv.org/pdf/2103.14031.pdf}{[link]} \\
    & CA-Net & 2020 & TMI & Dermoscopy, MRI & \href{https://github.com/HiLab-git/CA-Net}{[link]} \\
    & CaraNet & 2022 & SPIE Medical Imaging & Colonoscopy, MRI & \href{https://github.com/AngeLouCN/CaraNet}{[link]} \\
    & CENet & 2019 & TMI & Fundus Image, CT, Microscopy, OCT & \href{https://github.com/Guzaiwang/CE-Net}{[link]} \\
    & CE-Net & 2019 & Medical Imaging & Fundus, CT, Microscopy, OCT & \href{https://github.com/Guzaiwang/CE-Net}{[link]} \\
    & CFPNet-M & 2021 & Medical Imaging & Thermography, Microscopy, Colonoscopy, Dermoscopy, Fundus & \href{https://github.com/AngeLouCN/CFPNet-Medicine}{[link]} \\
    & CMU-Net & 2023 & ISBI & Ultrasound & \href{https://github.com/FengheTan9/CMU-Net}{[link]} \\
    & CMUNeXt & 2024 & ISBI & Ultrasound & \href{https://github.com/FengheTan9/CMUNeXt}{[link]} \\
    & CPCANet & 2023 & CMI&  MRI, Dermoscopy & \href{https://github.com/Cuthbert-Huang/CPCANet}{[link]} \\
    & CSCA U-Net & 2024 & AIIM & Colonoscopy, Pathology, Ultrasound & \href{https://github.com/xiaolanshu/CSCA-U-Net}{[link]} \\
    & DANet & 2024 & Plos one & Ultrasound & \href{https://github.com/AyushRoy2001/DAUNet}{[link]} \\
    & DCSAU-Net & 2023 & CBM & Colonoscopy, Microscopy, Dermoscopy & \href{https://github.com/xq141839/DCSAU-Net}{[link]} \\
    & DDS-UNet & 2024 & IET Image Processing & CT & \href{https://github.com/DeepPicEI/DDS-UNet}{[link]} \\
    & DoubleUNetPlus & 2020 & IEEE CBMS & Colonoscopy, Dermoscopy, Microscopy & \href{https://github.com/DebeshJha/2020-CBMS-DoubleU-Net}{[link]} \\
    & ERDUnet & 2024 & TCSVT & Microscopy, Dermoscopy, Colonoscopy, Pathology, MRI & \href{https://github.com/caijilia/ERDUnet}{[link]} \\
    & ESKNet & 2023 & CMPB & Ultrasound & \href{https://github.com/CGPxy/ESKNet}{[link]} \\
    & G-CASCADE & 2023 & WACV & CT, MRI, Dermoscopy, Colonoscopy & \href{https://github.com/SLDGroup/G-CASCADE}{[link]} \\
    & LFU-Net & 2023 & CMI& CT, MRI & \href{https://github.com/dengdy22/U-Nets/blob/main/LFU-Net}{[link]} \\
    & LV-UNet & 2024 & BIBM & Dermoscopy, Ultrasound, Colonoscopy & \href{https://github.com/juntaoJianggavin/LV-UNet}{[link]} \\
    & MALUNet & 2022 & BIBM & Dermoscopy & \href{https://github.com/JCruan519/MALUNet}{[link]} \\
    & MBSNet & 2021 & MSSP & Dermoscopy, Ultrasound, Colonoscopy & \href{https://github.com/YuLionel/MBSNet}{[link]} \\
    & MCA-Unet & 2023 &CMPBU & CT & \href{https://github.com/McGregorWwww/MCA-UNet}{[link]} \\
    & MDSA-UNet & 2025 & JBHI & Ultrasound, CT, Dermoscopy & \href{https://github.com/NEU-LX/MDSA-UNet}{[link]} \\
    & MEGANet & 2024 & WACV & Colonoscopy & \href{https://github.com/UARK-AICV/MEGANet}{[link]} \\
    & MMUNet & 2024 & BSPC & Histological image & \href{https://github.com/Yuanhaojun513/MMUNet}{[link]} \\
    & MSLAU-Net & 2025 & arXiv (cs.CV) & CT, MRI, Colonoscopy & \href{https://github.com/Monsoon49/MSLAU-Net}{[link]} \\
    & MSRF-Net & 2021 & JBHI & Colonoscopy, Microscopy, Dermoscopy & \href{https://github.com/vikram71198/MSRF_Net}{[link]} \\
    & MultiResUNet & 2020 & Neural networks & Microscopy, Dermoscopy, Colonoscopy, MRI & \href{https://github.com/nibtehaz/MultiResUNet}{[link]} \\
    & PraNet & 2020 & MICCAI & Colonoscopy & \href{https://github.com/DengPingFan/PraNet}{[link]} \\
    & ResNet34UnetPlus &2018 & TMI & Microscopy, CT, MRI & \href{https://github.com/MrGiovanni/UNetPlusPlus}{[link]} \\
    & ResUNetPlusPlus & 2019 & ISM & Colonoscopy & \href{https://github.com/DebeshJha/ResUNetPlusPlus}{[link]} \\
    & RollingUnet & 2024 & AAAI & Ultrasound, Histological image, Dermoscopy, Fundus & \href{https://github.com/Jiaoyang45/Rolling-Unet}{[link]} \\
    & SimpleUNet & 2025 & arXiv & Ultrasound, Dermoscopy, Colonoscopy & \href{https://github.com/Frankyu5666666/SimpleUNet}{[link]} \\
    & TA-Net & 2022 & WACV & Histological image & \href{https://github.com/shuchao1212/TA-Net}{[link]} \\
    & TinyU-Net & 2024 & MICCAI & Dermoscopy, CT & \href{https://github.com/ChenJunren-Lab/TinyU-Net}{[link]} \\
    & UACANet & 2021 & ACM MM & Colonoscopy & \href{https://github.com/plemeri/UACANet}{[link]} \\
    & ULite & 2023 & APSIPA & Dermoscopy, Microscopy, Histological image & \href{https://github.com/duong-db/U-Lite}{[link]} \\
    & U-Net & 2015 & MICCAI & Microscopy, Microscopy & \href{https://github.com/milesial/Pytorch-UNet}{[link]} \\
    & UNet3+ & 2020 & ICASSP & CT & \href{https://github.com/ZJUGiveLab/UNet-Version}{[link]} \\
    & UNeXt & 2022 & MICCAI & Dermoscopy, Ultrasound & \href{https://github.com/jeya-maria-jose/UNeXt-pytorch}{[link]} \\
    & UTANet & 2025 & AAAI & Histology Image, Microscopy, Abdominal CT, Dermoscopy & \href{https://github.com/AshleyLuo001/UTANet}{[link]} \\
    & ResU-KAN & 2025 & Applied Intelligence & Ultrasound, Histological, Colonoscopy & \href{https://github.com/Alfreda12/ResU-KAN}{[link]} \\
    & U-KAN & 2025 & AAAI & Ultrasound, Histological image, Colonoscopy & \href{https://github.com/CUHK-AIM-Group/U-KAN}{[link]} \\
    \hline
    \multirow{6}{*}{Mamba} 
    & Mamba-UNet & 2024 & CoRR & MRI, CT & \href{https://github.com/ziyangwang007/Mamba-UNet}{[link]} \\
    & MedVKAN & 2025 & arxiv & Microscopy, MRI, Ultrasound, CT & \href{https://github.com/beginner-cjh/MedVKAN}{[link]} \\
    & Swin-UMambaD & 2024 & TMI & MRI, Endoscopy, Microscopy & \href{https://github.com/JiarunLiu/Swin-UMamba}{[link]} \\
    & UltraLight-VM-UNet & 2025 & Patterns & Dermoscopy & \href{https://github.com/wurenkai/UltraLight-VM-UNet}{[link]} \\
    & VM-UNet & 2024 & CoRR & Dermoscopy, CT & \href{https://github.com/JCruan519/VM-UNet}{[link]} \\
    & VM-UNetV2 & 2024 & ISBRA & Dermoscopy, Colonoscopy & \href{https://github.com/nobodyplayer1/VM-UNetV2}{[link]} \\
     & AC-MambaSeg & 2024 &ICGTSD & Dermoscopy & \href{https://github.com/vietthanh2710/AC-MambaSeg}{[link]} \\
      & CFM-UNet & 2025 & Scientific Reports & CT,  MRI, Colonoscopy, MRI & \href{https://github.com/Jiacheng-Han/CFM-UNet}{[link]} \\
      & MUCM-Net & 2024 & CoRR & Dermoscopy & \href{https://github.com/chunyuyuan/MUCM-Net}{[link]} \\
      & Swin-UMamba & 2024 & MICCAI & MRI, Endoscopy, Microscopy & \href{https://github.com/JiarunLiu/Swin-UMamba}{[link]} \\
    \hline
    \multirow{3}{*}{RWKV} 
    & Zig-RiR & 2025 & TMI & Dermoscopy,  CT,  MRI, Microscopy & \href{https://github.com/txchen-USTC/Zig-RiR}{[link]} \\
    & RWKV-UNet & 2025 & CoRR & CT, MRI, Ultrasound, Colonoscopy, Dermoscopy, Histological image & \href{https://github.com/juntaoJianggavin/RWKV-UNet}{link} \\
    & U-RWKV & 2025 & MICCAI & Ultrasound, Colonoscopy, Dermoscopy, CT & \href{https://github.com/hbyecoding/U-RWKV}{link} \\
    \hline
    \multirow{9}{*}{Transformer} 
    & DC-ViT & 2024 & CVPR & Natural images & \href{https://github.com/Michelia-zhx/dcvit}{[link]} \\
    & ColonSegT & 2021 & IEEE ACCESS & Colonoscopy & \href{https://github.com/DebeshJha/ColonSegNet}{[link]} \\
    & ConvFormer & 2023 & MICCAI & Ultrasound, Dermoscopy, CT & \href{https://github.com/xianlin7/ConvFormer}{[link]} \\
    & CSWin-UNet & 2025 & Information Fusion & CT, MRI, Dermoscopy & \href{https://github.com/eatbeanss/CSWin-UNet}{[link]} \\
    & DAE-Former & 2023 & IWPIM & CT, Dermoscopy & \href{https://github.com/xmindflow/DAEFormer}{[link]} \\
    & MISSFormer & 2023 & TMI & CT, MRI & \href{https://github.com/ZhifangDeng/MISSFormer}{[link]} \\
    & Polyp-PVT & 2021 & arXiv & Colonoscopy & \href{https://github.com/DengPingFan/Polyp-PVT}{[link]} \\
    & SwinUnet & 2022 & ECCVW & CT, MRI & \href{https://github.com/HuCaoFighting/Swin-Unet}{[link]} \\
    & UNETR & 2022 & WACV & CT, MRI & \href{https://github.com/tamasino52/UNETR}{[link]} \\
    \hline
      \multirow{32}{*}{Hybrid} 
      & BEFUNet & 2024 & arXiv & CT, Microscopy, Dermoscopy & \href{https://github.com/Omid-Nejati/BEFUnet}{[link]} \\
      & CASCADE & 2023 & WACV & CT, MRI, Colonoscopy & \href{https://github.com/SLDGroup/CASCADE}{[link]} \\
      & CFFormer & 2025 & ESA & Ultrasound, Dermoscopy, Colonoscopy, CT, MRI & \href{https://github.com/JiaxuanFelix/CFFormer}{[link]} \\
      & DA-TransUNet & 2024 & FBB & CT, Colonoscopy, X-ray, Dermoscopy, Endoscopy & \href{https://github.com/SUN-1024/DA-TransUnet}{[link]} \\
      & DS-TransUNet & 2022 & TIM & Colonoscopy, Dermoscopy, Histology, Microscopy & \href{https://github.com/TianBaoGe/DS-TransUNet}{[link]} \\
      & D-TrAttUnet & 2024 & CBM & CT, Histology Image, Microscopy & \href{https://github.com/faresbougourzi/D-TrAttUnet}{[link]} \\
      & EMCAD & 2024 & CVPR & Colonoscopy, Dermoscopy, Ultrasound, CT, MRI & \href{https://github.com/SLDGroup/EMCAD}{[link]} \\
      & EViT-UNet & 2025 & ISBI & CT, Histology Image, Microscopy & \href{https://github.com/Retinal-Research/EVIT-UNET}{[link]} \\
      & FAT-Net & 2022 & MIA & Dermoscopy & \href{https://github.com/SZUcsh/FAT-Net}{[link]} \\
      & FCNFormer & 2022 & MICCAI & Colonoscopy & \href{https://github.com/CVML-UCLan/FCBFormer}{[link]} \\
      & GH-UNet & 2025 &  Digital Medicine & Dermoscopy, Colonoscopy, Fundus, MRI, CT & \href{https://github.com/xiachashuanghua/GH-UNet}{[link]} \\
      & H2Former & 2023 & TMI & Fundus, Colonoscopy, Dermoscopy, MRI, CT & \href{https://github.com/NKUhealong/H2Former}{[link]} \\
      & HiFormer & 2023 & WACV & CT, Dermoscopy, Microscopy & \href{https://github.com/amirhossein-kz/HiFormer}{[link]} \\
      & LeViT-UNet & 2023 & PRCV & CT, MRI & \href{https://github.com/apple1986/LeViT-UNet}{[link]} \\
      & LGMSNet & 2025 & ECAI & Ultrasound, Dermoscopy, Colonoscopy, CT & \href{https://github.com/cq-dong/LGMSNet}{[link]} \\
      & MedFormer & 2023 & arXiv & MRI, CT & \href{https://github.com/DL4mHealth/Medformer}{[link]} \\
      & MedT & 2021 & MICCAI & Ultrasound, Histology Image, Microscopy & \href{https://github.com/jeya-maria-jose/Medical-Transformer}{[link]} \\
      & MERIT & 2023 & MIDL & CT, MRI & \href{https://github.com/SLDGroup/MERIT}{[link]} \\
      & MFMSNet & 2023 & UMB & Ultrasound & \href{https://github.com/wrc990616/MFMSNet}{[link]} \\
      & Mobile U-ViT & 2025 & ACM MM & Ultrasound, Dermoscopy, Colonoscopy, CT & \href{https://github.com/FengheTan9/Mobile-U-ViT}{[link]} \\
      & MT-UNet & 2022 & ICASSP & CT, MRI & \href{https://github.com/Dootmaan/MT-UNet}{[link]} \\
      & Perspective-Unet & 2024 & MICCAI & CT, MRI & \href{https://github.com/tljxyys/Perspective-Unet}{[link]} \\
      & ScribFormer & 2024 & TMI & MRI, CT & \href{https://github.com/HUANGLIZI/ScribFormer}{[link]} \\
      & SCUNet++ & 2024 & WACV & CT & \href{https://github.com/JustlfC03/SCUNet-plusplus}{[link]} \\
      & SwinUNETR & 2021 & MICCAI & MRI & \href{https://github.com/Project-MONAI/research-contributions/tree/main/SwinUNETR}{[link]} \\
      & TransAttUnet & 2022 & TIM & Dermoscopy, X-ray, CT, Biological Image, Histology Image & \href{https://github.com/YishuLiu/TransAttUnet}{[link]} \\
      & TransFuse & 2021 & MICCAI & Colonoscopy, Dermoscopy, X-ray, MRI & \href{https://github.com/Rayicer/TransFuse}{[link]} \\
      & TransNorm & 2022 & IEEE Access & CT, Dermoscopy, Microscopy & \href{https://github.com/rezazad68/transnorm}{[link]} \\
      & TransResUNet & 2022 & CoRR & Colonoscopy & \href{https://github.com/nikhilroxtomar/TransResUNet}{[link]} \\
      & TransUNet & 2021 & arXiv & Abdominal CT, MRI & \href{https://github.com/Beckschen/TransUNet}{[link]} \\
      & UCTransNet & 2022 & AAAI & Histology Image, Microscopy, CT & \href{https://github.com/McGregorWwww/UCTransNet}{[link]} \\
      & UNetV2 & 2025 & ISBI & Dermoscopy, Colonoscopy & \href{https://github.com/yaoppeng/U-Net_v2}{[link]} \\
      & UTNet & 2021 & MICCAI & MRI & \href{https://github.com/yhygao/UTNet}{[link]} \\
    \Xhline{1px}
  \end{tabular}
  }
\end{table}

\section{Details of U-Score \label{app:uscore}}
Clinical deployment of segmentation models often requires operation under constrained resources. However, existing evaluations focus predominantly on segmentation performance, while failing to balance key computational factors such as model size, inference cost, and speed. This disconnect makes it difficult to assess real-world deployability. To bridge this gap, we introduce U-Score, a unified metric that quantifies the trade-off between performance and efficiency using quantile statistics under large-scale benchmark. Specifically, we report the 10th and 90th percentiles of IoU, Params, FLOPs, and FPS, as summarized in Tab.~\ref{tab:sourceq10} and~\ref{tab:zeroq10}. The formulation is represented as follow.

Given model $i$, we compute IoU $A_i$ across datasets, parameter $P_i$ in millions, FLOPs $G_i$ in GLOPs, and runtime speed $S_i$ in FPS. We normalize each component using the 10th and 90th percentiles computed over the model zoo. Let $Q_{10}^M$ and $Q_{90}^M$ denote the 10th and 90th percentiles of metric $M$. The normalized scores are defined as:
\begin{equation}
\begin{aligned}
a_i &= \mathrm{clip}\left( \frac{A_i - Q_{10}^{A}}{Q_{90}^{A} - Q_{10}^{A}},\ 0,\ 1 \right),\quad
p_i = \mathrm{clip}\left( \frac{\log Q_{90}^{P} - \log P_i}{\log Q_{90}^{P} - \log Q_{10}^{P}},\ 0,\ 1 \right), \\
g_i &= \mathrm{clip}\left( \frac{\log Q_{90}^{G} - \log G_i}{\log Q_{90}^{G} - \log Q_{10}^{G}},\ 0,\ 1 \right),\quad
s_i = \mathrm{clip}\left( \frac{S_i - Q_{10}^{S}}{Q_{90}^{S} - Q_{10}^{S}},\ 0,\ 1 \right).
\end{aligned}
\end{equation}

Then, we compute an efficiency subscore via the weighted harmonic mean of $p_i$, $g_i$, and $s_i$. Since we regard storage, cost, and speed as equally important, we set $w_P=w_G=w_S=\tfrac13$, leading to:
\begin{equation}
\mathrm{Eff}_i = \frac{3}{\tfrac{1}{p_i} + \tfrac{1}{g_i} + \tfrac{1}{s_i}}.
\end{equation}

Finally, we combine accuracy and efficiency via a harmonic mean. To balance the two factors equally, we set $\alpha=0.5$, yielding:
\begin{equation}
\mathrm{U\text{-}Score}_i = \frac{2}{\tfrac{1}{a_i} + \tfrac{1}{\mathrm{Eff}_i}}.
\end{equation}

\renewcommand{\multirowsetup}{\centering}  
\begin{table*}[!t]
\centering

\caption{In-domain per-dataset 10th and 90th percentiles of IoU, Params, FLOPs, and FPS. \label{tab:sourceq10}}
\resizebox{1\linewidth}{!}
{
\begin{tabular}{c c  c c  c c  c c  c c}
\Xhline{1.5px} % horizontal line

\multirow{2}{*}{Modality} & \multirow{2}{*}{Dataset} & \multicolumn{2}{c}{IoU (\%)} & \multicolumn{2}{c}{Params (M)} & \multicolumn{2}{c}{FLOPs (G)} & \multicolumn{2}{c}{FPS} \\
\cline{3-10}
& & $Q_{10}^{A}$ & $Q_{90}^{A}$ & $Q_{10}^{P}$ & $Q_{90}^{P}$ & $Q_{10}^{G}$ & $Q_{90}^{G}$ & $Q_{10}^{S}$ & $Q_{90}^{S}$ \\
\hline

\multirow{3}{*}{Ultrasound} & BUSI & 0.58 & 0.71 & 0.39 & 4.32 & 0.88 & 4.20 & 24.28 & 121.63 \\
 & BUSBRA & 0.78 & 0.84 & 0.39 & 4.32 & 0.88 & 4.20 & 24.28 & 121.63 \\
 & TNSCUI & 0.66 & 0.78 & 0.39 & 4.32 & 0.88 & 4.20 & 24.28 & 121.63 \\
\hline
\multirow{2}{*}{Dermoscopy} &  ISIC2018 & 0.81 & 0.84 & 0.39 & 4.32 & 0.88 & 4.20 & 24.28 & 121.63 \\
& SkinCancer & 0.79 & 0.85 & 0.39 & 4.32 & 0.88 & 4.20 & 24.28 & 121.63 \\
\hline
\multirow{2}{*}{Endoscopy} & Kvasir & 0.75 & 0.84 & 0.39 & 4.32 & 0.88 & 4.20 & 24.28 & 121.63 \\
& Robotool & 0.69 & 0.85 & 0.39 & 4.32 & 0.88 & 4.20 & 24.28 & 121.63 \\
\hline
\multirow{2}{*}{Fundus} & CHASE & 0.47 & 0.81 & 0.39 & 4.32 & 0.88 & 4.20 & 24.28 & 121.63 \\
& DRIVE & 0.15 & 0.62 & 0.39 & 4.32 & 0.88 & 4.20 & 24.28 & 121.63 \\
\hline
\multirow{2}{*}{Nuclei} & DSB2018 & 0.85 & 0.88 & 0.39 & 4.32 & 0.88 & 4.20 & 24.28 & 121.63 \\
& CellNuclear & 0.78 & 0.84 & 0.39 & 4.32 & 0.88 & 4.20 & 24.28 & 121.63 \\
\hline
\multirow{2}{*}{Histopathology} & Glas & 0.63 & 0.83 & 0.39 & 4.32 & 0.88 & 4.20 & 24.28 & 121.63 \\
& Monusac & 0.53 & 0.67 & 0.39 & 4.32 & 0.88 & 4.20 & 24.28 & 121.63 \\
\hline
\multirow{3}{*}{X-Ray} & Covidquex & 0.63 & 0.70 & 0.39 & 4.32 & 0.88 & 4.20 & 24.28 & 121.63 \\
& Montgomery & 0.92 & 0.96 & 0.39 & 4.32 & 0.88 & 4.20 & 24.28 & 121.63 \\
& DCA & 0.51 & 0.63 & 0.39 & 4.32 & 0.88 & 4.20 & 24.28 & 121.63 \\
\hline
\multirow{2}{*}{MRI} & ACDC & 0.73 & 0.85 & 0.39 & 4.32 & 0.88 & 4.20 & 24.28 & 121.63 \\
& Promise & 0.78 & 0.87 & 0.39 & 4.32 & 0.88 & 4.20 & 24.28 & 121.63 \\
\hline
CT & Synapse & 0.55 & 0.72 & 0.39 & 4.32 & 0.88 & 4.20 & 24.28 & 121.63 \\
\hline
OCT & Cystoidfluid & 0.63 & 0.83 & 0.39 & 4.32 & 0.88 & 4.20 & 24.28 & 121.63 \\

\Xhline{1.5px}
\end{tabular}
}
\end{table*}
\renewcommand{\multirowsetup}{\centering}  
\begin{table*}[!t]
\centering

\caption{Zero-shot per-dataset 10th and 90th percentiles of IoU, Params, FLOPs, and FPS. \label{tab:zeroq10}}
\resizebox{1\linewidth}{!}
{
\begin{tabular}{c c  c c  c c  c c  c c}
\Xhline{1.5px} % horizontal line

\multirow{2}{*}{Source} & \multirow{2}{*}{Target} & \multicolumn{2}{c}{IoU (\%)} & \multicolumn{2}{c}{Params (M)} & \multicolumn{2}{c}{FLOPs (G)} & \multicolumn{2}{c}{FPS} \\
\cline{3-10}
& & $Q_{10}^{A}$ & $Q_{90}^{A}$ & $Q_{10}^{P}$ & $Q_{90}^{P}$ & $Q_{10}^{G}$ & $Q_{90}^{G}$ & $Q_{10}^{S}$ & $Q_{90}^{S}$ \\
\hline
BUSI & BUS & 0.60 & 0.81 & 0.39 & 4.32 & 0.88 & 4.20 & 24.28 & 121.63 \\
BUSBRA & BUS & 0.78 & 0.85 & 0.39 & 4.32 & 0.88 & 4.20 & 24.28 & 121.63 \\
TNSCUI & TUCC & 0.56 & 0.64 & 0.39 & 4.32 & 0.88 & 4.20 & 24.28 & 121.63 \\
ISIC2018 & PH2 & 0.82 & 0.85 & 0.39 & 4.32 & 0.88 & 4.20 & 24.28 & 121.63 \\
Kvasir & CVC300 & 0.61 & 0.80 & 0.39 & 4.32 & 0.88 & 4.20 & 24.28 & 121.63 \\
Kvasir & CVC-ClinicDB & 0.60 & 0.75 & 0.39 & 4.32 & 0.88 & 4.20 & 24.28 & 121.63 \\
CHASE & STARE & 0.30 & 0.54 & 0.39 & 4.32 & 0.88 & 4.20 & 24.28 & 121.63 \\
Monusac & Tnbcnuclei & 0.25 & 0.44 & 0.39 & 4.32 & 0.88 & 4.20 & 24.28 & 121.63 \\
Montgomery & NIH-test & 0.58 & 0.82 & 0.39 & 4.32 & 0.88 & 4.20 & 24.28 & 121.63 \\
\Xhline{1.5px}
\end{tabular}
}
\end{table*}

\section{Implementation and Evaluation Details \label{app:implementation}}

\subsection{Foreground Characterization Metrics}
\label{app:characterization}
We employ three metrics to characterize dataset-level foreground properties: scale, boundary sharpness, and shape regularity.

\noindent\textbf{Foreground scale.} Foreground scale is measured as the ratio between the foreground area $A_f$ and the total image area $A_t$.  
\begin{equation}
A = \frac{A_f}{A_t}.
\end{equation}
We categorize samples as \emph{small-scale} if $A < 0.05$ and \emph{large-scale} otherwise.

\noindent\textbf{Shape complexity.} We quantify the sharpness of the segmented foreground boundaries using a composite score $S$ derived from two standard geometric descriptors: \textit{circularity} and \textit{solidity}. We categorize samples with $S < 0.5$ as \emph{irregular}, and those with $S \geq 0.5$ as \emph{regular}. The boundary sharpness score $S$ is defined as:
\begin{equation}
S = 0.5 \times \text{Circularity} + 0.5 \times \text{Solidity}.
\end{equation}

Circularity measures how close the shape is to a perfect circle. It is defined as: $\text{Circularity} = \frac{4\pi A_f}{P^2}$, where $A_f$ is the foreground area and $P$ is the perimeter of the contour. Solidity evaluates the extent to which a shape fills its convex hull. It is given by: $\text{Solidity} = \frac{A_f}{A_c}$, where $A_f$ is the foreground area and $A_c$ is the area of its convex hull.

\noindent\textbf{Boundary Sharpness.} We assess boundary sharpness using two complementary measures: \textit{boundary width} and \textit{boundary contrast}.  Given a binary mask $m$, we first construct a narrow boundary ring by applying morphological dilation and erosion. The boundary width is then computed as the ratio between the area of this ring and the contour perimeter: $w = \frac{\text{Area}(\text{Ring})}{P + \epsilon}$, where $P$ denotes the sum of contour perimeters. A larger $w$ indicates blurrier boundaries, while a smaller $w$ corresponds to sharper edges.  To evaluate intensity separation across the boundary, we form two narrow bands: one inside the mask and one outside, each of width $t$ pixels. Let $(\mu_{in}, \sigma_{in})$ and $(\mu_{out}, \sigma_{out})$ denote the mean and standard deviation of pixel intensities inside and outside the boundary band. The boundary contrast is defined as $\text{CNR} = \frac{|\mu_{in} - \mu_{out}|}{\sigma_{in} + \sigma_{out} + \epsilon}$. To obtain a unified measure of boundary clarity, we normalize both $w$ and CNR to $[0,1]$ across the dataset. A composite blur score is then computed as: 
\begin{equation}
B = \frac{w_{norm}}{w_{norm} + c_{norm} + \epsilon},
\end{equation}
where $w_{norm}$ and $c_{norm}$ are the normalized boundary width and contrast, respectively. We categorize samples with $b < 0.6$ as \emph{clear}, and those with $b \geq 0.6$ as \emph{blur}.

\subsection{Model Advisor Agent settings}
\label{app:agent}
We construct a comprehensive feature space that integrates both continuous and discretized descriptors from models and datasets. For model-level attributes, we discretize storage (parameter) into four scales (Tiny: 0-10M, Small: 10-50M, Medium: 50-200M, Large: $>$200M), computation cost (FLOPs) into three levels (Low: 0-10 GFLOPs, Medium: 10-100 GFLOPs, High: $>$100 GFLOPs), and inference speed (FPS) into three categories (Slow: $<$15 FPS, Medium: 15-60 FPS, Fast: $>$60 FPS). On the data characteristics side, we discretize foreground-related properties:foreground scale ($<0.05$ vs.\ $\ge 0.05$, denoting small vs.\ large targets), shape complexity ($<0.5$ vs.\ $\ge 0.5$, irregular vs.\ regular), and boundary sharpness ($<0.6$ vs.\ $\ge 0.6$, clear vs.\ blurry). We train an XGBRanker with the rank:pairwise objective on 18 in-domain datasets, reserving 2 datasets (BUSI and SkinCancer) for testing. Scores are normalized into relevance values within each dataset, with higher relevance indicating better relative performance. Dataset-level grouping is used to enforce within-dataset ranking consistency during training. At evaluation, the ranker outputs predicted scores, which are converted into ranked lists for each dataset. Performance is assessed using $\mathrm{NDCG}@50/20$ for ranking quality, MAP for precision under binary relevance, and correlation metrics (Spearman) to quantify alignment between predicted and ground-truth orderings. Finally, the agent exports the recommended models per test dataset, providing a practical reference list for downstream selection. 

\subsubsection{Data Preprocessing}
\textbf{Experimental Data Split.} For all experiments on the unified dataset, we used the same train-test split. For data without a clear train-test split in the dataset source, we adopted a random split; for data with a known train-test split in the dataset source, we followed the original split. The division method of each dataset is shown in Tab.~\ref{tab:datasets_summary}, where 'O' denotes our self-defined division and 'S' denotes the division consistent with the referenced data source.

\subsubsection{Retraining Details}
The experiments are utilizing eight NVIDIA H20 GPUs. A total of 100 models are trained with 6000 hours. The implementation was built on Python 3.9.0 and PyTorch 2.7.0. The model structure files are primarily obtained from the open-source code of the original models, with only minor modifications (e.g. input and output channels) to some input parameters to adapt to our framework.

Following prior works~\citep{UNeXt,CMU-Net,CMUNeXt,TinyU-Net,MobileUViT,U-RWKV,RWKV-UNet,LGMSNet}, we rescale all images to a resolution of $256 \times 256$ by default and apply standard data augmentation. For models that require a fixed input size (e.g., Swin Transformer variants designed for $224 \times 224$ inputs), we preserve their original settings without rescaling. Augmentations include random $90^\circ$ rotations, random horizontal and vertical flips, and normalization. To ensure fair comparisons, the same preprocessing pipeline is applied consistently across all experiments. Notably, for models that adopt deep supervision, we retain their original training strategy to enable accurate performance evaluation.

\begin{table}[htbp]
\centering
\caption{Hyperparameters in U-Bench}
\label{tab:hyperparameters}
\resizebox{1\linewidth}{!}
{
\begin{tabular}{ccccc}
\Xhline{1px}
\textbf{Optimizer} & \textbf{Learning Rate} & \textbf{Epochs} & \textbf{Random Seed} & \textbf{Batch Size} \\
\hline
SGD (Momentum=0.9, Weight Decay=0.0001) & 0.01 & 300 & 41 & 8 \\
\Xhline{1px}
\end{tabular}
}
\end{table}

\subsubsection{Hyper-Parameters in U-Bench}
Following previous work~\citep{LGMSNet,CMUNeXt,MobileUViT,TransUNet,Mamba-UNet,U-RWKV}, we unify training settings across all models to ensure fair comparisons, as summarized in Tab.~\ref{tab:hyperparameters}. Moreover, we use commonly adopted loss configurations~\citep{LGMSNet,CMUNeXt,MobileUViT,U-RWKV} to promote generalizable results and enable more equitable performance evaluation.
Specifically, for the ground truth \( y \) and the predicted output \( \hat{y} \), the loss function is defined as:
\begin{equation}
     \mathcal{L}=0.5\times BCE(\widehat{y},y)+Dice(\widehat{y},y),
\end{equation}
where BCE denotes the binary cross-entropy loss and Dice denotes the Dice loss. Note that for 3D data ACDC and Synapse, we follow CASCADE~\cite{CASCADE} weights $0.5\times BCE(\widehat{y},y), 0.7\times Dice(\widehat{y},y)$

\subsection{Metrics}
\label{app:f-metrics}

\noindent\textbf{Intersection over Union (IoU)} 
IoU quantifies the overlap between two regions (predicted $A$ and ground-truth $B$) as:
\begin{equation}
\text{IoU}(\widehat{Y},Y) = \frac{|\widehat{Y} \cap Y|}{|\widehat{Y} \cup Y|},
\end{equation}
where $|\widehat{Y} \cap Y|$ is the area of intersection, and $|\widehat{Y} \cup Y|$ is the area of union.

\noindent\textbf{U-Score.} 
To address the limitation that existing evaluations primarily focus on segmentation performance while failing to balance key computational factors such as model size, inference cost, and speed—making it difficult to assess practical deployment capabilities—we construct the U-Score based on quantile statistics. A detailed description is provided in Appendix~\ref{app:uscore}.

\noindent\textbf{Normalized Discounted Cumulative Gain (NDCG)}

NDCG evaluates the "usefulness" of a ranked list by accounting for two key factors: (1) the \textit{relevance} of each item, and (2) the \textit{position} of relevant items (penalizing lower-ranked relevant items via discounting). It is normalized to a range of $[0,1]$ to enable cross-task comparisons.

First, the \textit{Discounted Cumulative Gain (DCG)} is defined to measure the cumulative relevance of a ranked list up to position $k$ (denoted as $\text{DCG}@k$):
\begin{align}
\text{DCG}@k = \sum_{i=1}^{k} \frac{\text{rel}_i}{\log_2(i+1)}
\end{align}
where:
\begin{itemize}[leftmargin=*]
    \item $k$: The cutoff position (e.g., $k=10$ for NDCG@10, focusing on top-10 results).
    \item $\text{rel}_i$: The \textit{relevance score} of the $i$-th item in the ranked list. For binary relevance (relevant/irrelevant).
    \item $\log_2(i+1)$: The discount factor, which reduces the contribution of items ranked later (since users are less likely to inspect lower positions).
\end{itemize}

To normalize DCG across different queries/tasks (where the maximum possible relevance varies), the \textit{Ideal DCG (IDCG)}—the maximum possible DCG@k for a given set of items—is computed by ranking all relevant items in descending order of $\text{rel}_i$:
\begin{align}
\text{IDCG}@k = \sum_{i=1}^{\min(k, |R|)} \frac{\text{rel}'_i}{\log_2(i+1)}
\end{align}
where:
\begin{itemize}[leftmargin=*]
\item $R$: The set of all relevant items for the query/task.
\item $|R|$: The total number of relevant items.
\item $\text{rel}'_i$: The $i$-th highest relevance score among all items in $R$ (i.e., the ideal ranking).
\end{itemize}

NDCG@k is defined as the ratio of DCG@k to IDCG@k. To avoid division by zero (when no relevant items exist, $\text{IDCG}@k=0$), NDCG@k is set to 0 in this edge case:
\begin{align}
\text{NDCG}@k = 
\begin{cases} 
0 & \text{if } \text{IDCG}@k = 0, \\
\frac{\text{DCG}@k}{\text{IDCG}@k} & \text{otherwise}.
\end{cases}
\end{align}

For experiments with multiple queries/tasks (e.g., a retrieval dataset with 1k queries), the \textit{mean NDCG@k}—the average of NDCG@k across all queries—is reported. In Table~\ref{tab:recommend_wrap}, NDCG@k values for $k=5$ and $k=20$ are provided.

\noindent\textbf{Mean Average Precision (MAP)}
% \label{app:metrics:map}

MAP quantifies the average precision of relevant items in a ranked list, aggregated across all queries/tasks. It is particularly useful for scenarios where "early relevant items" (high precision at top positions) are critical (e.g., information retrieval, recommendation systems).

First, \textit{Average Precision (AP)} for a single query $q$ is defined as the average of the precision of the ranked list at the position of each relevant item:
\begin{align}
\text{AP}(q) = \frac{1}{|R_q|} \sum_{r \in R_q} \text{Prec}(k_r)
\end{align}
where:
\begin{itemize}[leftmargin=*]
    \item $q$: A single query (or task instance) from the query set $Q$;
    \item $R_q$: The set of all relevant items for query $q$ (if $|R_q|=0$, $\text{AP}(q)=0$ by convention);
    \item $k_r$: The position of relevant item $r$ in the ranked list for $q$;
    \item $\text{Prec}(k_r)$: The precision at position $k_r$, defined as $\text{Prec}(k_r) = \frac{\text{numRel}(k_r)}{k_r}$, where $\text{numRel}(k_r)$ is the number of relevant items in the top-$k_r$ positions.
\end{itemize}

For a set of $|Q|$ queries, MAP is the average of AP scores across all queries:
\begin{align}
\text{MAP} = \frac{1}{|Q|} \sum_{q \in Q} \text{AP}(q)
\end{align}

Similar to NDCG, MAP ranges from $[0,1]$: a value of 1 indicates all relevant items are ranked first (perfect precision at every relevant position), while 0 indicates no relevant items are retrieved.

\noindent\textbf{Spearman's Rank Correlation Coefficient}
\label{app:metrics:spearman}

Spearman's rank correlation coefficient quantifies the \textit{monotonic relationship} between two ranked variables. It is particularly useful for evaluating how well the order of items (e.g., predicted rankings by a model and ground-truth rankings) aligns, making it relevant for tasks where the consistency of relative ordering matters (e.g., comparing ranked recommendations or human judgments).

Formally, Spearman's rank correlation coefficient $\rho$ between two variables $X$ (e.g., model-generated ranks) and $Y$ (e.g., ground-truth ranks) (each with $n$ paired observations) is defined as:
\begin{align}
\rho = 1 - \frac{6 \sum_{i=1}^{n} d_i^2}{n(n^2 - 1)}
\end{align}
where:
\begin{itemize}[leftmargin=*]
    \item $d_i$: The difference between the rank of $X_i$ and the rank of $Y_i$ (i.e., $d_i = \text{rank}(X_i) - \text{rank}(Y_i)$).
    \item $n$: The total number of paired observations.
\end{itemize}
\clearpage
\section{Additional Results \label{app:result}}

\noindent

\begin{table}[t!]
\centering

    \caption{Top-10 performing variants across each dataset on source domains. Baseline U-Net is highlighted (gray background), and statistical significance of p-value is highlighted: \colorbox{g}{p$<$0.0001}, \colorbox{lg}{p$<$0.001}, \colorbox{y}{p$<$0.05}, \colorbox{lr}{p$\leq$0.05}, and \colorbox{r}{P $>$ 0.05 (Not significant)}}
    \resizebox{\linewidth}{!}{
        \begin{tabular}{r | c c c c c c | c c c c | c c c c}
            \Xhline{1px}
            \multirow{2}{*}{Rank} & \multicolumn{6}{c|}{Ultrasound} & \multicolumn{4}{c|}{Endoscopy} & \multicolumn{4}{c}{Dermoscopy} \\
            &  & BUSI &  & BUSBRA &  & TNSCUI &  & Kvasir &  & Robotool & ISIC2018 &  & SkinCancer \\
            \hline
            \gold & RWKV-UNet & \cellcolor{g}{72.32} & RWKV-UNet & \cellcolor{g}{84.76} & RWKV-UNet & \cellcolor{g}{80.06} & Swin-umamba & \cellcolor{lg}{85.56} & MEGANet & \cellcolor{g}{86.25} & RWKV-UNet & \cellcolor{g}{84.97} & RWKV-UNet & \cellcolor{lr}{87.48} \\
            \silver & PraNet & \cellcolor{g}{71.63} & EViT-UNet & \cellcolor{lg}{84.36} & MEGANet & \cellcolor{g}{79.01} & VMUNet & \cellcolor{y}{84.90} & RWKV-UNet & \cellcolor{g}{85.95} & CFFormer & \cellcolor{g}{84.89} & DA-TransUNet & \cellcolor{lr}{87.22} \\
            \bronze & Mobile U-ViT & \cellcolor{g}{71.59} & CaraNet & \cellcolor{lg}{84.31} & MFMSNet & \cellcolor{g}{78.91} & UACANet & \cellcolor{y}{84.81} & AURA-Net & \cellcolor{lg}{85.78} & MEGANet & \cellcolor{g}{84.50} & MSLAU-Net & \cellcolor{r}{86.80} \\
            \#4 & DA-TransUNet & \cellcolor{g}{71.47} & MFMSNet & \cellcolor{y}{84.27} & TA-Net & \cellcolor{g}{78.85} & CFFormer & \cellcolor{lg}{84.57} & TA-Net & \cellcolor{lg}{85.62} &  Swin-umamba & \cellcolor{g}{84.49} & PraNet & \cellcolor{r}{86.38} \\
            \#5 & MEGANet & \cellcolor{g}{71.47} & TA-Net & \cellcolor{y}{84.17} & UACANet & \cellcolor{g}{78.83} & RWKV-UNet & \cellcolor{lg}{84.53} & EViT-UNet & \cellcolor{lg}{85.39} & PraNet & \cellcolor{g}{84.42} & FCBFormer & \cellcolor{r}{86.33} \\
            \#6 & TransResUNet & \cellcolor{g}{71.27} & FAT-Net & \cellcolor{lr}{84.15} & EViT-UNet & \cellcolor{g}{78.75} & FCBFormer & \cellcolor{lg}{84.50} & TransResUNet & \cellcolor{lg}{85.30} & TransResUNet & \cellcolor{g}{84.38} & EMCAD & \cellcolor{lr}{86.20} \\
            \#7 & MFMSNet & \cellcolor{g}{71.24} & UACANet & \cellcolor{y}{84.07} & CaraNet & \cellcolor{g}{78.69} & PraNet & \cellcolor{y}{84.39} & MFMSNet & \cellcolor{lg}{85.21} & TA-Net & \cellcolor{g}{84.34} & MCA-UNet & \cellcolor{r}{86.17} \\
            \#8 & CFFormer & \cellcolor{g}{70.91} & FCBFormer & \cellcolor{y}{84.00} & Swin-umamba & \cellcolor{g}{78.64} & CASCADE & \cellcolor{y}{84.34} & CE-Net & \cellcolor{lg}{85.11} & CE-Net & \cellcolor{g}{84.32} & CaraNet & \cellcolor{r}{86.01} \\
            \#9 & ESKNet & \cellcolor{g}{70.88} & MEGANet & \cellcolor{lr}{83.99} & FAT-Net & \cellcolor{g}{78.57} & CENet & \cellcolor{y}{84.32} & PraNet & \cellcolor{lg}{85.10} & CaraNet & \cellcolor{g}{84.26} & TransNorm & \cellcolor{r}{85.86} \\
            \#10 & CASCADE & \cellcolor{lg}{70.81} & AURA-Net & \cellcolor{lr}{83.95} & UTANet & \cellcolor{g}{78.51} & MFMSNet & \cellcolor{lg}{84.32} & UACANet & \cellcolor{y}{84.65} & AURA-Net & \cellcolor{g}{84.25} & AURA-Net & \cellcolor{r}{85.56} \\
            \rowcolor{gray!15} & U-Net (\#68) & 65.58 & U-Net (\#41) & 82.91 & U-Net (\#58) & 75.99 & U-Net (\#70) & 80.11 & U-Net (\#23) & 81.24 & U-Net (\#61) & 82.78 & U-Net (\#77) & 80.94 \\
            \Xhline{1px}
        \end{tabular}
    }
    \vspace{0.1cm}

    \resizebox{\linewidth}{!}{
        \begin{tabular}{r | c c c c c c | c c c c | c c c c}
        \Xhline{1px}
        \multirow{2}{*}{Rank} & \multicolumn{6}{c|}{X-Ray} & \multicolumn{4}{c|}{MRI} & \multicolumn{4}{c}{Fundus} \\
        &  & Covidquex &  & Montgomery &  & DCA &  & Promise &  & ACDC &  & CHASE &  & DRIVE \\
        \hline
        \gold & AURA-Net & \cellcolor{g}{70.85} & RWKV-UNet & \cellcolor{r}{96.21} & DA-TransUNet & \cellcolor{r}{64.90} & RWKV-UNet & \cellcolor{y}{87.56} & CENet & \cellcolor{y}{85.54} & CMU-Net & \cellcolor{r}{84.33} & FCBFormer & \cellcolor{g}{64.25} \\
        \silver & RWKV-UNet & \cellcolor{lg}{70.75} & DA-TransUNet & \cellcolor{lr}{96.17} & UTANet & \cellcolor{r}{64.23} & FCBFormer & \cellcolor{r}{87.29} & Swin-umambaD & \cellcolor{y}{85.45} & AttU-Net & \cellcolor{r}{84.20} & MT-UNet & \cellcolor{g}{63.21} \\
        \bronze & CaraNet & \cellcolor{y}{70.61} & MEGANet & \cellcolor{r}{96.12} & EViT-UNet & \cellcolor{r}{63.81} & MFMSNet & \cellcolor{r}{87.26} & DoubleUNet & \cellcolor{lr}{85.33} & U-Net & 84.07 & ColonSegNet & \cellcolor{lg}{63.19} \\
        \#4 & EViT-UNet & \cellcolor{y}{70.33} & TransAttUnet & \cellcolor{r}{96.03} & MFMSNet & \cellcolor{r}{63.81} & EViT-UNet & \cellcolor{r}{87.05} & RWKV-UNet & \cellcolor{lr}{85.20} & UNet3plus & \cellcolor{r}{83.69} & UTNet & \cellcolor{g}{63.17} \\
        \#5 & TA-Net & \cellcolor{y}{70.20} & RollingUnet & \cellcolor{r}{96.01} & MEGANet & \cellcolor{r}{63.77} & Perspective-Unet & \cellcolor{r}{87.03} & DDANet & \cellcolor{lr}{85.11} & Perspective-Unet & \cellcolor{r}{82.86} & ESKNet & \cellcolor{g}{63.15} \\
        \#6 & MEGANet & \cellcolor{y}{70.19} & DDANet & \cellcolor{r}{95.97} & ESKNet & \cellcolor{r}{63.69} & MEGANet & \cellcolor{r}{87.00} & AttU-Net & \cellcolor{lg}{85.01} & UCTransNet & \cellcolor{r}{82.82} & CMU-Net & \cellcolor{lg}{62.85} \\
        \#7 & PraNet & \cellcolor{lr}{70.09} & MT-UNet & \cellcolor{r}{95.90} & DDANet & \cellcolor{r}{63.65} & TransResUNet & \cellcolor{r}{86.95} & EViT-UNet & \cellcolor{r}{84.91} & ESKNet & \cellcolor{r}{82.69} & Swin-umamba & \cellcolor{lr}{62.75} \\
        \#8 & CE-Net & \cellcolor{lr}{70.04} & TransResUNet & \cellcolor{r}{95.89} & RWKV-UNet & \cellcolor{r}{63.61} & U-KAN & \cellcolor{r}{86.89} & FCBFormer & \cellcolor{r}{84.90} & ColonSegNet & \cellcolor{r}{82.20} & UNet3plus & \cellcolor{lr}{62.54} \\
        \#9 & TransResUNet & \cellcolor{lr}{69.81} & Mobile U-ViT & \cellcolor{r}{95.89} & UTNet & \cellcolor{r}{63.54} & PraNet & \cellcolor{r}{86.88} & G-CASCADE & \cellcolor{r}{84.89} & MT-UNet & \cellcolor{r}{82.00} & RollingUnet & \cellcolor{lr}{62.49} \\
        \#10 & MFMSNet & \cellcolor{lr}{69.77} & UNet3plus & \cellcolor{r}{95.88} & U-Net & 63.30 & CMU-Net & \cellcolor{r}{86.87} & MSRFNet & \cellcolor{r}{84.78} & Swin-umamba & \cellcolor{r}{81.65} & D-TrAttUnet & \cellcolor{y}{62.48} \\
        \rowcolor{gray!15} & U-Net (\#31) & 68.52 & U-Net (\#11) & 95.87 & & & U-Net (\#29) & 86.30 & U-Net (\#23) & 84.32 & & &  U-Net (\#14) & 61.81 \\
        \hline
        \Xhline{1px}
        \end{tabular}
    }

    \vspace{0.1cm}

    \resizebox{\linewidth}{!}{
        \begin{tabular}{r | c c c c | c c c c | c c | c c}
        \Xhline{1px}
        \multirow{2}{*}{Rank} & \multicolumn{4}{c|}{Histopathology} & \multicolumn{4}{c|}{Nuclear} & \multicolumn{2}{c|}{CT} & \multicolumn{2}{c}{OCT} \\
        &  & Glas &  & Monusac &  & DSB2018 &  & CellNuclear  &  & Synapse &  & Cystoidfluid \\
        \hline
        \gold  & EMCAD & \cellcolor{y}{85.85} & MT-UNet & \cellcolor{lg}{69.27} & MT-UNet & \cellcolor{y}{88.74} & MT-UNet & \cellcolor{r}{84.93}  & CENet & \cellcolor{y}{74.70} & UNet3plus & \cellcolor{g}{85.76} \\
        \silver & RWKV-UNet & \cellcolor{lg}{85.75} & RWKV-UNet & \cellcolor{lr}{68.96} & DoubleUNet & \cellcolor{lr}{88.61} & TransAttUnet & \cellcolor{r}{84.88} & Perspective-Unet & \cellcolor{lr}{73.69} & Swin-umamba & \cellcolor{lg}{85.06} \\
        \bronze & CASCADE & \cellcolor{lr}{85.17} & UTANet & \cellcolor{lr}{68.39} & TransAttUnet & \cellcolor{lr}{88.49} & AURA-Net & \cellcolor{r}{84.87} & G-CASCADE & \cellcolor{lr}{73.54} & UTANet & \cellcolor{lg}{84.89} \\
        \#4 & MSLAU-Net & \cellcolor{r}{84.38} & CA-Net & \cellcolor{r}{68.39} & DCSAU-Net & \cellcolor{lr}{88.44}  & CA-Net & \cellcolor{r}{84.85} & CASCADE & \cellcolor{lr}{73.30} & MMUNet & \cellcolor{r}{84.21} \\
        \#5 & UTANet & \cellcolor{r}{84.22} & DDANet & \cellcolor{y}{68.38} & UTNet & \cellcolor{r}{88.39} & UTANet & \cellcolor{r}{84.77}  & AURA-Net & \cellcolor{lr}{73.25} & H2Former & \cellcolor{lr}{83.99} \\
        \#6 & DDANet & \cellcolor{r}{83.78} & TransAttUnet & \cellcolor{lr}{68.25} & D-TrAttUnet & \cellcolor{r}{88.27} & ColonSegNet & \cellcolor{r}{84.70}  & MEGANet & \cellcolor{r}{73.18} & Perspective-Unet & \cellcolor{r}{83.90} \\
        \#7 & MERIT & \cellcolor{r}{83.77} & UTNet & \cellcolor{r}{67.76} & ESKNet & \cellcolor{r}{88.23} &  DA-TransUNet & \cellcolor{r}{84.63} & DS-TransUNet & \cellcolor{r}{72.74} & FCBFormer & \cellcolor{lr}{83.84} \\
        \#8 & MBSNet & \cellcolor{r}{83.57} & EViT-UNet & \cellcolor{r}{67.61} & AURA-Net & \cellcolor{r}{88.23} & RollingUnet & \cellcolor{r}{84.62} & DoubleUNet & \cellcolor{y}{72.63}  & D-TrAttUnet & \cellcolor{r}{83.74} \\
        \#9 & CENet & \cellcolor{r}{83.47} & D-TrAttUnet & \cellcolor{r}{66.96} & LGMSNet & \cellcolor{r}{88.16} & RWKV-UNet & \cellcolor{r}{84.56}  & MSLAU-Net & \cellcolor{y}{72.60} & MedFormer & \cellcolor{r}{83.58} \\
        \#10 & U-Net & 83.30 & AttU-Net & \cellcolor{r}{66.96}  & DDANet & \cellcolor{r}{88.16} & FCBFormer & \cellcolor{r}{84.54}  & RWKV-UNet & \cellcolor{r}{72.56} & EViT-UNet & \cellcolor{r}{83.54} \\
        \rowcolor{gray!15} & & & U-Net (\#21) & 66.44  & U-Net (\#16) & 88.05 & U-Net (\#17) & 84.33  & U-Net (\#52) & 67.90  & U-Net (\#23) & 82.39 \\
        \Xhline{1px}
        \end{tabular}
    }
\label{tab:top10source}    
\end{table}

\renewcommand{\multirowsetup}{\centering} 
\begin{table}[t!]
  \centering
  \caption{Top-10 performing variants across each dataset on target domains. Baseline U-Net is highlighted (gray background), and statistical significance of p-value is highlighted: \colorbox{g}{p$<$0.0001}, \colorbox{lg}{p$<$0.001}, \colorbox{y}{p$<$0.05}, \colorbox{lr}{p$\leq$0.05}, and \colorbox{r}{P $>$ 0.05 (Not significant)}}
  \label{tab:top10zeroshot}
  % 第1行（两块并排）：
  \noindent
\begin{minipage}[t]{0.58\textwidth}
    \centering
    \resizebox{\linewidth}{!}{
        \begin{tabular}{r | c c c c c c}
            \Xhline{1px}
            \multirow{2}{*}{Rank} & \multicolumn{6}{c}{Ultrasound (Source $\rightarrow$ Target)} \\
            &  \multicolumn{2}{c}{BUSI $\rightarrow$ BUS} & \multicolumn{2}{c}{BUSBRA $\rightarrow$ BUS} & \multicolumn{2}{c}{TNSCUI $\rightarrow$ TUCC} \\
            \hline
            \gold & Swin-umamba & \cellcolor{g}{82.91} & MEGANet & \cellcolor{g}{86.62} & MSLAU-Net & \cellcolor{g}{66.15} \\
            \silver & EMCAD & \cellcolor{g}{82.83} & DoubleUNet & \cellcolor{g}{86.51} & MERIT & \cellcolor{g}{66.00} \\
            \bronze & CENet & \cellcolor{g}{82.70} & CENet & \cellcolor{g}{86.29} & EViT-UNet & \cellcolor{g}{65.83} \\
            \#4 & G-CASCADE & \cellcolor{g}{82.61} & FCBFormer & \cellcolor{g}{86.10} & Polyp-PVT & \cellcolor{g}{65.23} \\
            \#5 & DA-TransUNet & \cellcolor{g}{82.32} & CASCADE & \cellcolor{g}{85.96} & LGMSNet & \cellcolor{g}{65.16} \\
            \#6 & PraNet & \cellcolor{g}{82.13} & Polyp-PVT & \cellcolor{g}{85.65} & G-CASCADE & \cellcolor{g}{65.10} \\
            \#7 & CASCADE & \cellcolor{g}{82.11} & TransResUNet & \cellcolor{g}{85.62} & CaraNet & \cellcolor{g}{64.61} \\
            \#8 & TransNorm & \cellcolor{g}{82.11} & ResNet34UnetPlus & \cellcolor{g}{85.46} & H2Former & \cellcolor{g}{64.48} \\
            \#9 & MCA-UNet & \cellcolor{g}{81.88} & MCA-UNet & \cellcolor{g}{85.28} & MEGANet & \cellcolor{g}{64.37} \\
            \#10 & CaraNet & \cellcolor{g}{81.32} & G-CASCADE & \cellcolor{g}{85.27} & Swin-umamba & \cellcolor{g}{64.00} \\
            \rowcolor{gray!15} & U-Net (\#63) & 72.44 & U-Net (\# 79) & 81.37 & U-Net (\# 65) & 60.50 \\
            \Xhline{1px}
        \end{tabular}
    }
\end{minipage}
\hfill
\begin{minipage}[t]{0.398\textwidth}
    \centering
    \resizebox{\linewidth}{!}
        {
        \begin{tabular}{r | c c c c}
        \Xhline{1px}
        \multirow{2}{*}{Rank} & \multicolumn{4}{c}{Endoscopy (Source $\rightarrow$ Target)} \\
        &  \multicolumn{2}{c}{Kvasir $\rightarrow$ CVC300} & \multicolumn{2}{c}{Kvasir $\rightarrow$ CVC-ClinicDB} \\
        \hline
        \gold & PraNet & \cellcolor{y}{83.31} & PraNet & \cellcolor{y}{77.39} \\
        \silver & RWKV-UNet & \cellcolor{y}{82.14} & DS-TransUNet & \cellcolor{y}{77.38} \\
        \bronze & UACANet & \cellcolor{y}{81.72} & CASCADE & \cellcolor{lg}{77.19} \\
        \#4 & MERIT & \cellcolor{y}{81.39} & Swin-umambaD & \cellcolor{y}{76.83} \\
        \#5 & MFMSNet & \cellcolor{y}{81.24} & EMCAD & \cellcolor{lr}{76.56} \\
        \#6 & TA-Net & \cellcolor{y}{81.24} & TransResUNet & \cellcolor{y}{76.42} \\
        \#7 & EViT-UNet & \cellcolor{y}{81.11} & MFMSNet & \cellcolor{y}{76.33} \\
        \#8 & UTANet & \cellcolor{y}{80.78} & CFFormer & \cellcolor{lr}{76.18} \\
        \#9 & DS-TransUNet & \cellcolor{y}{80.58} & DoubleUNet & \cellcolor{lr}{75.72} \\
        \#10 & CASCADE & \cellcolor{y}{80.57} & MEGANet & \cellcolor{lr}{75.64} \\
        \rowcolor{gray!15} & U-Net (\#75) & 70.33 & U-Net (\#47) & 69.87 \\
        \Xhline{1px}
        \end{tabular}
        }
\end{minipage}%
  \par\medskip
  \noindent
\begin{minipage}[t]{0.225\textwidth}
    \centering
    % \caption{Zero-shot top10 on Dermoscopy}
    \resizebox{\linewidth}{!}{
        \begin{tabular}{r | c c}
            \Xhline{1px}
            \multirow{2}{*}{Rank} & \multicolumn{2}{c}{Dermoscopy} \\
            &  \multicolumn{2}{c}{ISIC2018 $\rightarrow$ PH2} \\
            \hline
            \gold & MSLAU-Net & \cellcolor{lg}{86.52} \\
            \silver & RWKV-UNet & \cellcolor{y}{86.00} \\
            \bronze & G-CASCADE & \cellcolor{y}{85.96} \\
            \#4 & MERIT & \cellcolor{y}{85.94} \\
            \#5 & MMUNet & \cellcolor{lr}{85.66} \\
            \#6 & H2Former & \cellcolor{lr}{85.39} \\
            \#7 & CMUNeXt & \cellcolor{lr}{85.32} \\
            \#8 & UACANet & \cellcolor{r}{85.12} \\
            \#9 & MFMSNet & \cellcolor{lr}{85.08} \\
            \#10 & MCA-UNet & \cellcolor{r}{85.08} \\
            \rowcolor{gray!15} & U-Net (\#47) & 84.00 \\
            \Xhline{1px}
        \end{tabular}
    }
\end{minipage}%
\hfill
\begin{minipage}[t]{0.235\textwidth}
    \centering
    % \caption{Zero-shot top10 on Fundus}
    \resizebox{\linewidth}{!}{
        \begin{tabular}{r | c c}
            \Xhline{1px}
            \multirow{2}{*}{Rank} & \multicolumn{2}{c}{Fundus} \\
            &  \multicolumn{2}{c}{CHASE $\rightarrow$ DRIVE} \\
            \hline
            \gold & MSRFNet & \cellcolor{g}{55.60} \\
            \silver & DS-TransUNet & \cellcolor{g}{55.52} \\
            \bronze & MBSNet & \cellcolor{g}{54.66} \\
            \#4 & RWKV-UNet & \cellcolor{g}{54.27} \\
            \#5 & CENet & \cellcolor{g}{53.78} \\
            \#6 & CSCAUNet & \cellcolor{g}{52.80} \\
            \#7 & MCA-UNet & \cellcolor{g}{52.69} \\
            \#8 & EViT-UNet & \cellcolor{g}{52.49} \\
            \#9 & Tinyunet & \cellcolor{g}{52.26} \\
            \#10 & TransResUNet & \cellcolor{g}{52.00} \\
            \rowcolor{gray!15} & U-Net (\#57) & 39.64 \\
            \Xhline{1px}
        \end{tabular}
    }
\end{minipage}%
\hfill
\begin{minipage}[t]{0.251\textwidth}
    \centering
    % \caption{Zero-shot top10 on X-Ray}
    \resizebox{\linewidth}{!}{
        \begin{tabular}{r | c c}
            \Xhline{1px}
            \multirow{2}{*}{Rank} & \multicolumn{2}{c}{X-Ray} \\
            & \multicolumn{2}{c}{Montgomery $\rightarrow$ NIH-test} \\
            \hline
            \gold & MEGANet & \cellcolor{g}{88.19} \\
            \silver & TransResUNet & \cellcolor{g}{87.69} \\
            \bronze & CaraNet & \cellcolor{g}{86.22} \\
            \#4 & DA-TransUNet & \cellcolor{g}{85.87} \\
            \#5 & PraNet & \cellcolor{g}{84.58} \\
            \#6 & MFMSNet & \cellcolor{g}{83.67} \\
            \#7 & Swin-umambaD & \cellcolor{g}{83.13} \\
            \#8 & TransUnet & \cellcolor{g}{83.03} \\
            \#9 & TransNorm & \cellcolor{g}{82.90} \\
            \#10 & RWKV-UNet & \cellcolor{g}{82.41} \\
            \rowcolor{gray!15} & U-Net (\#51) & 71.33 \\
            \Xhline{1px}
        \end{tabular}
    }
\end{minipage}
\hfill
\begin{minipage}[t]{0.265\textwidth}
    \centering
    % \caption{Zero-shot top10 on Histopathology}
    \resizebox{\linewidth}{!}{
        \begin{tabular}{r | c c}
            \Xhline{1px}
            \multirow{2}{*}{Rank} & \multicolumn{2}{c}{Histopathology} \\
            &  \multicolumn{2}{c}{Monusac $\rightarrow$ Tnbcnuclei} \\
            \hline
            \gold & TA-Net & \cellcolor{lg}{50.74} \\
            \silver & CENet & \cellcolor{lg}{48.03} \\
            \bronze & ResNet34UnetPlus & \cellcolor{lg}{46.44} \\
            \#4 & EMCAD & \cellcolor{lg}{46.41} \\
            \#5 & G-CASCADE & \cellcolor{y}{46.18} \\
            \#6 & UNetV2 & \cellcolor{lr}{45.32} \\
            \#7 & CSWin-UNet & \cellcolor{lr}{45.18} \\
            \#8 & DAEFormer & \cellcolor{lr}{45.16} \\
            \#9 & DA-TransUNet & \cellcolor{lg}{44.53} \\
            \#10 & MedVKAN & \cellcolor{lg}{44.42} \\
            \rowcolor{gray!15} & U-Net (\#91) & 26.05 \\
            \Xhline{1px}
        \end{tabular}
    }
\end{minipage}
\end{table}

\subsection{Per-Dataset Top-10 and U-Net Comparison}

We report the top-10 performing methods across each dataset, evaluated on both source and target domains. As shown in Tab.~\ref{tab:top10source} and Tab.~\ref{tab:top10zeroshot}. For reference, the position of the vanilla U-Net is highlighted with a gray background, and we also compute the statistical significance of each variant relative to U-Net.

\noindent\textbf{Top10 performance on Source domain.} On widely studied datasets and modalities-such as ultrasound, polyp segmentation, ISIC2018 (Dermoscopy), Synapse (CT), Drive (Fundus), ACDC (MRI), and Covidquex (X-ray)-most top-10 variants achieve significant improvements over U-Net. This trend is consistent with the increasing popularity of these datasets and 'novelity desion' for long-range dependency modeling, such as incorporating Transformers, Mamba, RWKV, and hybrid designs. In contrast, on other datasets and modalities the improvements remain marginal. For example, in Montgomery (X-ray lung segmentation), DCA, Chase (Fundus), nuclear segmentation, and Histopathology, the relative gains over U-Net are not significant. This suggests that progress in these modalities has been limited, because they rely on stable local patterns rather than long-range context. These observations highlight an important direction for future research: designing models that are modality-aware, particularly tailored for domains dominated by local and repetitive structures.

\noindent\textbf{Top10 performance on target domain.} On the target-domain datasets, nearly all top-10 methods achieve substantial improvements, highlighting the superior generalization ability of recent variants. These gains are primarily driven by two factors: the adoption of long-range dependency modeling and the increased model complexity. Together, these characteristics enhance the representational capacity and adaptability of the variants, which aligns with the prevailing trend toward more novel and increasingly complex model architectures.

\begin{figure}[!]
    \centering
    \includegraphics[width=0.8\linewidth]{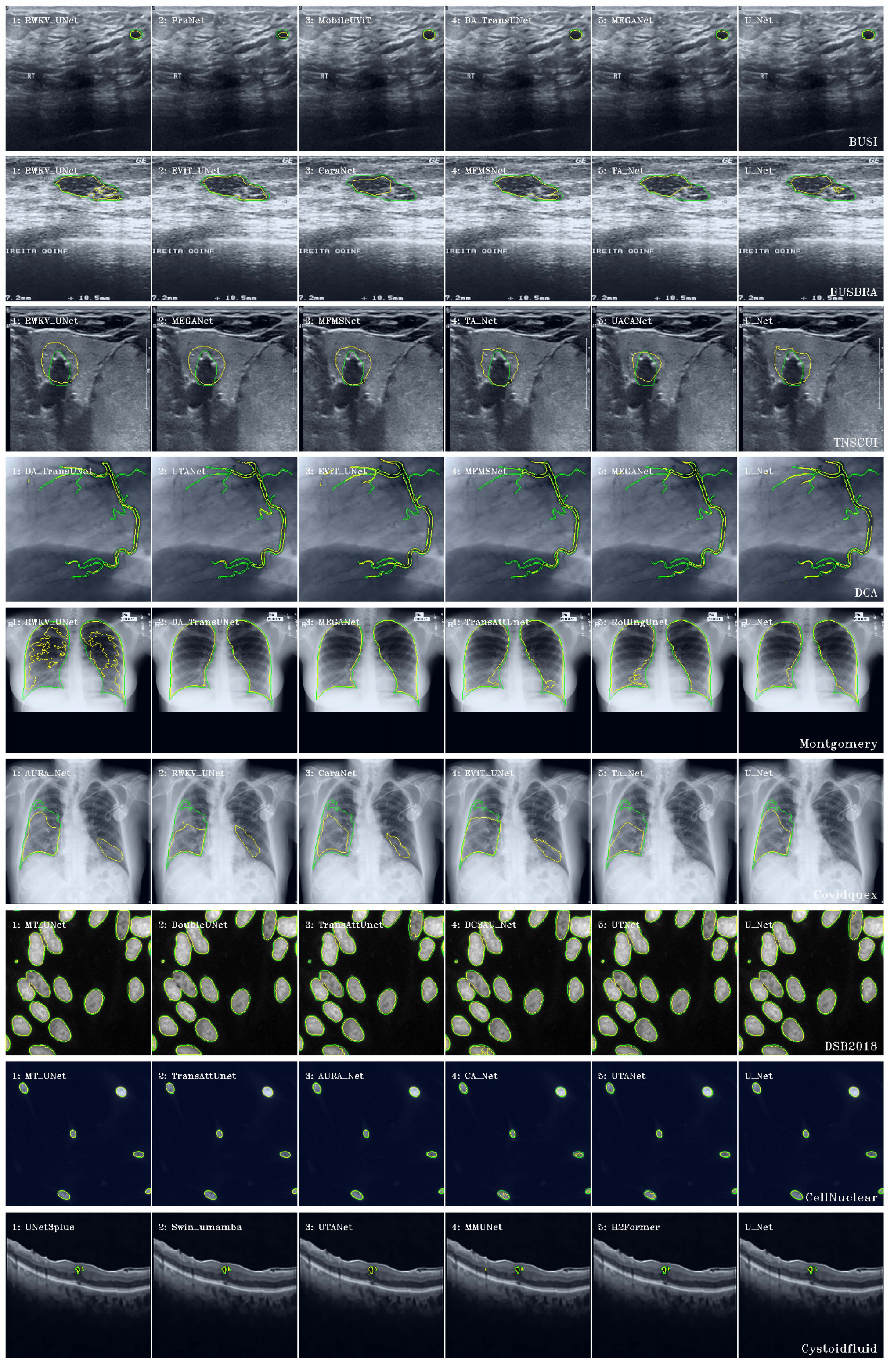}  
    \caption{Segmentation results of the Top 5 models and U-Net, where the green curve represents the ground truth and the yellow curve represents the model prediction.}
    \label{fig:top5visual1}
\end{figure}

\begin{figure}[!] 
    \centering 
    \includegraphics[width=0.8\linewidth]{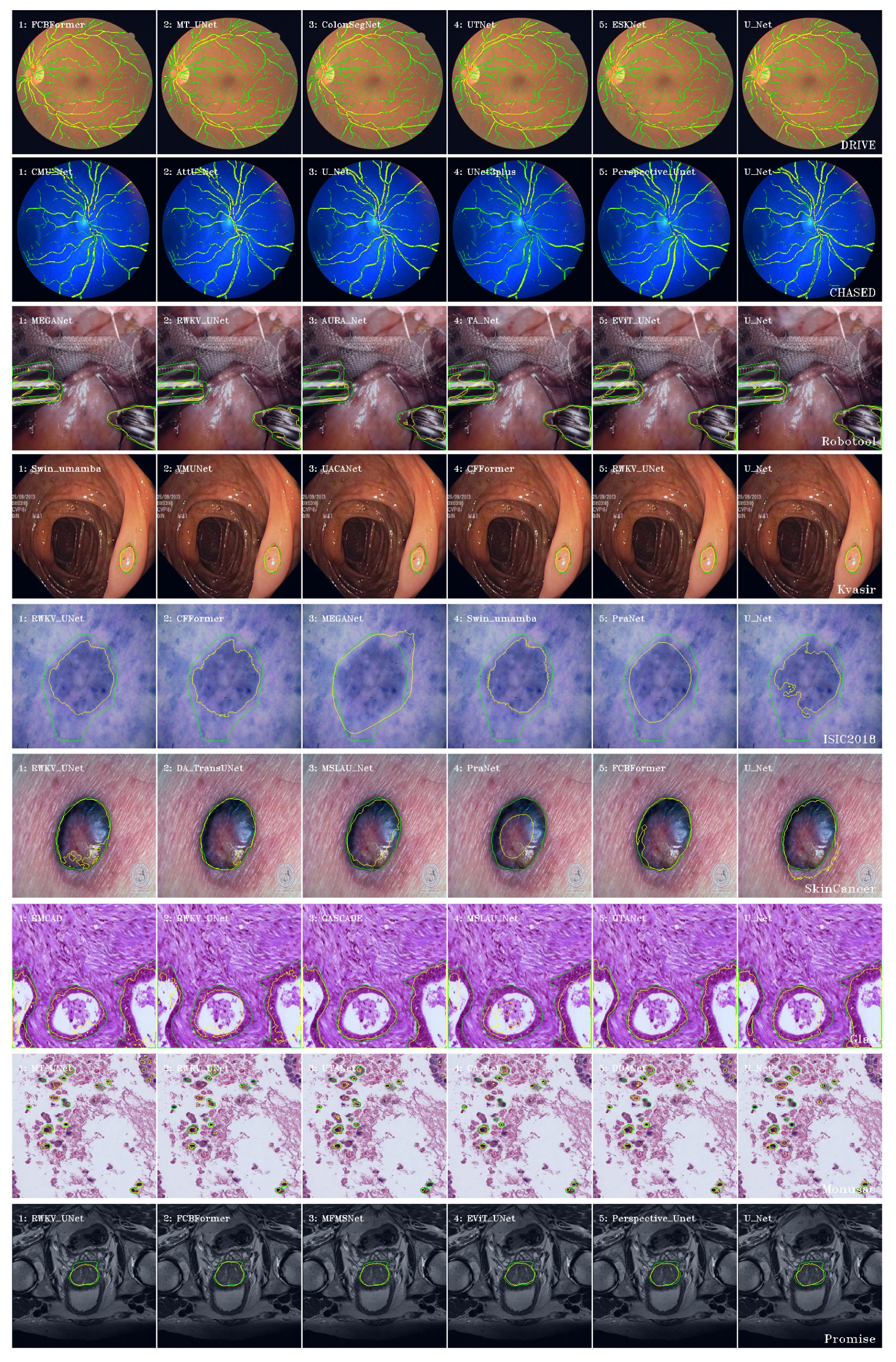}  
    \caption{Segmentation results of the Top 5 models and U-Net, where the green curve represents the ground truth and the yellow curve represents the model prediction.} 
    \label{fig:top5visual2} 
\end{figure}

In addition, we provide the visualization results of the top 5 models and U-net of the dataset for visualization analysis. The results are shown in Fig.~\ref{fig:top5visual1} and~\ref{fig:top5visual2}.

\clearpage
\section{Incorporate New Datasets and Algorithms in U-Stone}\label{app:reproduce}

\begin{figure}[htbp]
    \centering 
    \includegraphics[width=1\linewidth]{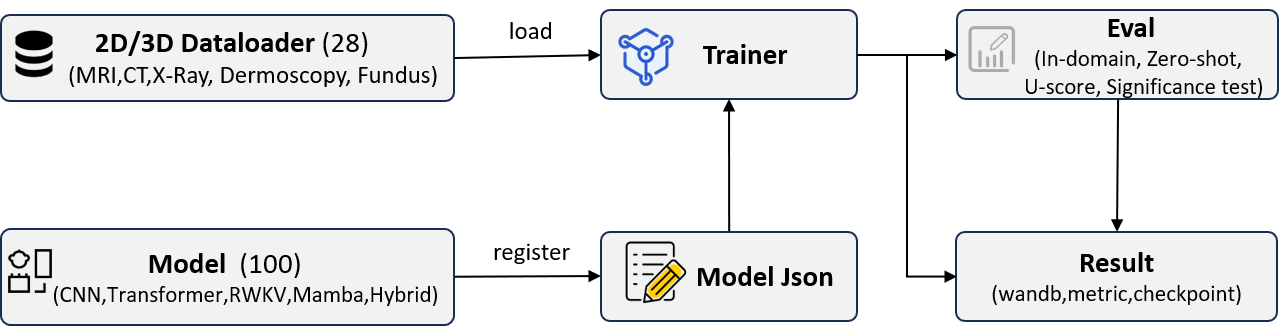}  
    \caption{Overall workflow of U-Stone. The Dataloader supports multiple medical imaging modalities (e.g., MRI, CT, X-Ray, Dermoscopy, Fundus). Models (with diverse architectures like CNN, Transformer, RWKV, Mamba, Hybrid) are registered and used by the Trainer for 2D$/$3D slice training. Evaluation covers In-domain$/$Zero-shot tasks, with results logged via tools like wandb.}
    \label{fig:system_arch} 
\end{figure}

We implement U-Stone using the PyTorch framework. Figure~\ref{fig:system_arch} illustrates the comprehensive workflow of U-Stone, a system tailored for medical image analysis. It features a versatile 2D/3D Dataloader that seamlessly accommodates multiple medical imaging modalities, including MRI, CT, X-Ray, Dermoscopy, and Fundus and so on. A rich assortment of models with diverse architectural designs-spanning CNN, Transformer, RWKV, Mamba, and Hybrid-are registered via a Model JSON configuration and then leveraged by the Trainer module for 2D/3D slice-wise training. The evaluation pipeline encompasses in-domain testing, zero-shot inference, statistical significance assessments and custom assessments of U-score.  Finally, results are systematically logged and visualized using tools such as Weight \& Biases (wandb), ensuring thorough tracking of metrics and checkpoints.

We demonstrate how to integrate new datasets and algorithms through example pseudocode Figure~\ref{fig:code}.

\subsection{Adding a New Dataset}
If the existing \texttt{Dataset} classes cannot meet your processing requirements, you can implement your own dataset with the structure shown as in Figure~\ref{fig:code} (a).

Additionally, you need to add your dataset name and loading method in the Dataloader file, as shown in the Figure~\ref{fig:code} (b).

\subsection{Adding a New Algorithm}
\begin{figure}[htbp] 
    \centering 
    \includegraphics[width=1\linewidth]{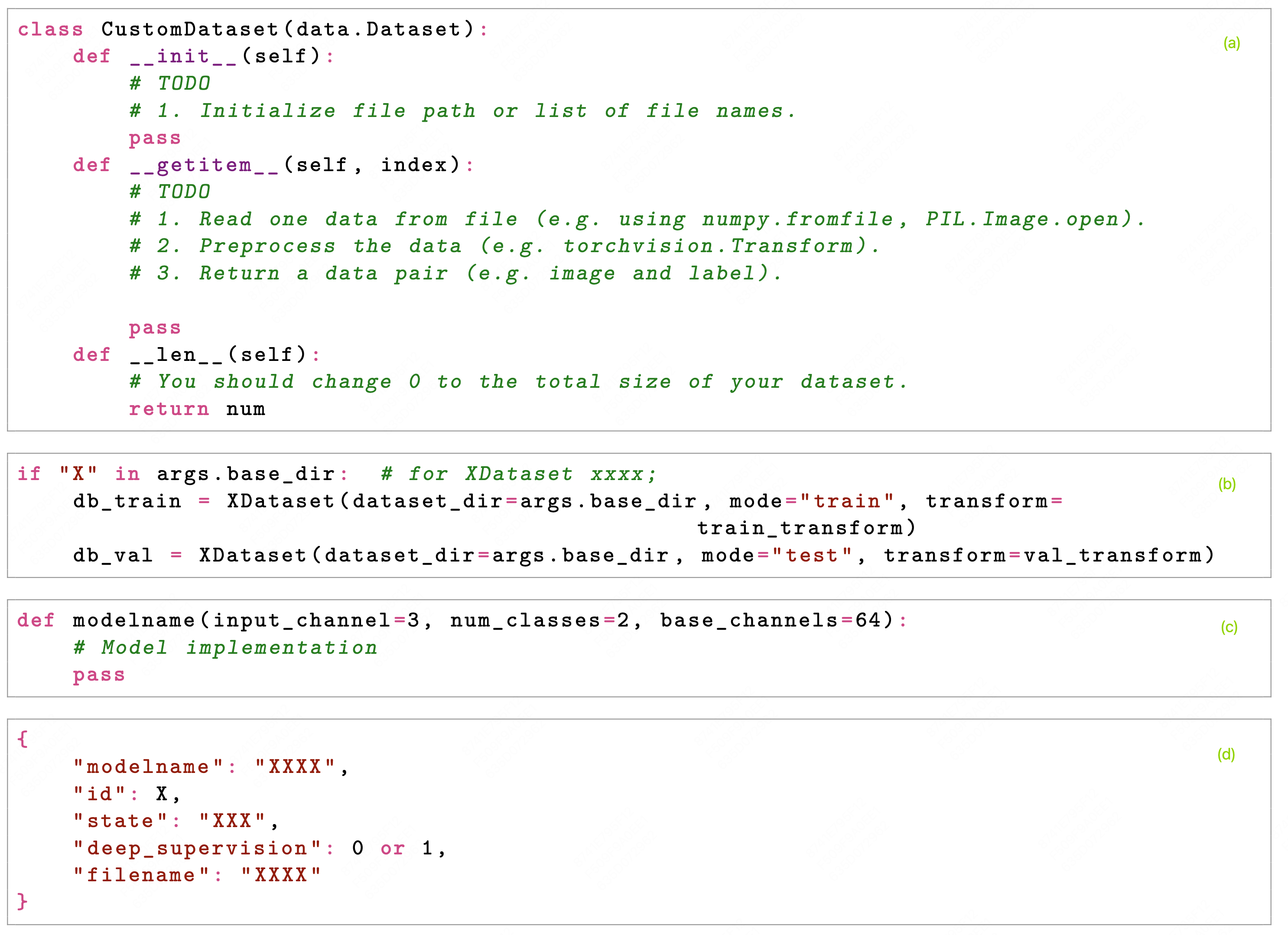}  
    \caption{Pseudocode display. (a) Pseudocode for datasets; (b) Pseudocode for data loading; (c) Pseudocode for model definition and input parameters; (d) Example of model registration using a JSON file}
    \label{fig:code}
\end{figure}

1. First, define your model in the \texttt{models} directory, ensuring the first two parameters are \texttt{input\_channel} and \texttt{num\_classes} to adapt to our project (as shown in Figure~\ref{fig:code} (c)).

2. Then, properly import your "modelname" in the \texttt{\_\_init\_\_.py} file under the \texttt{models} directory.

3. Finally, register your model in the \texttt{model\_id.json} file with the format shown in Figure~\ref{fig:code} (d). Note that \texttt{modelname}, \texttt{id}, and \texttt{deep\_supervision} are required fields, and \texttt{modelname} serves as the unique identifier for the model.

\clearpage 

\renewcommand{\multirowsetup}{\centering}  
\begin{table*}[!t]
\centering

\caption{Average performance of 100 u-shape medical image segmentation networks with IoU. Baseline U-Net is highlighted (gray background), and statistical significance of p-value is highlighted: \colorbox{g}{p$<$0.0001}, \colorbox{lg}{p$<$0.001}, \colorbox{y}{p$<$0.05}, \colorbox{lr}{p$\leq$0.05}, and \colorbox{r}{P $>$ 0.05 (Not significant)}}
\resizebox{1\linewidth}{!}
{
\begin{tabular}{r c | c c c | c c | c c | c c | c c c | c | c c c | c c | c | c | c}
\Xhline{1.5px} % horizontal line

\multirow{2}{*}{Rank} & \multirow{2}{*}{Network} 
& \multicolumn{3}{c|}{Ultrasound} & \multicolumn{2}{c|}{Dermoscopy} & \multicolumn{2}{c|}{Endoscopy} & \multicolumn{2}{c|}{Fundus} & \multicolumn{3}{c|}{Histopathology} & \multicolumn{1}{c|}{Nuclear} & \multicolumn{3}{c|}{X-Ray} & \multicolumn{2}{c|}{MRI} & \multicolumn{1}{c|}{CT} & \multicolumn{1}{c|}{OCT}   & \multirow{2}{*}{Avg} \\
% \cline{3-9}
% \cline{3-22}
& & BUSI & BUSBRA & TNSCUI & ISIC2018 & SkinCancer & Kvasir & Robotool & CHASE & DRIVE & DSB2018 & Glas & Monusac & Cell & Covidquex & Montgomery & DCA & ACDC & Promise & Synapse & Cystoidfluid \\

\hline

\gold & RWKV-UNet & \cellcolor{g}{72.32} & \cellcolor{g}{84.76} & \cellcolor{g}{80.06} & \cellcolor{g}{84.97} & \cellcolor{lr}{87.48} & \cellcolor{lg}{84.53} & \cellcolor{g}{85.95} & \cellcolor{r}{70.85} & \cellcolor{r}{59.75} & \cellcolor{r}{88.10} & \cellcolor{lg}{85.75} & \cellcolor{lr}{68.96} & \cellcolor{r}{84.56} & \cellcolor{lg}{70.75} & \cellcolor{r}{96.21} & \cellcolor{r}{63.61} & \cellcolor{lr}{85.20} & \cellcolor{y}{87.56} & \cellcolor{r}{72.56} & \cellcolor{r}{72.56} & 79.84 \\
\silver & UTANet & \cellcolor{y}{69.00} & \cellcolor{y}{83.93} & \cellcolor{g}{78.51} & \cellcolor{lg}{83.84} & \cellcolor{r}{84.48} & \cellcolor{lr}{83.06} & \cellcolor{y}{84.62} & \cellcolor{r}{79.81} & \cellcolor{r}{60.86} & \cellcolor{r}{87.70} & \cellcolor{r}{84.22} & \cellcolor{lr}{68.39} & \cellcolor{r}{84.77} & \cellcolor{r}{69.29} & \cellcolor{r}{95.76} & \cellcolor{r}{64.23} & \cellcolor{r}{84.73} & \cellcolor{r}{86.87} & \cellcolor{r}{69.73} & \cellcolor{lg}{69.73} & 79.43 \\
\end{tabular}
}
\end{table*}
\renewcommand{\multirowsetup}{\centering}  
\begin{table*}[!t]
\centering

\caption{Average performance of 100 u-shape medical image segmentation networks with U-Score. Baseline U-Net is highlighted (gray background), and statistical significance (calculated by IoU) of p-value is highlighted: \colorbox{g}{p$<$0.0001}, \colorbox{lg}{p$<$0.001}, \colorbox{y}{p$<$0.01}, \colorbox{lr}{p$\leq$0.05},and \colorbox{r}{P $>$ 0.05 (Not significant)}}
\resizebox{1\linewidth}{!}
{
\begin{tabular}{r c | c c c | c c | c c | c c | c c c | c | c c c | c c | c | c | c}
\Xhline{1.5px} % horizontal line

\multirow{2}{*}{Rank} & \multirow{2}{*}{Network} 
& \multicolumn{3}{c|}{Ultrasound} & \multicolumn{2}{c|}{Dermoscopy} & \multicolumn{2}{c|}{Endoscopy} & \multicolumn{2}{c|}{Fundus} & \multicolumn{3}{c|}{Histopathology} & \multicolumn{1}{c|}{Nuclear} & \multicolumn{3}{c|}{X-Ray} & \multicolumn{2}{c|}{MRI} & \multicolumn{1}{c|}{CT} & \multicolumn{1}{c|}{OCT}   & \multirow{2}{*}{Avg} \\
% \cline{3-9}
% \cline{3-22}
& & BUSI & BUSBRA & TNSCUI & ISIC2018 & SkinCancer & Kvasir & Robotool & CHASE & DRIVE & DSB2018 & Glas & Monusac & Cell & Covidquex & Montgomery & DCA & ACDC & Promise & Synapse & Cystoidfluid \\

\hline

\gold & LGMSNet & \cellcolor{y}{86.60} & \cellcolor{r}{85.79} & \cellcolor{g}{88.20} & \cellcolor{y}{83.21} & \cellcolor{r}{89.72} & \cellcolor{r}{77.54} & \cellcolor{r}{69.33} & \cellcolor{r}{86.33} & \cellcolor{r}{88.65} & \cellcolor{r}{90.80} & \cellcolor{r}{88.31} & \cellcolor{r}{84.05} & \cellcolor{r}{89.78} & \cellcolor{r}{80.45} & \cellcolor{r}{87.70} & \cellcolor{r}{86.33} & \cellcolor{r}{84.55} & \cellcolor{r}{86.55} & \cellcolor{r}{83.36} & \cellcolor{r}{82.59} & 84.99 \\
\silver & LV-UNet & \cellcolor{r}{78.85} & \cellcolor{r}{82.62} & \cellcolor{r}{85.36} & \cellcolor{lr}{87.93} & \cellcolor{r}{83.00} & \cellcolor{r}{84.97} & \cellcolor{r}{83.02} & \cellcolor{r}{68.19} & \cellcolor{r}{79.63} & \cellcolor{r}{71.97} & \cellcolor{r}{87.64} & \cellcolor{r}{82.50} & \cellcolor{r}{82.15} & \cellcolor{r}{91.70} & \cellcolor{r}{79.68} & \cellcolor{r}{90.90} & \cellcolor{r}{78.00} & \cellcolor{r}{86.67} & \cellcolor{r}{69.73} & \cellcolor{r}{84.68} & 81.96 \\
\end{tabular}
}
\end{table*}
\renewcommand{\multirowsetup}{\centering}  
\begin{table*}[!t]
\centering

\caption{Average zeroshot performance of 100 u-shape medical image segmentation networks with IoU. Source $\to$ Target. Baseline U-Net is highlighted (gray background), and statistical significance of p-value is highlighted: \colorbox{g}{p$<$0.0001}, \colorbox{lg}{p$<$0.001}, \colorbox{y}{p$<$0.01}, \colorbox{lr}{p$\leq$0.05},and \colorbox{r}{P $>$ 0.05 (Not significant)}}
\resizebox{0.93\linewidth}{!}
{
\begin{tabular}{r c | c c c | c c | c | c | c | c | c}
\Xhline{1.5px} % horizontal line

\multirow{2}{*}{Rank} & \multirow{2}{*}{Network} 
& \multicolumn{3}{c|}{Ultrasound} & \multicolumn{2}{c|}{Endoscopy} & \multicolumn{1}{c|}{Dermoscopy} & \multicolumn{1}{c|}{Fundus} & \multicolumn{1}{c|}{X-Ray} & \multicolumn{1}{c|}{Histopathology} & \multirow{2}{*}{Avg} \\
& & BUSI $\rightarrow$ BUS & BUSBRA $\rightarrow$ BUS & TNSCUI $\rightarrow$ TUCC & Kvasir $\rightarrow$ CVC300 & Kvasir $\rightarrow$ CVC-ClinicDB & ISIC2018 $\rightarrow$ PH2 & CHASE $\rightarrow$ Stare & Montgomery $\rightarrow$ NIH-test & Monusac $\rightarrow$ Tnbcnuclei & \\
\hline
\gold & RWKV-UNet & \cellcolor{g}{80.73} & \cellcolor{y}{84.73} & \cellcolor{g}{63.42} & \cellcolor{y}{82.14} & \cellcolor{y}{75.50} & \cellcolor{y}{86.00} & \cellcolor{g}{61.29} & \cellcolor{g}{82.41} & \cellcolor{y}{38.96} & 70.94 \\
\silver & DS-TransUNet & \cellcolor{lg}{78.70} & \cellcolor{y}{84.44} & \cellcolor{g}{61.95} & \cellcolor{y}{80.58} & \cellcolor{y}{77.38} & \cellcolor{r}{84.05} & \cellcolor{g}{57.65} & \cellcolor{g}{76.21} & \cellcolor{y}{44.20} & 70.07 \\
\bronze & TransResUNet & \cellcolor{g}{81.27} & \cellcolor{g}{85.62} & \cellcolor{g}{62.36} & \cellcolor{lr}{79.13} & \cellcolor{y}{76.42} & \cellcolor{r}{84.20} & \cellcolor{y}{50.77} & \cellcolor{g}{87.69} & \cellcolor{lr}{35.19} & 69.47 \\
\#4 & CENet & \cellcolor{g}{82.70} & \cellcolor{g}{86.29} & \cellcolor{g}{62.15} & \cellcolor{lr}{78.57} & \cellcolor{r}{73.14} & \cellcolor{r}{84.10} & \cellcolor{g}{53.32} & \cellcolor{r}{71.06} & \cellcolor{lg}{48.03} & 69.31 \\
\#5 & MFMSNet & \cellcolor{g}{79.78} & \cellcolor{y}{84.47} & \cellcolor{lg}{61.27} & \cellcolor{y}{81.24} & \cellcolor{y}{76.33} & \cellcolor{lr}{85.08} & \cellcolor{r}{48.94} & \cellcolor{g}{83.67} & \cellcolor{y}{43.98} & 69.08 \\
\end{tabular}
}
\end{table*}
\renewcommand{\multirowsetup}{\centering}  
\begin{table*}[!t]
\centering

\caption{Average zeroshot performance of 100 u-shape medical image segmentation networks with U-Score. Source $\to$ Target. Baseline U-Net is highlighted (gray background), and statistical significance of p-value is highlighted: \colorbox{g}{p$<$0.0001}, \colorbox{lg}{p$<$0.001}, \colorbox{y}{p$<$0.01}, \colorbox{lr}{p$\leq$0.05},and \colorbox{r}{P $>$ 0.05 (Not significant)}}
\resizebox{0.93\linewidth}{!}
{
\begin{tabular}{r c | c c c | c c | c | c | c | c | c}
\Xhline{1.5px} % horizontal line

\multirow{2}{*}{Rank} & \multirow{2}{*}{Network} 
& \multicolumn{3}{c|}{Ultrasound} & \multicolumn{2}{c|}{Endoscopy} & \multicolumn{1}{c|}{Dermoscopy} & \multicolumn{1}{c|}{Fundus} & \multicolumn{1}{c|}{X-Ray} & \multicolumn{1}{c|}{Histopathology} & \multirow{2}{*}{Avg} \\
% \cline{3-9}
% \cline{3-22}
& & BUSI $\rightarrow$ BUS & BUSBRA $\rightarrow$ BUS & TNSCUI $\rightarrow$ TUCC & Kvasir $\rightarrow$ CVC300 & Kvasir $\rightarrow$ CVC-ClinicDB & ISIC2018 $\rightarrow$ PH2 & CHASE $\rightarrow$ Stare & Montgomery $\rightarrow$ NIH-test & Monusac $\rightarrow$ Tnbcnuclei & \\
\hline
\gold & LV-UNet & \cellcolor{r}{79.62} & \cellcolor{r}{79.18} & \cellcolor{g}{97.84} & \cellcolor{y}{96.96} & \cellcolor{r}{76.12} & \cellcolor{r}{99.23} & \cellcolor{r}{92.89} & \cellcolor{g}{92.65} & \cellcolor{r}{25.00} & 82.67 \\
\silver & LGMSNet & \cellcolor{y}{84.79} & \cellcolor{r}{78.74} & \cellcolor{g}{90.80} & \cellcolor{r}{76.60} & \cellcolor{lr}{89.51} & \cellcolor{r}{90.21} & \cellcolor{r}{80.85} & \cellcolor{g}{78.81} & \cellcolor{r}{39.90} & 78.41 \\
\bronze & MBSNet & \cellcolor{lr}{77.02} & \cellcolor{r}{74.05} & \cellcolor{g}{79.08} & \cellcolor{lr}{83.56} & \cellcolor{r}{79.75} & \cellcolor{r}{82.13} & \cellcolor{r}{87.32} & \cellcolor{r}{43.20} & \cellcolor{r}{25.09} & 70.02 \\
\#4 & CMUNeXt & \cellcolor{r}{9.41} & \cellcolor{r}{68.19} & \cellcolor{g}{78.86} & \cellcolor{r}{80.54} & \cellcolor{r}{60.39} & \cellcolor{lr}{88.85} & \cellcolor{y}{72.94} & \cellcolor{r}{52.42} & \cellcolor{y}{86.90} & 67.85 \\
\#5 & U-KAN & \cellcolor{y}{70.78} & \cellcolor{y}{72.11} & \cellcolor{g}{71.77} & \cellcolor{r}{62.44} & \cellcolor{r}{63.35} & \cellcolor{r}{50.87} & \cellcolor{r}{73.45} & \cellcolor{r}{55.98} & \cellcolor{lr}{64.91} & 64.96 \\
\end{tabular}
}
\end{table*}

\renewcommand{\multirowsetup}{\centering}  
\begin{table*}[!t]
\centering

\caption{Per-dataset source ranking of 100 u-shape medical image segmentation networks with IoU}
\resizebox{1\linewidth}{!}
{
\begin{tabular}{r | c c c | c c | c c | c c | c  }
\Xhline{1.5px} % horizontal line

\multirow{2}{*}{Rank} 
\multirow{2}{*}{Rank} & \multicolumn{3}{c|}{Ultrasound} & \multicolumn{2}{c|}{Dermoscopy} & \multicolumn{2}{c|}{Endoscopy} & \multicolumn{2}{c|}{Fundus}  & \multicolumn{1}{c}{Nuclear} \\
% \cline{3-9}
% \cline{3-22}
& BUSI & BUSBRA & TNSCUI & ISIC2018 & SkinCancer & Kvasir & Robotool & CHASE & DRIVE  & Cell\\

\hline
\gold	&	RWKV-UNet	&	RWKV-UNet	&	RWKV-UNet	&	RWKV-UNet	&	RWKV-UNet	&	Swin-umamba	&	MEGANet	&	CMU-Net	&	FCBFormer	&	MT-UNet	\\
\silver	&	PraNet	&	EViT-UNet	&	MEGANet	&	CFFormer	&	DA-TransUNet	&	VMUNet	&	RWKV-UNet	&	AttU-Net	&	MT-UNet	&	TransAttUnet	\\
\bronze	&	MobileUViT	&	CaraNet	&	MFMSNet	&	MEGANet	&	MSLAU-Net	&	UACANet	&	AURA-Net	&	U-Net	&	ColonSegNet	&	AURA-Net	\\
\#4	&	DA-TransUNet	&	MFMSNet	&	TA-Net	&	Swin-umamba	&	PraNet	&	CFFormer	&	TA-Net	&	UNet3+	&	UTNet	&	CA-Net	\\
\#5	&	MEGANet	&	TA-Net	&	UACANet	&	PraNet	&	FCBFormer	&	RWKV-UNet	&	EViT-UNet	&	Perspective-Unet	&	ESKNet	&	UTANet	\\
\#6	&	TransResUNet	&	FAT-Net	&	EViT-UNet	&	TransResUNet	&	EMCAD	&	FCBFormer	&	TransResUNet	&	UCTransNet	&	CMU-Net	&	ColonSegNet	\\
\end{tabular}
}
\end{table*}

\renewcommand{\multirowsetup}{\centering}  
\begin{table*}[!t]
\centering

\caption{Per-dataset source ranking of 100 u-shape medical image segmentation networks with IoU}
\resizebox{1\linewidth}{!}
{
\begin{tabular}{r | c c c | c c c | c c | c | c  }
\Xhline{1.5px} % horizontal line

\multirow{2}{*}{Rank} & \multicolumn{3}{c|}{Histopathology} & \multicolumn{3}{c|}{X-Ray} & \multicolumn{2}{c|}{MRI} & \multicolumn{1}{c|}{CT} & \multicolumn{1}{c}{OCT} \\
% % \cline{3-9}
% % \cline{3-22}
& DSB2018 & Glas & Monusac & Covidquex & Montgomery & DCA & ACDC & Promise & Synapse & Cystoidfluid \\

\hline
\gold	&	MT-UNet	&	EMCAD	&	MT-UNet	&	AURA-Net	&	RWKV-UNet	&	DA-TransUNet	&	CENet	&	RWKV-UNet	&	CENet	&	UNet3+\\
\silver	&	DoubleUNet	&	RWKV-UNet	&	RWKV-UNet	&	RWKV-UNet	&	DA-TransUNet	&	UTANet	&	Swin-umambaD	&	FCBFormer	&	Perspective-Unet	&	Swin-umamba\\
\bronze	&	TransAttUnet	&	CASCADE	&	UTANet	&	CaraNet	&	MEGANet	&	EViT-UNet	&	DoubleUNet	&	MFMSNet	&	G-CASCADE	&	UTANet\\
\#4	&	DCSAU-Net	&	MSLAU-Net	&	CA-Net	&	EViT-UNet	&	TransAttUnet	&	MFMSNet	&	RWKV-UNet	&	EViT-UNet	&	CASCADE	&	MMUNet\\
\#5	&	UTNet	&	UTANet	&	DDANet	&	TA-Net	&	RollingUnet	&	MEGANet	&	DDANet	&	Perspective-Unet	&	AURA-Net	&	H2Former\\
\#6	&	D-TrAttUnet	&	DDANet	&	TransAttUnet	&	MEGANet	&	DDANet	&	ESKNet	&	AttU-Net	&	MEGANet	&	MEGANet	&	Perspective-Unet\\
\end{tabular}
}
\end{table*}

\renewcommand{\multirowsetup}{\centering}  
\begin{table*}[!t]
\centering

\caption{Per-dataset target ranking of 100 u-shape medical image segmentation networks with IoU. Source $\to$ Target.}
\resizebox{0.93\linewidth}{!}
{
\begin{tabular}{r | c c c | c c | c | c | c | c }
\Xhline{1.5px} % horizontal line

\multirow{2}{*}{Rank} & \multicolumn{3}{c|}{Ultrasound} & \multicolumn{2}{c|}{Endoscopy} & \multicolumn{1}{c|}{Dermoscopy} & \multicolumn{1}{c|}{Fundus} & \multicolumn{1}{c|}{X-Ray} & \multicolumn{1}{c}{Histopathology} \\
% \cline{3-9}
% \cline{3-22}
& BUSI $\rightarrow$ BUS & BUSBRA $\rightarrow$ BUS & TNSCUI $\rightarrow$ TUCC & Kvasir $\rightarrow$ CVC300 & Kvasir $\rightarrow$ CVC-ClinicDB & ISIC2018 $\rightarrow$ PH2 & CHASE $\rightarrow$ Stare & Montgomery $\rightarrow$ NIH-test & Monusac $\rightarrow$ Tnbcnuclei \\
\hline
\#1 & Swin-umamba & MEGANet & MSLAU-Net & PraNet & PraNet & MSLAU-Net & RWKV-UNet & MEGANet & TA-Net\\
\#2 & EMCAD & DoubleUNet & MERIT & RWKV-UNet & DS-TransUNet & RWKV-UNet & DS-TransUNet & TransResUNet & CENet\\
\#3 & CENet & CENet & EViT-UNet & UACANet & CASCADE & G-CASCADE & SwinUNETR & CaraNet & U-Net++\\
\#4 & G-CASCADE & FCBFormer & Polyp-PVT & MERIT & Swin-umambaD & MERIT & FCBFormer & DA-TransUNet & EMCAD\\
\#5 & DA-TransUNet & CASCADE & LGMSNet & MFMSNet & EMCAD & MMUNet & MCA-UNet & PraNet & G-CASCADE\\
\#6 & PraNet & Polyp-PVT & G-CASCADE & TA-Net & TransResUNet & H2Former & UNETR & MFMSNet & UNetV2\\
\end{tabular}
}
\end{table*}

\renewcommand{\multirowsetup}{\centering}  
\begin{table*}[!t]
\centering

\caption{Per-dataset source ranking of 100 u-shape medical image segmentation networks with U-Score}
\resizebox{1\linewidth}{!}
{
\begin{tabular}{r | c c c | c c | c c | c c | c }
\Xhline{1.5px} % horizontal line

\multirow{2}{*}{Rank} 
& \multicolumn{3}{c|}{Ultrasound} & \multicolumn{2}{c|}{Dermoscopy} & \multicolumn{2}{c|}{Endoscopy} & \multicolumn{2}{c|}{Fundus}& \multicolumn{1}{c}{Nuclear}  \\
% \cline{3-9}
% \cline{3-22}
& BUSI & BUSBRA & TNSCUI & ISIC2018 & SkinCancer & Kvasir & Robotool & CHASE & DRIVE & Cell\\

\hline

\gold	&	LGMSNet	&	CMUNeXt	&	LGMSNet	&	LV-UNet	&	LGMSNet	&	LV-UNet	&	LV-UNet	&	LGMSNet	&	Tinyunet	&	Tinyunet	\\
\silver	&	CMUNeXt	&	LGMSNet	&	LV-UNet	&	LGMSNet	&	CMUNeXt	&	LGMSNet	&	LGMSNet	&	CMUNeXt	&	ULite	&	ULite	\\
\bronze	&	MBSNet	&	MBSNet	&	CMUNeXt	&	Mobile U-ViT	&	LV-UNet	&	Mobile U-ViT	&	Mobile U-ViT	&	MBSNet	&	LFU-Net	&	LGMSNet	\\
\#4	&	LV-UNet	&	LV-UNet	&	MBSNet	&	MBSNet	&	U-KAN	&	MambaUnet	&	MBSNet	&	U-RWKV	&	SimpleUNet	&	UNeXt	\\
\#5	&	Mobile U-ViT	&	Tinyunet	&	Tinyunet	&	DCSAU-Net	&	ULite	&	MBSNet	&	CMUNeXt	&	Tinyunet	&	LGMSNet	&	CMUNeXt	\\
\#6	&	MDSA-UNet	&	Mobile U-ViT	&	Mobile U-ViT	&	CMUNeXt	&	MUCM-Net	&	U-KAN	&	RWKV-UNet	&	Mobile U-ViT	&	CMUNeXt	&	U-RWKV	\\
\end{tabular}
}
\end{table*}

\renewcommand{\multirowsetup}{\centering}  
\begin{table*}[!t]
\centering

\caption{Per-dataset source ranking of 100 u-shape medical image segmentation networks with U-Score}
\resizebox{1\linewidth}{!}
{
\begin{tabular}{r | c c c | c c c | c c | c |  c }
\Xhline{1.5px} % horizontal line

\multirow{2}{*}{Rank} 
& \multicolumn{3}{c|}{Histopathology}  & \multicolumn{3}{c|}{X-Ray} & \multicolumn{2}{c|}{MRI} & \multicolumn{1}{c|}{CT} & \multicolumn{1}{c}{OCT} \\
% \cline{3-9}
% \cline{3-22}
& DSB2018 & Glas & Monusac& Covidquex & Montgomery & DCA & ACDC & Promise & Synapse & Cystoidfluid \\

\hline

\gold &	LGMSNet	&	Tinyunet	&	MBSNet	&	LV-UNet	&	Tinyunet	&	LV-UNet	&	MBSNet	&	UNeXt	&	LGMSNet	&	ULite\\
\silver &	ULite	&	LGMSNet	&	U-RWKV	&	MBSNet	&	LGMSNet	&	CMUNeXt	&	Tinyunet	&	LV-UNet	&	Tinyunet	&	Tinyunet\\
\bronze &	U-RWKV	&	LV-UNet	&	LGMSNet	&	LGMSNet	&	MBSNet	&	U-RWKV	&	LGMSNet	&	LGMSNet	&	MBSNet	&	UNeXt\\
\#4 &	MBSNet	&	MBSNet	&	CMUNeXt	&	CMUNeXt	&	CMUNeXt	&	LGMSNet	&	CMUNeXt	&	MBSNet	&	CMUNeXt	&	LV-UNet\\
\#5 &	CMUNeXt	&	CMUNeXt	&	SimpleUNet	&	Tinyunet	&	UNeXt	&	MBSNet	&	SimpleUNet	&	CMUNeXt	&	Mobile U-ViT	&	MBSNet\\
\#6 &	Tinyunet	&	SimpleUNet	&	LV-UNet	&	Mobile U-ViT	&	U-RWKV	&	Tinyunet	&	LV-UNet	&	U-RWKV	&	U-KAN	&	CMUNeXt\\
\end{tabular}
}
\end{table*}

\renewcommand{\multirowsetup}{\centering}  
\begin{table*}[!t]
\centering

\caption{Per-dataset target ranking of 100 u-shape medical image segmentation networks with U-Score. Source $\to$ Target.}
\resizebox{0.93\linewidth}{!}
{
\begin{tabular}{r | c c c | c c | c | c | c | c }
\Xhline{1.5px} % horizontal line

\multirow{2}{*}{Rank} & \multicolumn{3}{c|}{Ultrasound} & \multicolumn{2}{c|}{Endoscopy} & \multicolumn{1}{c|}{Dermoscopy} & \multicolumn{1}{c|}{Fundus} & \multicolumn{1}{c|}{X-Ray} & \multicolumn{1}{c}{Histopathology} \\
% \cline{3-9}
% \cline{3-22}
& BUSI $\rightarrow$ BUS & BUSBRA $\rightarrow$ BUS & TNSCUI $\rightarrow$ TUCC & Kvasir $\rightarrow$ CVC300 & Kvasir $\rightarrow$ CVC-ClinicDB & ISIC2018 $\rightarrow$ PH2 & CHASE $\rightarrow$ Stare & Montgomery $\rightarrow$ NIH-test & Monusac $\rightarrow$ Tnbcnuclei \\
\hline
\gold & LGMSNet & LV-UNet & LV-UNet & LV-UNet & LGMSNet & LV-UNet & ULite & LV-UNet & CMUNeXt\\
\silver & LV-UNet & LGMSNet & LGMSNet & MBSNet & MBSNet & LGMSNet & LV-UNet & MUCM-Net & MALUNet\\
\bronze & MBSNet & SwinUNETR & MBSNet & CMUNeXt & LV-UNet & CMUNeXt & SwinUNETR & LGMSNet & MDSA-UNet\\
\#4 & Mobile U-ViT & Mobile U-ViT & CMUNeXt & LGMSNet & MambaUnet & U-RWKV & CMUNeXt & MALUNet & DCSAU-Net\\
\#5 & U-KAN & MBSNet & U-RWKV & MambaUnet & Mobile U-ViT & MBSNet & MUCM-Net & SwinUNETR & UltraLight-VM-UNet\\
\#6 & SwinUNETR & UNeXt & U-KAN & Mobile U-ViT & U-KAN & CFPNet-M & CFPNet-M & Mobile U-ViT & U-KAN\\
\end{tabular}
}
\end{table*}
\clearpage 

\clearpage
\bibliography{main}

% \appendix
% \input{sections/appendix}
% \appendix
\end{document}